%% file: AAAA_batched_bandits_camera_ready.tex
\documentclass{article}

\usepackage{commands} 
\usepackage[font=footnotesize,labelfont=bf]{caption}

\usepackage{xr-hyper}
\usepackage{hyperref}
     \usepackage[final,nonatbib]{neurips_2020}

\usepackage[utf8]{inputenc} % allow utf-8 input
\usepackage[T1]{fontenc}    % use 8-bit T1 fonts
\usepackage{hyperref}       % hyperlinks
\usepackage{url}            % simple URL typesetting
\usepackage{booktabs}       % professional-quality tables
\usepackage{amsfonts}       % blackboard math symbols
\usepackage{nicefrac}       % compact symbols for 1/2, etc.
\usepackage{microtype}      % microtypography
\usepackage{color}

\title{Inference for Batched Bandits}

\author{%
  Kelly W. Zhang \\
  Department of Computer Science\\
  Harvard University\\
  \texttt{kellywzhang@seas.harvard.edu} \\
   \And
  Lucas Janson \\
   Departments of Statistics \\
    Harvard University\\
   \texttt{ljanson@fas.harvard.edu} \\
   \AND
   Susan A. Murphy \\
   Departments of Statistics and Computer Science \\
    Harvard University\\
   \texttt{samurphy@fas.harvard.edu} \\
}

\begin{document}

\maketitle

\begin{abstract}
As bandit algorithms are increasingly utilized in scientific studies and industrial applications, there is an associated increasing need for reliable inference methods based on the resulting adaptively-collected data.
In this work, we develop methods for inference on data collected in batches using a bandit algorithm. We first prove that the ordinary least squares estimator (OLS), which is asymptotically normal on independently sampled data, is \textit{not} asymptotically normal on data collected using standard bandit algorithms when there is no unique optimal arm. This asymptotic non-normality result implies that the naive assumption that the OLS estimator is approximately normal can lead to Type-1 error inflation and confidence intervals with below-nominal coverage probabilities. Second, we introduce the Batched OLS estimator (BOLS) that we prove is (1) asymptotically normal on data collected from both multi-arm and contextual bandits and (2) robust to non-stationarity in the baseline reward.
\end{abstract}

\section{Introduction}
Due to their regret minimizing guarantees bandit algorithms have been increasingly used in in real-world sequential decision-making problems, like online advertising \cite{li2010contextual}, mobile health \cite{yom2017encouraging}, and online education \cite{rafferty2019statistical}.
However, for many real-world problems it is not enough to just minimize regret on a particular problem instance. 
For example, suppose we have run an online education experiment using a bandit algorithm where we test different types of teaching strategies. 
When designing a new online course, ideally we could use the data from the previous experiment to inform the design, e.g., under-performing arms could be eliminated or modified.
Moreover, to help others designing online courses we would like to be able to publish our findings about how different teaching strategies compare in their performance.
This example demonstrates the need for statistical inference methods on bandit data, which allow practitioners to draw generalizable knowledge from the data they have collected (e.g., how much better one teaching strategy is compared to another) for the sake of scientific discovery and informed decision making.

In this work we will focus on methods to construct confidence intervals for the margin---the difference in expected rewards of two bandit arms---from batched bandit data. 
Rather than constructing high probability confidence intervals, we are interested in constructing confidence intervals by using the asymptotic distribution of estimators to approximate their finite sample distribution. Asymptotic approximation methods for statistical inference has a long history of being successful in science and leads to much narrower confidence intervals than those constructed using high probability bounds.
Most statistical inference methods based on asymptotic approximation assume that treatments are assigned independently \cite{imbens2015causal}.
\bo{However, bandit data violates this independence assumption because it is collected \textit{adaptively}, meaning previous actions and rewards inform future action selections.}
The non-independence makes statistical inference more challenging, e.g., estimators like the sample mean are often biased on bandit data \cite{nie2017adaptively,shin2019sample}. 

Throughout, we focus on the batched bandit setting, in which arms of the bandit are pulled in batches. For our asymptotic analysis we fix the total number of batches, $T$, and allow the arm pulls in each batch, $n$, to go to infinity. \textit{Note that we do not need or expect $n$ to go to infinity for real-world experiments; we use the asymptotic distribution of estimators to approximate their finite-sample distribution when constructing confidence intervals.} 
We focus on the batched setting because it closely reflects many of the problem settings where bandit algorithms are applied. For example, in many mobile health \cite{yom2017encouraging,klasnja2019efficacy,liao2020personalized} and online education problems \cite{kizilcec2020scaling,rafferty2019statistical} multiple users use apps / take courses simultaneously, so a batch corresponds to the number of unique users the bandit algorithm acts on at once. The batched setting is even common in online recommendations and advertising because it is impractical to update the bandit after every action if many users visit the site simultaneously \cite{tang2014ensemble,schwartz2017customer,han2020sequential,li2019cascading}. 
In many such experimental settings the length of the study, $T$, cannot be arbitrarily adjusted, e.g., in online education, courses generally cannot be made arbitrarily long, and clinical trials often run for a standard amount of time that depends on the domain science (e.g. the length of mobile health studies is a function of the scientific community's belief in how long it should take for users to form a habit). On the other hand, the number of users, $n$, can in principle grow as large as funding allows. 

Additionally, in our batched setting, we assume that the means of the arms can change over time, i.e., from batch to batch, which reflects the temporal non-stationarity that is prevalent in many real world bandit application problems. 
For example, in online recommendation systems, the click through rate of a given recommendation typically varies over time, e.g., breaking news articles become less popular over time \cite{tang2014ensemble,li2019cascading}. Online education and mobile health are also highly non-stationary problems because users tend to disengage over time, so the same notification may be much less effective if sent near the end of an experiment than sent near the beginning \cite{eysenbach2005law,kizilcec2013deconstructing,druce2019maximizing}.
Our statistical inference method does not need to assume that the number of stationary time periods in the experiment is large and is robust to temporal non-stationarity from batch to batch.

\bo{The first contribution of this work is proving that on bandit data, rather surprisingly, whether standard estimators are asymptotically normal can depend on whether the margin is zero.}
%We prove that when data is collected using common bandit algorithms, the action selection probabilities only concentrate if there is a unique optimal arm.
We prove that for common bandit algorithms, the arm selection probabilities only concentrate if there is a unique optimal arm.
Thus, for two-arm bandits, the arm selection probabilities do not concentrate when the margin---the difference in the expected rewards between the arms---is zero.
We show that this leads the ordinary least squares (OLS) estimator to be asymptotically normal when the margin is non-zero, and asymptotically \textit{not} normal when the margin is zero.
\textit{Since the OLS estimator does not converge uniformly (over values of the margin), standard inference methods (normal approximations, bootstrap\footnote{Note that the validity of bootstrap methods rely on uniform convergence \cite{romano2012uniform}.}) can lead to inflated Type-1 error and unreliable confidence intervals on bandit data.}

\bo{The second contribution of this work is introducing the Batched OLS (BOLS) estimator, which can be used for reliable inference---even in non-stationary settings---on data collected with batched bandits.}
%Regardless of whether the margin is zero or not, the BOLS estimator for the margin for both multi-arm and contextual bandits is asymptotically normal. 
We prove that, regardless of whether the margin is zero or not, the BOLS estimator for the margin for both multi-arm and contextual bandits is asymptotically normal and thus can be used for both hypothesis testing and obtaining confidence intervals.
Moreover, BOLS is also automatically robust to non-stationarity in the rewards and can be used for constructing valid confidence intervals even if there is non-stationarity in the baseline reward, i.e., if the rewards of the arms change from batch to batch, but the margin remains constant.
If the margin itself is also non-stationary, BOLS can also be used for constructing simultaneous confidence intervals for the margins for each batch.

\section{Related Work}

\bo{Batched Bandits} ~
Much work on batched bandits focuses on minimizing regret \cite{perchet2016batched,gao2019batched} or identifying the best arm with high probability \cite{agarwal2017learning,jun2016top}. The best arm identification literature utilizes high probability confidence bounds to construct confidence intervals for bandit parameters; we will discuss this method in the next section.
%Note that best arm identification is distinct from obtaining a confidence interval for the margin, as the former identifies the best arm with high probability (assuming there is a best arm), while the latter provides guarantees regarding the sign and \textit{magnitude} of the margin. %can be used to test whether one arm is better than the other and 
Note that in contrast to other batched bandit literature that allow batch sizes to be adjusted adaptively \cite{perchet2016batched}, here we do not have adaptive control over the batch sizes.

Batched bandits are closely related to multistage adaptive clinical trials, in which between each batch (or stage of the trial) the data collection procedure can be adjusted depending on the outcome of the previous batches.
Our Batched OLS estimator is most closely related to ``stage-wise" p-values for group sequential trials that are computed on each stage separately \cite{wassmer2016group}.
p-value combination tests are commonly used to combine stage-wise p-values, when the sequence of p-values are shown to be independent or \textit{p-clud}, meaning that under the null each p-value has a Uniform(0,1) distribution conditional on past p-values \cite{wassmer2016group}.
%One of the most common methods used for inference using stage-wise p-values are p-value combination tests on p-values that are shown to be \textit{p-clud}, meaning that under the null each p-value has a Uniform(0,1) distribution conditional on past p-values \cite{wassmer2016group,brannath2012probabilistic}.
%liu2006design
\cite{liu2002unified} formally establish the independence of stage-wise p-values for two-stage trials in which there are a countable number of adaptive rules; note that this rules out bandit algorithms with real-valued arm selection probabilities, like Thompson Sampling.
\cite{brannath2012probabilistic} establishes the p-clud property for two-stage adaptive clinical trials under the assumption that the distribution of the second stage data is known conditioned on the decision rule and first stage data under the null hypothesis.
Neither of these methods are sufficient for obtaining independent p-values for adaptive trials (1) with an arbitrary number of stages, (2) where exact distribution of rewards is unknown, and (3) where the action selection probabilities can be real numbers, like for Thompson Sampling.

\bo{High Probability Confidence Intervals} ~
High probability confidence intervals provide stronger guarantees than those constructed using asymptotic approximations. In particular, these bounds are guaranteed to hold for finite number of observations and often even hold uniformly over all $n$ and $T$. These types of bounds are used throughout the bandit and reinforcement learning literature to construct confidence intervals for bandit parameters \cite{howard2018uniform,kaufmann2018mixture}, prove regret bounds \cite{abbasi2011improved,lattimore2020bandit}, and provide guarantees regarding best arm identification \cite{jamieson2014lil,jamieson2014best}. The primary drawback of high probability confidence intervals is that they are much more conservative than those constructed using asymptotic approximations. This means that many more observations will be needed to get a confidence interval of the same width or for a statistical test to have the same power when using high probability confidence intervals compared to those constructed using asymptotic approximation. Since the cost of increasing the the number of users in a study can be large, being able to construct narrow---yet reliable---confidence intervals is crucial to many applications.

In our simulations we compare our method to high probability confidence bounds constructed using the self-normalized martingale bound of \cite{abbasi2011improved}. This bound is guaranteed to hold on adaptively collected data and is commonly used in the proof of regret bounds for bandit algorithms. We find that all the approaches based on asymptotic approximations (which we discuss next), significantly outperform the statistical test constructed using a self-normalized martingale bound in terms of power.
Moreover, despite the weaker guarantees of statistical inference based on asymptotic approximations, they are generally able to provide reliable coverage of confidence intervals and type-1 error control. 

% High probability confidence intervals are bounds on the errors of estimators, e.g., the sample mean, that are guaranteed to hold with high probability---even for finite number of observations.

\bo{Adaptive Inference based on Asymptotic Approximations} ~
%Much of the recent work on inference for adaptively-collected data focuses on characterizing and reducing the bias of the OLS estimator in finite samples \cite{deshpande, nie2017adaptively, shin2019sample}. 
%\cite{deshpande} and \cite{athey} develop estimators that they prove are asymptotically Normal on adaptively collected data under certain conditions.
%\cite{deshpande} develop the W-decorrelated estimator which is an adjusted version of OLS that requires choosing a tuning parameter $\lambda$, which allows practitioners to trade off bias for variance. 
%In the two-arm bandit setting, we find that the W-decorrelated estimator down-weights samples that occur later in the study and up-weights samples from earlier in the study; see Appendix \ref{appendix:wdecorrelated}.
%The Adaptively-Weighted Augmented-Inverse-Probability-Weighted Estimator (AW-AIPW) for multi-arm bandits reweights the samples of a regular AIPW estimator with adaptive weights that are non-anticipating \cite{athey}.
%Note that neither the W-decorrelated estimator nor the AW-AIPW estimator have guarantees in non-stationary settings.
A common approach in the literature for performing inference on bandit data is to use adaptive weights, which are weights that are a function of the history.
An early example of using adaptive weights is that of \cite{luedtke2016statistical} and \cite{luedtke2018parametric}, who use adaptive weights in estimating the expected reward under the optimal policy when one has access to i.i.d. observational data.
%They cannot use standard inferential techniques because they must estimate the optimal policy and the expected reward under the optimal policy using the same data. 
They use an Augmented-Inverse-Probability-Weighted %based 
estimator with adaptive weights that are a function of the estimated standard deviation of the reward.
%As mentioned in \cite{athey}, 
\cite{luedtke2018parametric} conjecture that their approach can be adapted to the adaptive sampling case.  %SAM: need to avoid athey's language otherwise plagurism!
%, but do not pursue it further.
Subsequently \cite{athey}  developed the adaptively weighted method for inference on bandit data to produce the Adaptively-Weighted Augmented-Inverse-Probability-Weighted Estimator (AW-AIPW) for data collected via multi-arm bandits.
They prove a central limit theorem (CLT) for AW-AIPW when the adaptive weights satisfy certain conditions. 
Note, however, the AW-AIPW estimator does not have guarantees in non-stationary settings.
%In the stationary multi-arm bandit case, we make similar assumptions to those that \cite{athey} use to prove asymptotic Normality of the AW-AIPW estimator; however, the AW-AIPW estimator does not have guarantees in non-stationary settings.
%establish sufficient conditions for the adaptive weights for their AW-AIPW central limit theorem (CLT).
%For their CLT to hold, they require that either the action selection probabilities concentrate (in Section \ref{section:olsnonnormal} we discuss why this condition fails to hold for common bandit algorithms) or that the model for their AIPW estimator is consistent.

Adaptive weights are also used by \cite{deshpande} to form the W-decorrelated estimator, a debiased version of OLS, that is asymptotically normal.
In the multi-arm bandit setting, the adaptive weights are a function of the number of times an arm was chosen previously.
We found that in the two-arm setting, the W-decorrelated estimator down-weights rewards from later in the study (Appendix \ref{appendix:wdecorrelated}).
%In the multi-arm bandit setting, the adaptive weights are a function a chosen parameter $\lambda$ and the number of times an arm was chosen previously.
%develop the W-decorrelated estimator, which is an adjusted version of OLS, with adapative weights. In the multi-arm bandit case, their weights are a function of a tuning parameter $\lambda$ and the number of times that arm was sampled so far in the study. 
%The choice of $\lambda$ allows practitioners to trade off bias for variance. 
%They prove a CLT for their estimator when $\lambda$ is a high probability lower bound of the minimum eigenvalue of $\under{\boldX}^\top \under{\boldX}$.
%For two-arm bandits, we find that the W-decorrelated estimator down-weights rewards from later in the study (Appendix \ref{appendix:wdecorrelated}). %and up-weights rewards from earlier in the study
%In the two-arm bandit setting, we find that the W-decorrelated estimator down-weights samples that occur later in the study and up-weights samples from earlier in the study (Appendix \ref{appendix:wdecorrelated}).
\cite{deshpande2019online} introduce the Online Debiased Estimator that also has bias guarantees on adaptive data, but in the more challenging high-dimensional linear regression setting.
% which debiases estimates of parameters of high-dimensional linear regression models on adaptive data;  % bounds on the bias and 
They prove the asymptotic normality of their estimator in the Gaussian autoregressive time series and the two-batch settings. %on the bias and 
%\cite{deshpande2019online} extend the W-decorrelated estimator to the high-dimensional linear regression setting when the parameters are sparse; they propose the online debiasing procedure, which regarding the unbiasedness and asymptotic normality of their estimator in the Gaussian autoregressive time series and the two batch settings.
%statistical inference on linear regression parameters on high-dimensional adaptive data when using LASSO regularization; also consider the two batch setting.
Note that none of these estimation methods have guarantees in non-stationary bandit settings. % \sam{we got that email from desphande's student about extension to non-stationary settings.  they have an arXiv paper--I doubt we can get away with not mentioning this.   Did their arXiv paper appear after or before ours?  The way we describe their work would be affected by the answer to prior question.}
\cite{lai1982least} provide conditions under which  the OLS estimator is asymptotically normal on adaptively collected data. 
However, as noted in \cite{villar2015multi, deshpande, athey}, classical inference techniques developed for i.i.d. data often empirically have inflated Type-1 error on bandit data.
%However, \cite{villar2015multi} examine a variety of bandit algorithms and empirically find that tests for i.i.d. data have inflated finite-sample Type-1 error rates when the data is collected using these algorithms.
%Lastly, \cite{lai1982least} prove that the OLS estimator is asymptotically normal on adaptively collected data under certain conditions. 
%In Section \ref{section:olsnormal}, we examine the conditions of the CLT of \cite{lai1982least} and determine settings in which the necessary conditions are satisfied. 
%\sam{I don't think Lai proved the conditions are ``necessary''  please check.   let's not call conditions necessary unless they are!   I suggest to replace the prior sentences by:  (please correct the following for accuracy)}
In Section \ref{section:olsnormal}, we discuss the restrictive nature of \cite{lai1982least}'s CLT conditions.
%Then in Section \ref{section:olsnonnormal}, we characterize settings in which particular CLT conditions are violated and to our knowledge, provide a first proof showing that these conditions are necessary, that is, violations lead to  asymptotic non-normality of the OLS estimator on bandit data.
%Our results theoretically justify the results of \cite{villar2015multi}, who examine a variety of bandit algorithms and empirically find that tests for i.i.d. data have inflated finite-sample Type-1 error rates when the data is collected using these algorithms.
%\sam{above reads as outline of paper--if this is intentional then need to also refer to Section 5 with our estimator}

%\todo[inline]{here}
%\begin{itemize}
%	\item Leudke and van der laan
%	\item Athey - constant weighting? (adaptive weighting?)
%	\item p-value combination - p-clud
%\end{itemize}
%\todo[inline]{here}

\section{Problem Formulation}

\paragraph{Setup and Notation}
%Though our results generalize to $K$-arm, contextual bandits (see Section \ref{section:contextBOLS}), we first focus on the two-arm bandit for expositional simplicity.
%Suppose there are $T$ timesteps or batches in a study.
%In each timestep $t \in [1 \colon T]$, we select $n$ binary actions $\{ A_{t,i}^{(n)} \}_{i=1}^n \in \{0, 1\}^n$.
%%Note that our asymptotic results will hold even if the number of arm pulls in each batch vary, as long as the number of arm pulls is independent of the history of actions and rewards.
%We then observe independent rewards $\{ R_{t,i}^{(n)} \}_{i=1}^n$, one for each action sampled.
%Note that the distribution of these random variables changes with the batch size, $n$.
%For example, the distribution of the actions one chooses for the $2^{\TN{nd}}$ batch, $\{ A_{2,i}^{(n)} \}_{i=1}^n$, may change if one has observed $n=10$ vs. $n=100$ samples $( A_{1,i}^{(n)}, R_{1,i}^{(n)} )_{i=1}^n$ in the first batch.
%We introduce an $(n)$ superscript on these random variables as a reminder that their distribution changes with the batch size $n$. 
%%We introduce an $(n)$ superscript on these random variables as a reminder that their distribution changes with $n$, but often drop the $(n)$ superscript for readability.
%%We introduce an $(n)$ superscript on these random variables as a reminder that their distribution changes with $n$; however, we often drop the $(n)$ superscript for readability.
Though our results generalize to $K$-arm, contextual bandits (see Section \ref{section:contextBOLS}), we first focus on the two-arm bandit for expositional simplicity.
Suppose there are $T$ timesteps or batches in a study.
In each batch $t \in [1 \colon T]$, we select $n$ binary actions $\{ A_{t,i} \}_{i=1}^n \in \{0, 1\}^n$.
%Note that our asymptotic results will hold even if the number of arm pulls in each batch vary, as long as the number of arm pulls is independent of the history of actions and rewards.
We then observe independent rewards $\{ R_{t,i} \}_{i=1}^n$, one for each action selected.
Note that the distribution of these random variables changes with the batch size, $n$.
For example, the distribution of the actions one chooses for the $2^{\TN{nd}}$ batch, $\{ A_{2,i} \}_{i=1}^n$, may change if one has observed $n=10$ vs. $n=100$ samples $\{ A_{1,i}, R_{1,i} \}_{i=1}^n$ in the first batch.
For readability, we omit indexing random variables by $n$, except for the variables $H_{t-1}^{(n)}$ and $\pi_t^{(n)}$, and filtrations like $\G_{t-1}^{(n)}$ to be introduced next.
%We introduce an $(n)$ superscript on these random variables as a reminder that their distribution changes with the batch size $n$. 
%We introduce an $(n)$ superscript on these random variables as a reminder that their distribution changes with $n$, but often drop the $(n)$ superscript for readability.
%We introduce an $(n)$ superscript on these random variables as a reminder that their distribution changes with $n$; however, we often drop the $(n)$ superscript for readability.

%We define two-arm bandit algorithms as functions $\{ \MC{A}_t \}_{t=1}^T$ such that $\MC{A}_t ( H_{t-1}^{(n)} ) =: \pi_{t}^{(n)} \in [0, 1]$, where $H_{t-1}^{(n)} := \{ A_{t',i}, R_{t',i} : i \in [1 \colon n], t' \in [1 \colon t-1] \}$ is the history prior to batch $t$.  \sam{if we need to save space, I don't think we need to introduce $\MC{A}_t$ as we could get away with just using the $\pi_t^{(n)} := \PP( A_{t,i} = 1 | \HH_{t-1}^{(n)})$ notation}
%For the two-arm bandit, we define the \textit{action selection probability}, $\pi_t^{(n)}$, where $\pi_t^{(n)} := \PP( A_{t,i} = 1 | \HH_{t-1}^{(n)})$.
%For each $t \hspace{-0.75mm} \in \hspace{-0.75mm} [1 \colon T]$, the bandit selects actions $\{ A_{t,i} \}_{i=1}^n \hspace{-0.75mm} \iidsim \hspace{-0.75mm} \TN{Bernoulli} ( \pi_t^{(n)} )$ conditional on $H_{t-1}^{(n)}$, where $H_{t-1}^{(n)} := \{ A_{t',i}, R_{t',i} \}_{i=1,t'=1}^{i=n,t'=t-1}$ is the history prior to batch $t$.
For each $t \hspace{-0.75mm} \in \hspace{-0.75mm} [1 \colon T]$, the bandit selects actions $\{ A_{t,i} \}_{i=1}^n \hspace{-0.75mm} \iidsim \hspace{-0.75mm} \TN{Bernoulli} ( \pi_t^{(n)} )$ conditional on $H_{t-1}^{(n)} := \{ A_{t',i}, R_{t',i} \}_{i=1,t'=1}^{i=n,t'=t-1}$, the history prior to batch $t$.
%$H_{t-1}^{(n)} := \{ A_{t',i}, R_{t',i} : i \in [1 \colon n], t' \in [1 \colon t-1] \}$
Note, the \textit{action selection probability} $\pi_t^{(n)} := \PP( A_{t,i} = 1 | H_{t-1}^{(n)})$ depends on the history $H_{t-1}^{(n)}$.
%Note, $\pi_t^{(n)} := \PP( A_{t,i} = 1 | H_{t-1}^{(n)})$, the \textit{action selection probability}, is a function of the history $H_{t-1}^{(n)}$.% and that $\pi_t^{(n)} := \PP( A_{t,i} = 1 | H_{t-1}^{(n)})$.
%, so we often refer to $\pi_t^{(n)}$ as the \textit{action selection probability}. 
We assume the following conditional mean for rewards:
 \begin{equation}
 \label{eqn:meanReward}
 \E \big[ R_{t,i} \big| H_{t-1}^{(n)}, A_{t,i} \big] 
= (1 - A_{t,i} ) \beta_{ t, 0 } +  A_{t,i} \beta_{ t, 1 } .
\end{equation}
%We define two-arm bandit algorithms as functions $\{ \MC{A}_t \}_{t=1}^T$ such that $\MC{A}_t ( H_{t-1}^{(n)} ) =: \pi_{t}^{(n)} \in [0, 1]$, 
%where $H_{t-1}^{(n)} := \big\{ A_{t',i}^{(n)}, R_{t',i}^{(n)} : i \in [1 \colon n], t' \in [1 \colon t-1] \big\}$ is the history prior to batch $t$.  
%The bandit selects actions such that for each $t \in [1 \colon T]$, $\{ A_{t,i}^{(n)} \}_{i=1}^n \iidsim \TN{Bernoulli} ( \pi_t^{(n)} )$ conditionally on $ H_{t-1}^{(n)} $.  
%Note, $\pi_t^{(n)} := \PP( A_{t,i} = 1 | \HH_{t-1}^{(n)})$. We assume the following conditional mean for the rewards:
% \begin{equation}
% \label{eqn:meanReward}
% \E \big[ R_{t,i}^{(n)} \big| H_{t-1}^{(n)}, A_{t,i}^{(n)} \big] 
%= (1 - A_{t,i}^{(n)} ) \beta_{ t, 0 } +  A_{t,i}^{(n)} \beta_{ t, 1 } .
%\end{equation}
Note in equation \eqref{eqn:meanReward} we condition on $H_{t-1}^{(n)}$ because the conditional mean of the reward does not depend on prior rewards or actions.
%\sam{make a statement here about why we are conditioning on $H_{t-1}^{(n)}$ in above--that is the conditional mean of the  reward does not depend on prior rewards or actions.}
Let $\boldX_{t,i} := [1 - A_{t,i}, A_{t,i} ]^\top \in \real^2$; note $\boldX_{t,i}$ is higher dimensional when we add more arms and/or context variables.
%Also define $N_{t,1} := \sum_{i=1}^n A_{t,i}$ and $N_{t,0} := \sum_{i=1}^n (1-A_{t,i})$, the number of times each arm is sampled in the $t^{ \TN{th} }$ batch. 
We define the errors as $\epsilon_{t,i} := R_{t,i} - ( \boldX_{t,i} )^\top \bs{\beta}_t$.
Equation \eqref{eqn:meanReward} implies that $\{ \epsilon_{t,i} : i \in [1 \colon n], t \in [1 \colon T] \}$ are a martingale difference array with respect to the filtration $\{ \G_{t}^{(n)} \}_{t=1}^T$, where $\G_t^{(n)} := \sigma \big( H_{t-1}^{(n)} \cup \{ A_{t,i} \}_{i=1}^n \big)$; thus, $\E[ \epsilon_{t,i} | \G_{t-1}^{(n)} ] = 0, \forall t, i, n$.
%Let $\boldX_{t,i}^{(n)} := [1 - A_{t,i}^{(n)}, A_{t,i}^{(n)} ]^\top \in \real^2$ ($\boldX_{t,i}^{(n)}$ will be higher dimensional when we add more arms and/or context variables).
%Also define $N_{t,1}^{(n)} := \sum_{i=1}^n A_{t,i}^{(n)}$ and $N_{t,0}^{(n)} := \sum_{i=1}^n (1-A_{t,i}^{(n)})$, the number of times each arm is sampled in the $t^{ \TN{th} }$ batch. 
%We define the errors as $\epsilon_{t,i}^{(n)} := R_{t,i}^{(n)} - ( \boldX_{t,i}^{(n)} )^\top \bs{\beta}_t$.
%Equation \eqref{eqn:meanReward} implies that $\big\{ \epsilon_{t,i}^{(n)} : i \in [1 \colon n], t \in [1 \colon T] \big\}$ are a martingale difference array with respect to the filtration $\{ \G_{t}^{(n)} \}_{t=1}^T$, where $\G_t^{(n)} := \sigma \big( H_{t-1}^{(n)} \cup \{ A_{t,i}^{(n)} \}_{i=1}^n \big)$. 
%So $\E[ \epsilon_{t,i}^{(n)} | \G_{t-1}^{(n)} ] = 0, \forall t, i, n$.
%\paragraph{Parameters of Interest}
The parameters $\bs{\beta}_t= ( \beta_{t, 0}, \beta_{t, 1} )$ can change across  batches $t \in [1 \colon T]$, which allows for non-stationarity between batches.
Assuming that $\bs{\beta}_t = \bs{\beta}_{t'}$ for all $t, t' \in [1 \colon T]$ simplifies to the stationary mean case.
%Our goal is to estimate and obtain confidence intervals for $\bs{ \beta }_t$ for $t \in [1 \colon T]$ when the data is collected using bandit algorithms. % like Thompson Sampling and $\epsilon$-greedy.
%Specifically we want an estimator that has an asymptotic distribution that approximates its finite-sample distribution well, so we can both perform hypothesis testing for the zero margin and construct a confidence interval for the margin. 
%Specifically we want an estimator that has an asymptotic distribution that approximates its finite-sample distribution well, so we can construct a confidence interval for the margin and perform hypothesis testing for the zero margin. 
 
\paragraph{Action Selection Probability Constraint (Clipping)}
%When running an experiment with adaptive sampling, it is desirable to minimize regret as much as possible.
%However, in order to perform inference on the resulting data it is also necessary to guarantee that the bandit algorithm explores sufficiently.
In order to perform inference on bandit data it is necessary to guarantee that the bandit algorithm explores sufficiently. 
%Greater exploration in the multi-arm bandit case means sampling the treatments with closer to equal probability, rather than sampling one action almost exclusively. 
For example, the CLTs for both the W-decorrelated \cite{deshpande} and the AW-AIPW \cite{athey} estimators have conditions that implicitly require that the bandit algorithms cannot sample any given action with probability that goes to zero or one arbitrarily fast.
Greater exploration also increases the power of statistical tests regarding the margin \cite{yao2020power}. %i.e., it makes it more probable that a true discovery will be made from the collected data
Moreover, if there is non-stationarity in the margin between batches, it is desirable for the bandit algorithm to continue exploring. %, so it can adjust to these changes and not almost exclusively sample one arm that is no longer the best.
We explicitly guarantee exploration by constraining the probability that any given action can be sampled (see Definition \ref{def:clipping}).
%We explicitly guarantee exploration by constraining the probability that any given action can be sampled, as per Definition \ref{def:clipping} below.
%In Definition \ref{def:clipping} below, we explain the constraint we put  bandit algorithms. 
We allow the action selection probabilities $\pi_t^{(n)}$ to converge to $0$ and/or $1$ at some rate.
%In particular, we will require that for some non-random sequences $\pi_{\min}^{(n)}, \pi_{\max}^{(n)}$ where $0 < \pi_{\min}^{(n)} \leq \pi_{\max}^{(n)} < 1$, there exists some $m$ such that for all $n \geq m$, $\pi_t^{(n)} \in [\pi_{\min}^{(n)}, \pi_{\max}^{(n)}]$.
%In particular, we will require that for some non-random sequence $\pi_{\min}^{(n)}$ where $0 < \pi_{\min}^{(n)} \leq 0.5$, there exists some $m$ such that for all $n \geq m$, $\pi_t^{(n)} \in [\pi_{\min}^{(n)}, 1-\pi_{\min}^{(n)}]$.
%Note that we allow $\pi_{\min}^{(n)}$ to change with $n$, so $\pi_{\min}^{(n)}$ can go to zero at some rate.
%In Appendix \ref{appendix:BOLS} we introduce an even weaker version of clipping that is sufficient for Theorem \ref{thm:bols}, which we do not introduce here for expositional simplicity.

\begin{mydef}
\label{def:clipping}
%Let $\pi_{\min}^{(n)}, \pi_{\max}^{(n)}$ be non-random bounded sequences with $0 < \pi_{\min}^{(n)} \leq \pi_{\max}^{(n)} < 1$.
%A \bo{fixed} clipping rate means that for all $t \in [1 \colon T]$, $\pi_t^{(n)}$ satisfies \eqref{eqn:clipping} for $\pi_{\min}^{(n)}, \pi_{\max}^{(n)}$ constant for all $n$:
A clipping constraint with rate $f(n)$ means that $\pi_t^{(n)}$ satisfies the following:
\begin{equation}
\lim_{n \to \infty} \PP \big( \pi_t^{(n)} \in [ f(n), 1-f(n) ] \big) = 1
\label{eqn:clipping}
\end{equation}
%A clipping rate \bo{dominated} by positive function $f(n)$ where $f(n) \to 0$ means that $\pi_t^{(n)}$ satisfies \eqref{eqn:clipping} for a sequences $\pi_{\min}^{(n)}, \pi_{\max}^{(n)}$ such that $\frac{ \pi_{\min}^{(n)} }{ f(n) } \to \infty$ and $\frac{ 1 - \pi_{\max}^{(n)} }{ f(n) } \to \infty$.
\end{mydef}

%\begin{mydef}
%\label{def:clipping}
%Let $\pi_{\min}^{(n)}$ be a non-random bounded sequence with $\pi_{\min}^{(n)} \in [0, 0.5]$.
%A \bo{fixed} clipping rate means that for all $t \in [1 \colon T]$, $\pi_t^{(n)}$ satisfies \eqref{eqn:clipping} for $\pi_{\min}^{(n)}$ constant for all $n$:
%\begin{equation}
%\lim_{n \to \infty} \PP \bigg( \forall m \geq n, \pi_t^{(m)} \in [ \pi_{\min}^{(m)}, 1 - \pi_{\min}^{(m)} ] \bigg) = 1
%\label{eqn:clipping}
%\end{equation}
%A clipping rate dominated by positive function $f(n)$ where $f(n) \to 0$ means that $\pi_t^{(n)}$ satisfies \eqref{eqn:clipping} for a sequence $\pi_{\min}^{(n)}$ such that $\frac{ \pi_{\min}^{(n)} }{ f(n) } \to \infty$.
%\end{mydef}

\section{Asymptotic Distribution of the Ordinary Least Squares Estimator}
Suppose we are in the stationary case, and we would like to estimate $\bs{\beta}$.
Consider the OLS estimator:
%\begin{equation*}
$\bs{ \betahat }^{\OLS} = ( \under{\boldX}^\top \under{\boldX} )^{-1} \under{\boldX}^\top \boldR$,
%\end{equation*}
where $\under{\boldX} := [ \boldX_{1, 1}, .., \boldX_{1, n}, .., \boldX_{T,1}, .., \boldX_{T,n}]^\top \in \real^{nT \by 2}$
and $\boldR := [ R_{1,1}, .., R_{1,n}, .., R_{T,1}, .., R_{T,n} ]^\top \in \real^{nT}$.
Note that $\under{\boldX}^\top \under{\boldX} = \sum_{t=1}^T \sum_{i=1}^n \boldX_{t,i} \boldX_{t,i}^\top$.

\subsection{Conditions for Asymptotically Normality of the OLS estimator}
\label{section:olsnormal} % (no adaptive sampling)
If $(\boldX_{t,i}, \epsilon_{t,i})$ are i.i.d., $\E[ \epsilon_{t,i} ] = 0$, $\E[ \epsilon_{t,i}^2] = \sigma^2$, and the first two moments of $\boldX_{t,i}$ exist, a classical result from statistics \cite{amemiya1985advanced} is that the OLS estimator is asymptotically normal, i.e., as $n \to \infty$,
%\label{eqn:clt}
\begin{equation*}
( \under{\boldX}^\top \under{\boldX} )^{1/2} ( \bs{ \betahat }^{\OLS} - \bs{\beta} ) \Dto \N( \bs{0}, \sigma^2 \under{\bs{I}}_p ).
\end{equation*}
\cite{lai1982least} generalize this result by proving that the OLS estimator is still asymptotically normal in the adaptive sampling case when $\under{\boldX}^\top \under{\boldX}$ satisfies a certain stability condition.
To show that a similar result holds for the batched setting, we generalize the CLT of \cite{lai1982least} to triangular arrays (required since the distribution of our random variables vary as the batch size, $n$, changes), as stated in Theorem \ref{thm:triangularCLT}.
%Note, we must consider triangular array asymptotics since the distribution of our random variables vary as the batch size, $n$, changes.
%For example, the distribution of the actions one chooses for the $2^{\TN{nd}}$ batch, $\{ A_{2,i}^{(n)} \}_{i=1}^n$, may change if one has observed $n=10$ vs. $n=100$ samples $( A_{1,i}^{(n)}, R_{1,i}^{(n)} )_{i=1}^n$ in the first batch.

\begin{condition}[Moments]
\label{cond:moments}
For all $t, n, i$, $\E [ \epsilon_{t, i}^2 \big| \G_{t-1}^{(n)} ] = \sigma^2$ and
$\E \big[ \epsilon_{t,i}^4 \big| \G_{t-1}^{(n)} \big] < M < \infty$.
%For all $t, n, i$, $\E [ (\epsilon_{t, i}^{(n)} )^2 \big| \G_{t-1}^{(n)} ] = \sigma^2$ and
%$\E \big[ ( \epsilon_{t,i}^{(n)} )^4 \big| \G_{t-1}^{(n)} \big] < M < \infty$.
\end{condition}

\begin{condition}[Stability]
\label{cond:banditstability}
For some non-random sequence of scalars $\{ a_i \}_{i=1}^\infty$, as $n \to \infty$, \\
%\begin{equation*}
\centerline{$a_n \cdot \frac{1}{nT} \sum_{t=1}^T \sum_{i=1}^n A_{t,i} \Pto 1$}
%\end{equation*}
\end{condition}

%\begin{theorem}[Triangular array version of Lai \& Wei (1982), Theorem 3]
\begin{theorem}[Triangular array version \cite{lai1982least}, Theorem 3]
\label{thm:triangularCLT}
Assuming Conditions \ref{cond:moments} and \ref{cond:banditstability}, as $n \to \infty$,
%\begin{equation*}
\centerline{$\big( \under{\boldX}^{ \top} \under{\boldX} \big)^{1/2} ( \bs{\betahat}^{\OLS} - \bs{\beta}) 
\Dto \N( 0, \sigma^2 \under{\bo{I}}_p)$.}
%\end{equation*}
%\begin{equation*}
%\big( \under{\boldX}^{(n), \top} \under{\boldX}^{(n)} \big)^{1/2} ( \bs{\betahat}^{\OLS} - \bs{\beta}) 
%\Dto \N( 0, \sigma^2 \under{\bo{I}}_p)
%\end{equation*}
\end{theorem}
Note that in the bandit setting, Condition \ref{cond:banditstability} means that prior to running the experiment, the asymptotic rate at which arms will be selected is predictable. 
%Intuitively, this means the asymptotic rates arms are selected are the same across multiple identical bandit experiments with different random seeds.
We will show that Condition \ref{cond:banditstability} is in a sense necessary for the asymptotic normality of OLS. %, i.e., equation \eqref{eqn:OLSCLT}.
In Corollary \ref{corollary:sufficientcond} below we state that Conditions \ref{cond:moments} and \ref{cond:condiid}, and a non-zero margin are sufficient for stability Condition \ref{cond:banditstability}. 
Later, we will show that when the margin is zero, Condition \ref{cond:banditstability} does not hold for many common bandit algorithms and prove that this leads the OLS estimator to be asymptotically non-normal.

\begin{condition}[Conditionally i.i.d. actions] 
\label{cond:condiid}
For each $t \in [1 \colon T]$,
$\{ A_{t,i} \}_{i=1}^n \iidsim \TN{Bernoulli} ( \pi_t^{(n)} )$ i.i.d. over $i \in [1 \colon n]$ conditional on $H_{t-1}^{(n)}$.
%For each $t \in [1 \colon T]$,
%$A_{t,i}^{(n)} \iidsim \TN{Bernoulli} ( \pi_t^{(n)} )$ i.i.d. over $i \in [1 \colon n]$ conditionally on $H_{t-1}^{(n)}$.
\end{condition}

\smallskip
\begin{corollary}[Sufficient conditions for Theorem \ref{thm:triangularCLT}]
If Conditions \ref{cond:moments} and \ref{cond:condiid} hold, and \bo{the margin is non-zero}, data collected in batches using $\epsilon$-greedy, Thompson Sampling, or UCB with clipping constraint with $f(n) = c~$ for some $0 < c \leq \frac{1}{2}$ (see Definition \ref{def:clipping}) satisfy Theorem \ref{thm:triangularCLT} conditions. 
\label{corollary:sufficientcond}
\end{corollary}

\begin{figure}[t]
\begin{center}
\centerline{ \includegraphics[width=0.49 \textwidth]{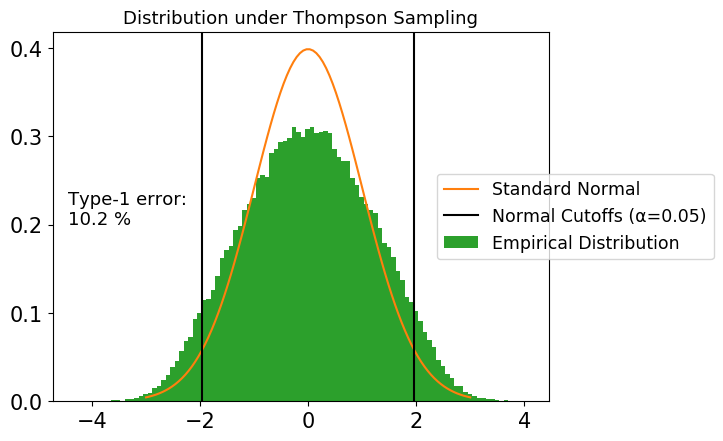} \includegraphics[width=0.49\textwidth]{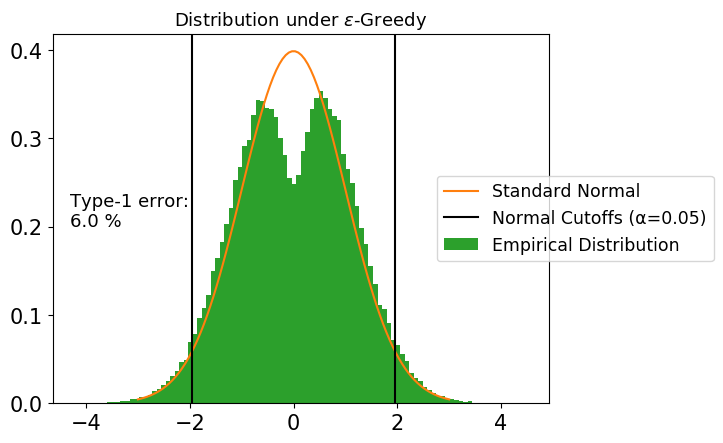}}
\caption{Empirical distribution of the Z-statistic ($\sigma^2$ is known) of the OLS estimator for the margin.
All simulations are with no margin ($\beta_1 = \beta_0 = 0$); $\N(0, 1)$ rewards; $T=25$; and $n=100$. 
For $\epsilon$-greedy, $\epsilon = 0.1$.}
\label{fig:nonnorm}
\end{center}
\vspace{-5mm}
\end{figure}

\subsection{Asymptotic Non-Normality under No Margin}
\label{section:olsnonnormal}
We prove the conjecture of \cite{deshpande} that when the margin is zero, the OLS estimator is asymptotically non-normal under common bandit algorithms, including Thompson Sampling, $\epsilon$-greedy, and UCB. % (Theorem \ref{thm:nonnormTS})
Thus as seen in Figure~\ref{fig:nonnorm}, assuming the OLS estimator is approximately Normal on bandit data can lead to inflated Type-1 error, even asymptotically.
%As seen in Figure~\ref{fig:nonnorm}, the Z-statistic of the OLS estimator of the margin can have fat tails, which can lead to poor control of Type-1 error in hypothesis testing.  
The asymptotic non-normality of OLS occurs when the margin is zero because when there is no unique optimal arm, $\pi_t^{(n)}$ does not concentrate as $n \to \infty$ (Appendix \ref{appendix:nonnormality}).
% note that the non-concentration of $\pi_t^{(n)}$ violates Condition \ref{cond:banditstability} \sam{necessary?} for the previously stated CLT for the OLS estimator, Theorem \ref{thm:triangularCLT}. %; see Appendix \ref{appendix:nonnormality} for details.

%\begin{comment}
%As illustrated in Figure~\ref{fig:nonnorm}, the Z-statistics for the OLS estimator of the margin can have fatter tails than a Normal distribution, which can lead to poor performance of  confidence intervals and/or the control of Type-1 error in hypothesis testing.  
%%%%%%%%%%%5
%the noise variance $\sigma^2$ is known to be $1$ %under Thompson Sampling
We state the asymptotic non-normality result for Thompson Sampling in Theorem \ref{thm:nonnormTS}; see Appendix \ref{appendix:nonnormality} for the proof and similar results for $\epsilon$-greedy and UCB. 
It is sufficient to prove asymptotic non-normality for $T=2$. % and the margin is the same for each batch with fixed clipping. 
Note, $\hat{\Delta}^{\OLS}$ is the difference in the sample means for each arm, so $\hat{\Delta}^{\OLS} = \betahat_1^{\OLS} - \betahat_0^{\OLS}$.
%The OLS estimator of $\Delta$ is simply the difference in the sample means for each arm, i.e., $\hat{\Delta}^{\OLS} = \betahat_1^{\OLS} - \betahat_0^{\OLS}$.  %=\beta_ 1 - \beta_0
The Z-statistic of $\hat{\Delta}^{\OLS}$, which is asymptotically normal under i.i.d. sampling, is as follows:  %(z-statistic under the null $H_0 \colon \Delta =0$) 
%(assuming known noise variance $\sigma^2$) 
\begin{equation}
\label{eqn:zstat}
\sqrt{ \frac{ \sum_{t=1}^2 \sum_{i=1}^n A_{t,i} ) ( \sum_{t=1}^2 \sum_{i=1}^n 1 - A_{t,i} ) }{ 2 \sigma^2 n } } \big( \hat{\Delta}^{\OLS} - \Delta \big).
\end{equation}
%\sqrt{ \frac{ (N_{1,1} + N_{2,1}) (N_{1,0} + N_{2,0}) }{ 2 \sigma^2 n } } \big( \hat{\Delta}^{\OLS} - \Delta \big).
%\begin{theorem}[Non-uniform convergence of the OLS estimator of the margin for Thompson Sampling]
\begin{theorem}[Asymptotic non-normality of OLS estimator under zero margin for Thompson Sampling]
\label{thm:nonnormTS}
Let $T=2$ and $\pi_1^{(n)} = \frac{1}{2}$.
%Let $T=2$ and $\pi_1^{(n)} = \frac{1}{2}$.% for all $n$.
%We put independent standard normal priors on the means of each arm, $\tilde{\beta}_0, \tilde{\beta}_1 \iidsim \N(0, 1)$.
%We put independent standard normal priors on the means of each arm $\tilde{\beta}_0 \sim \N(0, 1)$ and $\tilde{\beta}_1 \sim \N(0, 1)$.
%The algorithm assumes noise variance $\sigma^2 = 1$. 
If $\epsilon_{t,i} \iidsim \N(0, 1)$, we have independent normal priors on arm means $\tilde{\beta}_0, \tilde{\beta}_1 \iidsim \N(0, 1)$, and $\pi_2^{(n)} = \pi_{\min} \vee [ (1-\pi_{\max}) \wedge \PP ( \tilde{\beta}_1 > \tilde{\beta}_0 \big| H_1^{(n)} ) ]$ for constants $\pi_{\min}, \pi_{\max}$ with $0 < \pi_{\min} \leq \pi_{\max} < 1$,
then \eqref{eqn:zstat} is asymptotically \bo{not} normal when the margin is zero.
%the OLS estimator Z-statistic
 %the Z-statistic of 
%, i.e., %when the margin is zero there exists an $x \in \real$ such that
%\begin{equation*}
%$\lim_{n \to \infty} \sup_{x \in \real} |F_n(x) - \Phi(x) | \not= 0$,
%\end{equation*}
%where $\Phi$ is the standard Normal distribution CDF and $F_n$ is the CDF of the Z-statistic of the OLS estimator, .
\end{theorem}
%\end{comment}
%then the z-statistic for the OLS estimator is asymptotically normal when  if and only if the margin is zero.
%This means that if $\Delta \neq 0$, for all $x \in \real$,
%\begin{equation*}
%\lim_{n \to \infty} |F_n(x) - \Phi(x) | = 0;
%\end{equation*}
%otherwise, if $\Delta = 0$, there exists some $x \in \real$ such that
%\begin{equation*}
%\lim_{n \to \infty} |F_n(x) - \Phi(x) | \not= 0.
%\end{equation*}
%where $\Phi$ is the CDF of a standard Normal distribution and $F_n$ is the CDF of the z-statistic for the OLS estimator, \eqref{eqn:zstat}.s
\begin{figure}[t]
\begin{center}
\centerline{\includegraphics[width=0.45\columnwidth]{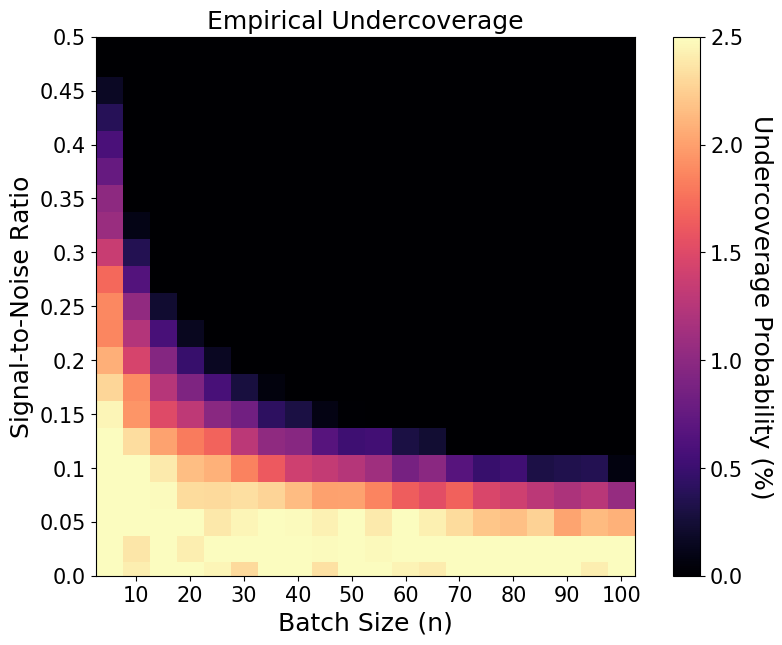}}
\caption{Empirical undercoverage probabilities (coverage probability below $95\%$) of confidence intervals using on a normal approximation for the OLS estimator. %the OLS estimator and constructed using the quantiles of a standard Normal. 
We use Thompson Sampling with $\N(0,1)$ priors, a clipping constraint of $0.05 \leq \pi_t^{(n)} \leq 0.95$, $\N(0,1)$ rewards, $T=25$, and known $\sigma^2$.
Standard errors are $< 0.001$.} %for the estimates used in the plot above
\label{fig:nonuniform}
\end{center}
\vspace{-10mm}
\end{figure}
%Note that when there is no margin, there is no unique optimal, regret-minimizing policy.
%In the sequential treatment regime literature, sequential decision making problems that do not have unique optimal policies are called ``exceptional laws"  \cite{luedtke2016statistical, robins2004optimal, robins2014discussion}. 
Since the OLS estimator is asymptotically normal when $\Delta \neq 0$ (Corollary \ref{corollary:sufficientcond}) and asymptotically \textit{not} Normal when $\Delta = 0$, the OLS estimator does not converge \textit{uniformly} on data collected under standard bandit algorithms. %Theorem \ref{thm:nonnormTS} %(Theorem \ref{thm:nonnormTS}; Appendix \ref{appendix:nonnormality})
%By Corollary \ref{corollary:sufficientcond}, we know the OLS estimator is asymptotically normal when $\Delta \neq 0$.
%And since Theorem \ref{thm:nonnormTS} shows that the OLS estimator is asymptotically \textit{not} normal when $\Delta = 0$, we know that the OLS estimator does not converge \textit{uniformly} on adaptively-collected data.
The non-uniform convergence of the OLS estimator precludes us from using a normal approximation to perform hypothesis testing and construct confidence intervals (see \cite{kasy2019uniformity}).
%See \cite{kasy2019uniformity} for a review of the problems with asymptotic approximations under non-uniform convergence. 
%Since the OLS estimator will be asymptotically normal when $\Delta \neq 0$ and asymptotically non-normal when $\Delta = 0$, we get that the OLS estimator does not converge \textit{uniformly} on adaptively-collected data (see \cite{kasy2019uniformity} for a review of non-uniform convergence and asymptotic approximations). 
%Note, this non-uniform convergence result means that bootstrap techniques are not applicable in this setting because the validity of bootstrap relies on uniform convergence (typically satisfied by Asymptotic Normality) \cite{romano2012uniform}.
%\todo[inline]{exceptional law discussion}
In real-world applications, there is rarely exactly zero margin. However, the non-uniform convergence of the OLS estimator at zero margin is still practically important because
%In real world applications, there is rarely exactly zero margin. 
%However, the discontinuity in the asymptotic distribution of the OLS estimator at zero margin is still practically important. 
%For constructing confidence intervals, we use the asymptotic distribution to approximate the finite-sample distribution. %hypothesis testing and 
the asymptotic distribution of the OLS estimator when the margin is zero is indicative of the finite-sample distribution when the margin is statistically difficult to differentiate from zero, i.e., when the signal-to-noise ratio, $\frac{|\Delta|}{ \sigma }$, is low. 
Figure \ref{fig:nonuniform} shows that \textit{even when the margin is non-zero, when the signal-to-noise ratio is low, confidence intervals constructed using a normal approximation have coverage probabilities below the nominal level. %cutoffs based on the normal distribution have coverage probabilities below the nominal level.
Moreover, for any batch size $n$ and noise variance $\sigma^2$, there exists a non-zero margin size with a finite-sample distribution that is poorly approximated by a normal distribution.}

\section{Batched OLS Estimator}

\subsection{Batched OLS Estimator for Multi-Arm Bandits}
We now introduce the Batched OLS (BOLS) estimator that is asymptotically normal under a large class of bandit algorithms, even when the margin is zero.
Instead of computing the OLS estimator on all the data, we compute the OLS estimator for each batch and normalize it by the variance estimated from that batch. 
%Instead of computing the OLS estimator on the data from all batches together, we compute the OLS estimator from each batch and normalize it by the variance estimated from that batch. 
For each $t \in [1 \colon T]$, the BOLS estimator of the margin $\Delta_t := \beta_{t,1}- \beta_{t,0}$ is:
\begin{equation*}
\hat{\Delta}_t^{\TN{BOLS}} 
= \frac{ \sum_{i=1}^{n} (1-A_{t,i} ) R_{t,i} }{ \sum_{i=1}^{n} 1 - A_{t,i} } - \frac{ \sum_{i=1}^{n} A_{t,i} R_{t,i} }{ \sum_{i=1}^{n} A_{t,i} }.
\end{equation*}
%\begin{equation*}
%\hat{\Delta}_t^{\TN{BOLS}} 
%= \frac{ \sum_{i=1}^{n} (1-A_{t,i} ) R_{t,i} }{ N_{t,0} } - \frac{ \sum_{i=1}^{n} A_{t,i} R_{t,i} }{ N_{t,1} }.
%\end{equation*}
%Note that BOLS bears resemblance to estimator of \cite{luedtke2016statistical}  for the expected reward under an optimal policy, since it is a weighted combination of a statistic computed on batches.
%However, the estimator of \cite{luedtke2016statistical} assumes an observational study with i.i.d. action and reward tuples (see their section 2), which means their results do not apply to bandit data.
%In contrast, we do not assume i.i.d. action and reward tuples because for bandit data, the rewards in a previous batch can affect which actions are chosen in future batches. 
%%%%%%%%%%%%%%%%
\begin{theorem}[Asymptotic Normality of Batched OLS estimator for multi-arm bandits]
\label{thm:bols} Assuming Conditions \ref{cond:moments} (moments) and \ref{cond:condiid} (conditionally i.i.d. actions), and a clipping rate $f(n) = \frac{1}{n^\alpha}$~ for some $0 \leq \alpha < 1$ (see Definition \ref{def:clipping}), %\footnote{It is straightforward to show that these results hold in the case that batches are different sizes (for non-adaptively chosen batch sizes) as the size of the smallest batch goes to infinity.}
\begin{equation*}
\begin{bmatrix} 
\sqrt{ \frac{ ( \sum_{i=1}^{n} 1 - A_{1,i} ) ( \sum_{i=1}^{n} A_{1,i} ) }{ n } } ( \hat{\Delta}_1^{\TN{BOLS}} - \Delta_1 )  \\
\sqrt{ \frac{ ( \sum_{i=1}^{n} 1 - A_{2,i} ) ( \sum_{i=1}^{n} A_{2,i} ) }{ n } } ( \hat{\Delta}_2^{\TN{BOLS}} - \Delta_2 )   \\
\vdots \\
\sqrt{ \frac{ ( \sum_{i=1}^{n} 1 - A_{T,i} ) ( \sum_{i=1}^{n} A_{T,i} ) }{ n } } ( \hat{\Delta}_T^{\TN{BOLS}} - \Delta_T )   \\
 \end{bmatrix}
\Dto \N(0, \sigma^2 \under{\bo{I}}_T ) \vspace{-2mm}
\end{equation*}
%Thus, for any non-random $\boldc = [c_1, c_2, ..., c_T]^\top \in \real^T$ such that $|| \boldc ||_2 = 1$,
%\begin{equation*}
%\sum_{t=1}^T c_t \sqrt{ \frac{ N_{t,0} N_{t,1} }{ n } } ( \hat{\Delta}_t^{\TN{BOLS}} - \Delta_t ) \Dto \N(0, \sigma^2).
%\end{equation*}
\end{theorem}
It is straightforward to generalize Theorem \ref{thm:bols} to the case that batches are different sizes but the size of

the smallest batch goes to infinity and  the batch size is independent of the history.
%Note that our asymptotic results will hold even if the number of arm pulls in each batch vary, 

%%%%%%%%%%%%%%%%
%The simplest case is to let $c_t = \frac{1}{ \sqrt{T} }$ for all $t$. 
By Theorem \ref{thm:bols}, for the stationary margin case, we can test $H_0: \Delta = c$ vs. $H_1: \Delta \neq c$ with the following statistic, which is asymptotically normal under the null:
\begin{equation}
\label{eqn:testStat}
	\frac{1}{ \sqrt{T} } \sum_{t=1}^T \sqrt{ \frac{ ( \sum_{i=1}^{n} 1 - A_{t,i} ) ( \sum_{i=1}^{n} A_{t,i} ) }{ n \sigma^2 } } ( \hat{\Delta}_t^{\TN{BOLS}} - c ).
\end{equation}
This type of test statistic---a weighted combination of asymptotically independent normals---a special case of the inverse normal p-value combination test, has been used in simple settings in which the studies (e.g., batches) are independent (e.g., when conducting meta-analyses across multiple studies) [26]. Here the ability to use this type of test statistic is novel since, due to the bandit algorithm, the batches are {\it not} independent. The work here demonstrates asymptotic independence and thus for large $n$ the Z-statistics from each batch should be approximately independently distributed.

The key to proving asymptotic normality for BOLS is that the following ratio converges in probability to one: 
%\begin{equation*}
%$\frac{ N_{t,1} N_{t,0} }{ n } \frac{1}{ n \pi_t^{(n)} (1 - \pi_t^{(n)})} \Pto 1$.
$\frac{ ( \sum_{i=1}^{n} 1 - A_{t,i} ) ( \sum_{i=1}^{n} A_{t,i} ) }{ n } \frac{1}{ n \pi_t^{(n)} (1 - \pi_t^{(n)})} \Pto 1$.
%\end{equation*}
Since $\pi_t^{(n)} \in \G_{t-1}^{(n)}$, $\frac{1}{ n \pi_t^{(n)} (1 - \pi_t^{(n)})}$ is a constant given $\G_{t-1}^{(n)}$.
Thus, even if $\pi_t^{(n)}$ does not concentrate, we are still able to apply the martingale CLT \cite{dvoretzky1972asymptotic} to prove asymptotic normality.
See Appendix \ref{appendix:triangularCLT} for more details.

\subsection{Batched OLS Estimator for Contextual Bandits}
\label{section:contextBOLS}
For contextual $K$-arm bandits, %the parameters of interest are $\bs{\beta}_t = ( \bs{\beta}_{t,0}, \bs{\beta}_{t,1}, ..., \bs{\beta}_{t,K-1} )^\top \in \real^{Kd}$. 
for any two arms $x, y \in [0 \colon K-1]$, we can estimate the margin between them $\bs{\Delta}_{t, x-y} := \bs{\beta}_{t,x} - \bs{\beta}_{t,y} \in \real^d$. % for all $t \in [1 \colon T]$. 
In each batch, we observe context vectors $\{ \boldC_{t,i} \}_{i=1}^n$ for $\boldC_{t,i} \in \real^d$.
%We can deterministically set the first element of $\boldC_{t,i}^{(n)}$ to $1$ to allow for a non-zero intercept.
%For our new definition of the history, 
%$H_{t-1}^{(n)} := \{ \boldC_{t,i}, A_{t,i}, R_{t,i} \colon i \in [1 \colon n], t \in [1 \colon T] \}$ 
We redefine the history $H_{t-1}^{(n)} := \{ \boldC_{t',i}, A_{t',i}, R_{t',i} \}_{i=1,t'=1}^{i=1,t'=t-1}$ and define the filtration $\F_t^{(n)} := \sigma \big( H_{t-1}^{(n)} \cup \{ A_{t,i}, \boldC_{t,i} \}_{i=1}^n \big)$.
%Our policy $\bs{\pi}_t^{(n)} \in \real^K$ is now a vector an actions $A_{t,i}^{(n)} \in [0 \colon K-1]$ and $A_{t,i}^{(n)} \sim \TN{Categorial} \big( \bs{\pi}_t^{(n)} (\boldC_{t,i}^{(n)}) \big) - 1$.
%Moreover, $\bs{\pi}_t^{(n)}$ as it is not only a function of $H_{t-1}^{(n)}$, but also of context $\boldC_{t,i}^{(n)}$.
%Our policy $\bs{\pi}_t^{(n)} \colon \real^d \to \real^K$ is a function that takes a context vector $\boldC_{t,i}^{(n)}$ and outputs a vector representing the probability that any action is chosen, i.e., an actions $A_{t,i}^{(n)} \in [0 \colon K-1]$ and $A_{t,i}^{(n)} \sim \TN{Categorial} \big( \bs{\pi}_t^{(n)} (\boldC_{t,i}^{(n)}) \big) - 1$.
%We let $\pi_{t,i}^{(n)} (k) := \PP( A_{t,i}^{(n)} = k | \boldC_{t,i}^{(n)}, H_{t-1}^{(n)} )$.
%Our policy $\bs{\pi}_{t,i}^{(n)} \in \real^K$ is now a vector representing the probability that any action is chosen, i.e., an actions $A_{t,i}^{(n)} \in [0 \colon K-1]$ and $A_{t,i}^{(n)} \sim \TN{Categorial}( \bs{\pi}_{t,i}^{(n)} ) - 1$.
The action selection probabilities $\bs{\pi}_t^{(n)}$ are now functions of the context, 
%so $\bs{\pi}_t^{(n)} \colon \real^d \to [0, 1]^K$ and 
so $\bs{\pi}_t^{(n)} (\boldC_{t,i}) \in [0,1]^K$ is a vector where the $k^{\TN{th}}$ dimension equals $\PP( A_{t,i} = k | \HH_{t-1}^{(n)}, \boldC_{t,i})$. %of $\bs{\pi}_t^{(n)} (\boldC_{t,i})$
% [ \PP( A_{t,i} = 0 | \HH_{t-1}^{(n)}, \boldC_{t,i}), \PP( A_{t,i} = 1 | \HH_{t-1}^{(n)}, \boldC_{t,i}), ..., \PP( A_{t,i} = K-1 | \HH_{t-1}^{(n)}, \boldC_{t,i}) ]$.
%We define contextual bandit algorithms are functions $\{ \MC{A}_t \}_{t=1}^T$ such that $\MC{A}_t ( H_{t-1}^{(n)}, \boldC_{t,i} ) =: \bs{\pi}_{t,i}^{(n)} \in [0, 1]^K$.  \sam{if we need to save space, I don't think we need to introduce $\MC{A}_t$ as we could get away with just using the $\pi_t^{(n)} := \PP( A_{t,i} = 1 | \HH_{t-1}^{(n)}, \boldC_{t,i})$ notation}
%Note, $\bs{\pi}_{t,i}^{(n)}$ is indexed by $i$ as it depends on the context $\boldC_{t,i}$.
%Policy $\bs{\pi}_{t,i}^{(n)} \in \real^K$ is a vector of action selection probabilities, i.e., action $A_{t,i} \in [0 \colon K-1]$ and $A_{t,i} \sim \TN{Categorial}( \bs{\pi}_{t,i}^{(n)} ) - 1$. 
We assume the following conditional mean model of the reward:
%\begin{equation*}
$\E \big[ R_{t,i} \big| \F_{t-1}^{(n)} \big] 
= \sum_{k=0}^{K-1} \II_{ (A_{t,i} = k) } \boldC_{t,i}^\top \bs{\beta}_{t,k} $
%\end{equation*}
%\begin{equation*}
%\E \big[ R_{t,i}^{(n)} \big| \F_{t-1}^{(n)} \big] 
%= A_{t,i}^{(n)} \boldC_{t,i}^\top \bs{\beta}_{t,1} + (1 - A_{t,i}^{(n)} ) \boldC_{t,i}^\top \bs{\beta}_{t,0}
%\end{equation*}
and let
$\epsilon_{t,i} := R_{t,i} - \sum_{k=0}^{K-1} \II_{ (A_{t,i}  = k) } \boldC_{t,i}^\top \bs{\beta}_{t,k} $.
%We let $\pi_{t,i}^{(n)} (k) := \PP( A_{t,i}^{(n)} = k | \boldC_{t,i}^{(n)}, H_{t-1}^{(n)} )$.
%$\epsilon_{t,i}^{(n)} := R_{t,i}^{(n)} - A_{t,i}^{(n)} \boldC_{t,i}^\top \bs{\beta}_{t,1} - (1 - A_{t,i}^{(n)} ) \boldC_{t,i}^\top \bs{\beta}_{t,0}$.
%is now a vector representing the probability that any action is chosen, 
%Note, $\bs{\pi}_t^{(n)}$ is now a function of the context $\boldC_{t,i}^{(n)}$.
%Note that $\pi_{t,k}^{(n)} \big( \boldC_{t,i}^{(n)} \big) = \PP( A_{t,i}^{(n)} = k | \boldC_{t,i}^{(n)}, H_{t-1}^{(n)} )$.
%Moreover, $\bs{\pi}_{t,i}^{(n)}$ is now indexed by $i$ as it is not only a function of $H_{t-1}^{(n)}$, but also of context $\boldC_{t,i}^{(n)}$.

\begin{condition}[Conditionally i.i.d. contexts]
\label{cond:iidcontext}
For each $t$, $\boldC_{t,1}, \boldC_{t,2}, ..., \boldC_{t,n}$ are i.i.d. and its first two moments, $\bs{\mu}_t, \underline{ \bs{ \Sigma } }_t$, are non-random given $H_{t-1}^{(n)}$, i.e., $\bs{\mu}_t, \underline{ \bs{ \Sigma } }_t \in \sigma(H_{t-1}^{(n)})$.
\end{condition}

\begin{condition}[Bounded context]
\label{cond:boundedcontext}
$|| \boldC_{t,i} ||_{\max} \leq u$ for all $i, t, n$ for some constant $u$.  %\sam{should this be $\boldC_{t,i}$ instead?}
Also, the minimum eigenvalue of $\underline{ \bs{ \Sigma } }_t$ is lower bounded, i.e., $\lambda_{\min} ( \underline{ \bs{ \Sigma } }_t ) > l > 0$.
\end{condition}

%\todo[inline]{rate stuff!}
%\begin{mydef}
%\label{def:condclipping}
%%Recall that $\MC{A}_t( H_{t-1}^{(n)}, \boldC_{t,i}^{(n)}) = \bs{\pi}_{t,i}^{(n)} \in [0,1]^K$.
%A conditional clipping constraint with rate $f(n)$ means that the action selection probabilities $\bs{\pi}_{t,i}^{(n)} := \MC{A}( H_{t-1}^{(n)}, \boldC_{t,i}) \in [0,1]^K$ satisfy the following: %$\pi_t^{(n)}$ satisfies the following:
%%\begin{equation*}
%%\inf_{ \boldc \in \real^d} \PP \bigg( \pi_{t,i} \in \big[ f(n), 1-f(n) \big]^K ~\bigg|~ \boldC_{t,i} = \boldc \bigg) \to 1
%%\end{equation*}
%%Let $\pi_{\min}, \pi_{\max}$ be constants with $0 < \pi_{\min} \leq \pi_{\max} < 1$. 
%%\bo{Conditional} fixed clipping means that $\bs{\pi}_t^{(n)}$ satisfies the following for all $t \in [1 \colon T]$, as $n \to \infty$,
%\begin{equation*}
%\PP \big( \forall ~\boldc \in \real^d, \MC{A}_t( H_{t-1}^{(n)}, \boldc) \in \big[ f(n), 1-f(n) \big]^K \big) \to 1
%\end{equation*}
%\end{mydef}

\begin{mydef}
\label{def:condclipping}
A conditional clipping constraint with rate $f(n)$ means that the action selection probabilities $\bs{\pi}_t^{(n)}: \real^d \to [0,1]^K$ satisfy the following: 
\begin{equation*}
\PP \big( \forall ~\boldc \in \real^d, \bs{\pi}_t^{(n)}( \boldc) \in \big[ f(n), 1-f(n) \big]^K \big) \to 1
\end{equation*}
\end{mydef}
\vspace*{-2mm}
%Conditional fixed clipping constrains the probability that any arm is sampled to be bounded away from zero and one. 
%This condition could potentially be weakened, as in the multi-arm bandit case; we leave this to future work.
For each $t \in [1 \colon T]$, we have the OLS estimator for $\bs{ \Delta }_{t, x-y}$:
%\begin{equation*}
$\bs{ \hat{\Delta} }_t^{\OLS} := \big[ \underline{\boldC}_{t,x}^{-1} + \underline{\boldC}_{t,y}^{-1} \big]^{-1} \big( \bs{\betahat}_{t,x}^{\OLS}  - \bs{\betahat}_{t,y}^{\OLS} \big)$,
%\end{equation*}
where $\underline{\boldC}_{t,k} := \sum_{i=1}^n \II_{ A_{t,i}^{(n)} = k } \boldC_{t,i} \boldC_{t,i}^\top \in \real^{d \by d}$, 
$\bs{\betahat}_{t,k}^{\OLS} =  \underline{\boldC}_{t,k}^{-1} \sum_{i=1}^n \II_{ A_{t,i}^{(n)} = k } \boldC_{t,i} R_{t,i}$. 

\begin{theorem}[Asymptotic Normality of Batched OLS estimator for contextual bandits]
\label{thm:bolscontext}
Assuming Conditions \ref{cond:moments} (moments)\footnote{Assume an analogous moment condition for the contextual bandit case, where $\G_t^{(n)}$ is replaced by $\F_t^{(n)}$.}, \ref{cond:condiid} (conditionally i.i.d. actions), \ref{cond:iidcontext}, and \ref{cond:boundedcontext}, and a conditional clipping rate $f(n) = c$~ for some $0 \leq c < \frac{1}{2}$ (see Definition \ref{def:condclipping}),
%$f(n) = \frac{1}{n^\alpha}$ for some $0 \leq \alpha < \frac{1}{2}$ (see Definition \ref{def:condclipping}),
\begin{equation*}
\begin{bmatrix} 
\big[ \underline{\boldC}^{-1} _{1,x} + \underline{\boldC}_{1,y}^{-1} \big]^{ 1/2 } ( \bs{ \hat{\Delta} }_1^{\OLS} - \bs{\Delta}_{1,x-y} )  \\
\big[ \underline{\boldC}^{-1} _{2,x} + \underline{\boldC}_{2,y}^{-1} \big]^{ 1/2 } ( \bs{ \hat{\Delta} }_2^{\OLS} - \bs{\Delta}_{2,x-y} )   \\
\vdots \\
\big[ \underline{\boldC}^{-1}_{T,x} + \underline{\boldC}_{T,y}^{-1} \big]^{ 1/2 } ( \bs{ \hat{\Delta} }_T^{\OLS} - \bs{\Delta}_{T,x-y} )   \\
 \end{bmatrix}
\Dto \N(0, \sigma^2 \under{\bo{I}}_{Td} ).
\end{equation*}
%Thus, for any non-random $\boldc = [c_1, c_2, ..., c_T]^\top \in \real^T$ such that $|| \boldc ||_2 = 1$,
%\begin{equation*}
%\sum_{t=1}^T c_t \big[ \underline{\boldC}^{-1} _{t,0} + \underline{\boldC}_{t,1}^{-1} \big]^{ 1/2 } ( \hat{ \bs{\Delta} }_t^{\TN{BOLS}} - \bs{\Delta} ) \Dto \N(\bs{0}, \sigma^2 \under{\bo{I}}_d ).
%\end{equation*}
\end{theorem}

\subsection{Batched OLS Statistic for Non-Stationary Bandits}
Many real-world problems we would like to use bandit algorithms for have non-stationary over time.
For example, in online advertising, the effectiveness of an ad may change over time due to exposure to competing ads and general societal changes that could affect perceptions of an ad.
We may believe that the expected reward for a given action may vary over time, but that the margin is constant from batch to batch. In the online advertising setting, this would mean whether one ad is always better than another is stable, but the overall effectiveness of both ads may change over time.
In this case, we can simply use the BOLS test statistic described earlier in equation \eqref{eqn:testStat} to test $H_0 \colon \Delta = 0$ vs. $H_1 \colon \Delta \neq 0$.
Note that the BOLS test statistic for the margin is robust to non-stationarity in the baseline reward without any adjustment.
Moreover, in our simulation settings we estimate the variance $\sigma^2$ separately for each batch, which allows for non-stationarity in the variance between batches as well; see Appendix \ref{appendix:simulations} for variance estimation details and see Section \ref{section:simulations} for simulation results.
Additionally, in the case that we believe that the margin itself may vary from batch to batch, the BOLS test statistic can also be used to construct confidence regions that contain the true margin $\Delta_t$ for each batch simultaneously; see Appendix \ref{appendix:nste} for details.

\section{Simulation Experiments}
\label{section:simulations}

\paragraph{Procedure}
\label{sec:simprocedure}
We focus on the two-arm bandit setting and test whether the margin is zero, specifically $H_0 \colon \Delta = 0$ vs. $H_1 \colon \Delta \neq 0$.
We perform experiments for when the noise variance $\sigma^2$ is estimated. % of the error $\epsilon_{t,i}$
We assume homoscedastic errors throughout. See Appendix \ref{appendix:noisevar} for more details about how we estimate the noise variance and more details regarding our experimental setup.
%In Figures \ref{fig:stationary} we display results for stationary bandits, in Figure \ref{fig:nonstationarybaseline} we display results for bandits with a non-stationary baseline line reward, and in Figure \ref{fig:nonstationary} we show results for non-stationary bandits.
In Figures \ref{fig:type1} and \ref{fig:power}, we display results for stationary bandits and in Figure \ref{fig:nonstationarybaseline} we show results for bandits with non-stationary baseline rewards.
See Appendix \ref{appendix:nste} for results for bandits with non-stationary margins. %We display results for general non-stationary bandits in .

%We found that several of the estimators, primarily OLS and AW-AIPW (with variance stabilizing weights), have inflated finite-sample Type-1 error. 
%Since Type-1 control is a hard constraint, %solutions with inflated Type-1 error are \textit{infeasible} solutions.
%we adjust the critical values of these estimators to ensure no Type-1 error inflation. %(see Appendix \ref{appendix:?} for details). 
%However, note that it is unfeasible to make these cutoff adjustment for real experiments (unless one found the worst case setting), as there are many nuisance parameters---like the expected rewards for each arm and the variance of the noise---which can affect these cutoff values.
%
%We found that several of the estimators, primarily OLS and AW-AIPW, have inflated finite-sample Type-1 error. 
%Since Type-1 control is a hard constraint, solutions with inflated Type-1 error are \textit{infeasible} solutions.
%We make adjustments to no Type-1 error inflation, but note that it is unfeasible to make these cutoff adjustment for real experiments (unless one found the worst case setting), as there are many nuisance parameters---like the expected rewards for each arm and the variance of the noise---which can affect these cutoff values.

In our simulations, we found that OLS and AW-AIPW have inflated Type-1 error. %several of the estimators, primarily %finite-sample 
Since Type-1 control is a hard constraint, solutions with inflated Type-1 error are \textit{infeasible} solutions.
%For the sake of comparison, i
In the power plots, we adjust the cutoffs of the estimators to ensure proper Type-1 error control;
% under the null; 
 if an estimator has inflated Type-1 error under the null, in the power simulations we use a critical value estimated using the simulations under the null.
Note that it is unfeasible to make these cutoff adjustment for real experiments (unless one found the worst case setting), as there are many nuisance parameters---like the expected rewards for each arm and the noise variance---which can affect cutoff values.
%For BOLS, we use cutoffs based on the Student-t distribution rather than the normal distribution, as it is relatively straightforward to determine the number of degrees of freedom needed in the correction; see Appendix \ref{appendix:noisevar} for details. 
%We do not make a similar correction for the other estimators we compare to because it is unclear how to determine the number of degrees of freedom that should be used.

\begin{figure*}[ht]
\centering
  \includegraphics[width=0.47\textwidth]{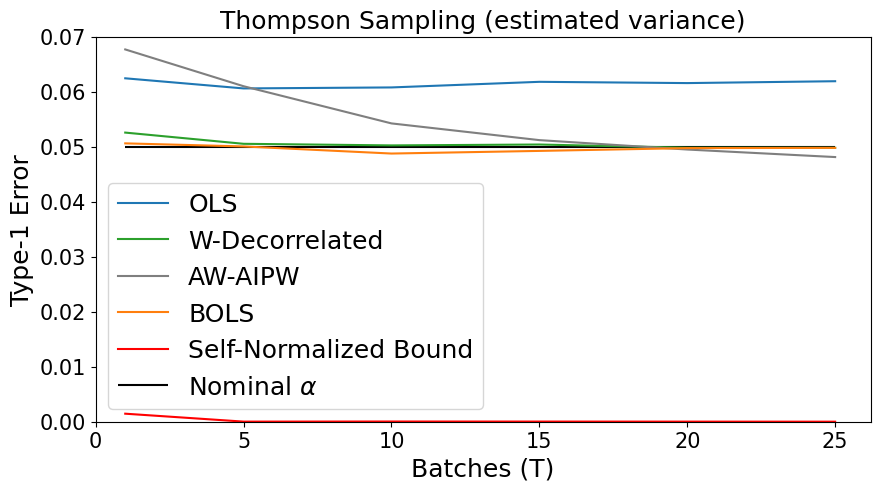}
  ~
  \includegraphics[width=0.47\textwidth]{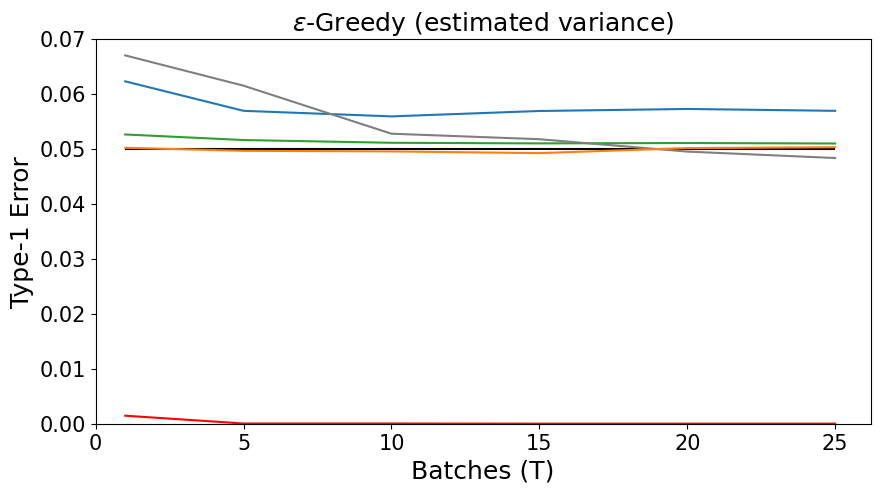}
\caption{ \bo{Stationary Setting:} Type-1 error for a two-sided test of $H_0 \colon \Delta = 0$ vs. $H_1 \colon \Delta \neq 0$ ($\alpha=0.05$). % for estimators of margin 
 We set $\beta_1 = \beta_0 = 0$, $n = 25$, and a clipping constraint of $0.1 \leq \pi_t^{(n)} \leq 0.9$.
 We use 100k Monte Carlo simulations and standard errors are $<0.001$.}
 \label{fig:type1stationary}
\end{figure*}

\begin{figure*}[ht]
\centering
  \includegraphics[width=0.47\textwidth]{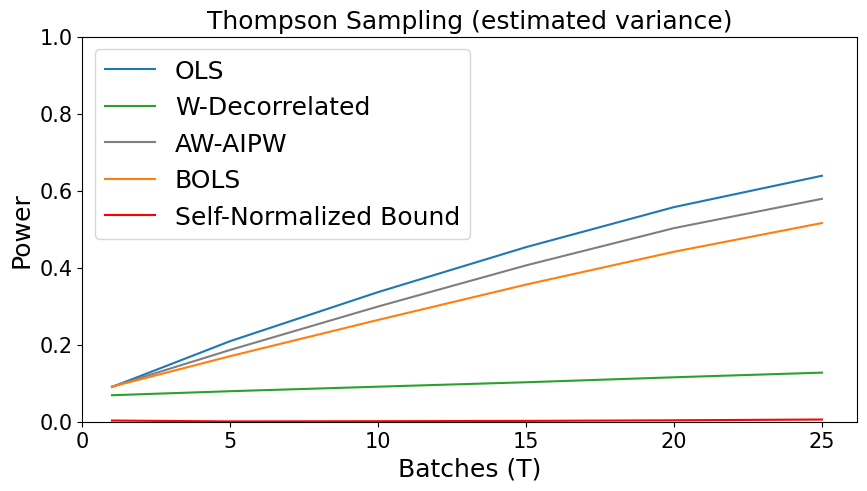}
  ~
  \includegraphics[width=0.47\textwidth]{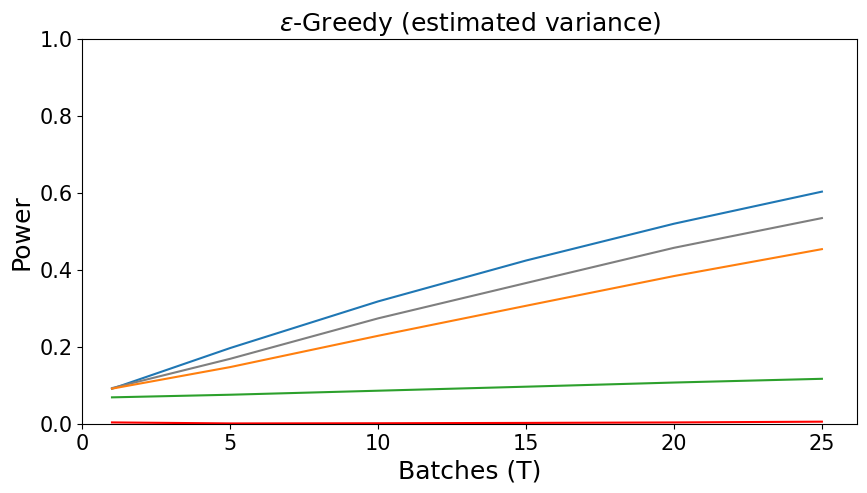}
\caption{ \bo{Stationary Setting:} Power for a two-sided test of $H_0 \colon \Delta = 0$ vs. $H_1 \colon \Delta \neq 0$ ($\alpha=0.05$).
 We set $\beta_1 = 0$, $\beta_0 = 0.25$, $n = 25$, and a clipping constraint of $0.1 \leq \pi_t^{(n)} \leq 0.9$.
 We use 100k Monte Carlo simulations and standard errors are $<0.002$.
 We account for Type-1 error inflation as described in Section \ref{sec:simprocedure}.} %Section \ref{sec:simprocedure}
 \label{fig:powerstationary}
\end{figure*}

\paragraph{Results}
Figure \ref{fig:type1stationary} shows that for small sample sizes ($nT \lesssim 300$), BOLS has more reliable Type-1 error control than AW-AIPW with variance stabilizing weights. 
After $nT \geq 500$ samples, AW-AIPW has proper Type-1 error, and by Figure \ref{fig:powerstationary} it always has slightly greater power than BOLS in the stationary setting.
The W-decorrelated estimator has reliable Type-1 error control, but very low power compared to AW-AIPW and BOLS. Finally the high probability, self-normalized martingale bound of \cite{abbasi2011improved}, which we use for hypothesis testing, has very low power compared to the asymptotic approximation statistical inference methods.

In Figure \ref{fig:nonstationarybaseline}, we display simulation results for the non-stationary baseline reward setting. 
\textit{Whereas other estimators have no Type-1 error guarantees, BOLS still has proper Type-1 error control in the non-stationary baseline reward setting.
Moreover, BOLS can have much greater power than other estimators when there is non-stationarity in the baseline reward.} %as demonstrated in Figure \ref{fig:nonstationarybaseline}, 
Overall, BOLS is favorable over other estimators in small-sample settings or when one wants to be robust to non-stationarity in the baseline reward---at the cost of losing a little power if the environment is stationary.
%Overall, it makes sense to choose BOLS over other estimators (e.g. AW-AIPW) in small-sample settings or whenever the experimenter wants to be robust to non-stationarity in the baseline reward---at the cost of losing a little power if the environment is stationary.

\clearpage
\vspace{-2mm}
\vspace{-2mm}
\begin{figure*}[ht]
\centering
  \includegraphics[width=0.47\textwidth]{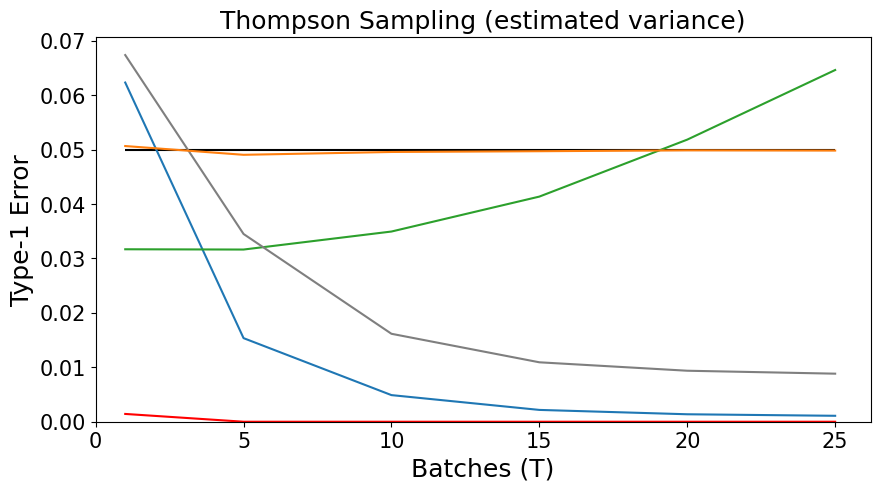}
  ~
  \includegraphics[width=0.47\textwidth]{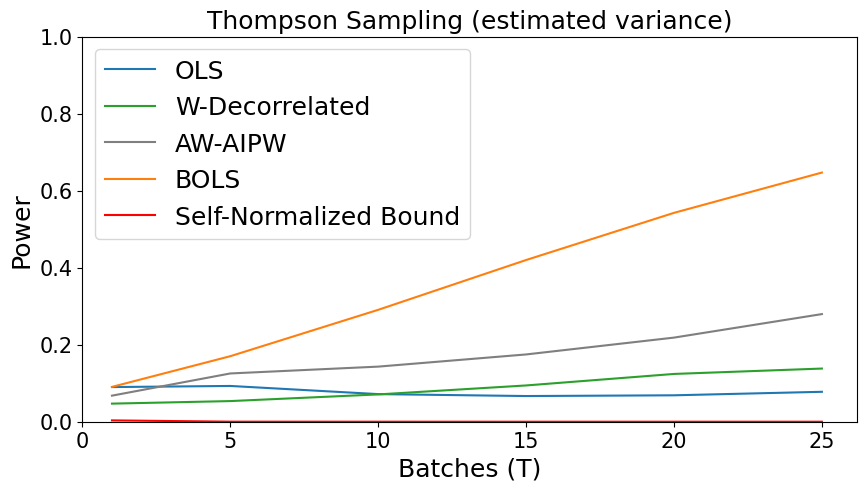}
  \vspace{-2mm}
  \includegraphics[width=0.47\textwidth]{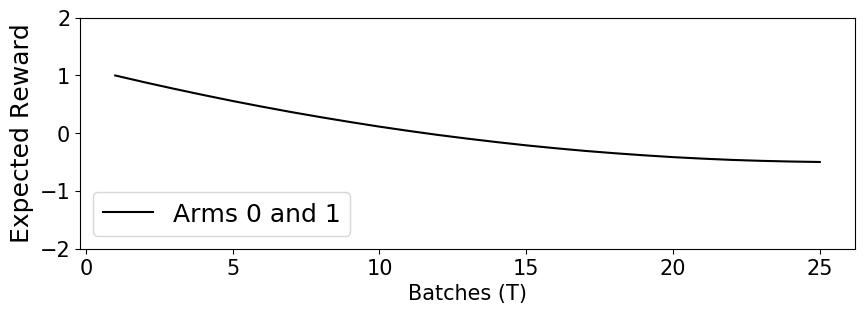}
  ~
  \includegraphics[width=0.47\textwidth]{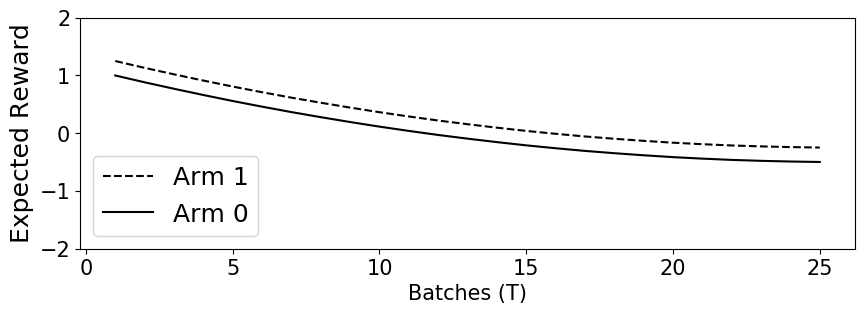} %or estimators of margin 
\caption{ \bo{Non-stationary baseline reward setting:} Type-1 error (upper left) and power (upper right) for a two-sided test of $H_0 \colon \Delta = 0$ vs. $H_1 \colon \Delta \neq 0$ ($\alpha=0.05$).
 In the lower two plots we plot the expected rewards for each arm; note the margin is constant across batches. %are plotted in the lower two plots
% The expected rewards for each arm over batches are plotted in the lower two plots; note that the margin is constant across batches.
 We use $n=25$ and a clipping constraint of $0.1 \leq \pi_t^{(n)} \leq 0.9$.
 We use 100k Monte Carlo simulations and standard errors are $<0.002$. }
% We adjust critical values to account for Type-1 error inflation as described in Section \ref{sec:simprocedure}.}
 \label{fig:nonstationarybaseline}
\end{figure*}
\vspace{-4mm}

\section{Discussion}
%BOLS has slightly less, yet comparable, power to the AW-AIPW estimator, but more reliable Type-1 error in small samples.
%One major advantage of BOLS though is that it is robust to non-stationarity in the baseline reward.
%Specifically, whereas other estimators have no Type-1 error guarantees, BOLS it still has proper Type-1 error control in the non-stationary baseline reward setting.
%Moreover, as demonstrated in Figure \ref{fig:nonstationarybaseline}, BOLS can have much greater power than other estimators when there are non-stationarity.
%In the sequential treatment regime literature, sequential decision making problems that do not have unique optimal policies are called ``exceptional laws"  \cite{luedtke2016statistical, robins2004optimal, robins2014discussion}. 
%Assuming a non-zero margin for standard bandit algorithms, as required by the CLT of \cite{lai1982least}, implies no exceptional laws.
%We show that the OLS estimator is asymptotically non-normal for exceptional laws. 
%Since the OLS estimator is a special case of the method-of-moments estimator \cite{Hazelton2011} (see Chapter 6 of \cite{cameron2005microeconometrics}), our results suggest that all method-of-moments estimators on adaptively-collected data may be asymptotically non-normal for exceptional laws. 
%\todo{MOM estimators}
%Note that when there is no margin, there is no unique optimal, regret-minimizing policy.
We found that the OLS estimator is asymptotically non-normal when the margin is zero due to the non-concentration of the action selection probabilities. 
Since the OLS estimator is a canonical example of a method-of-moments estimator \cite{Hazelton2011}, our results suggest that the inferential guarantees of standard method-of-moments estimators may fail to hold on adaptively collected data when there is no unique optimal, regret-minimizing policy.
We develop the Batched OLS estimator, which is asymptotically normal even when the action selection probabilities do not concentrate.
An open question is whether batched versions of general method-of-moments estimators could similarly be used for adaptive inference.
%We address the problem of non-concentration of action selection probabilities when the margin is zero with the Batched OLS estimator, which is asymptotically normal even when the action selection probabilities do not concentrate.
%Since the OLS estimator is a canonical example of a method-of-moments estimator, an open question is whether the batched estimator technique will work for general method-of-moments estimators.

\clearpage

\section*{Broader Impact}
Our work has the positive impact of encouraging the use of valid statistical inference methods on bandit data, which ultimately leads to more reliable scientific conclusions.
In addition, by providing a valid statistical inference method on bandit data, our work facilitates the use of bandit algorithms in experimentation. 
%uthors are required to include a statement of the broader impact of their work, including its ethical aspects and future societal consequences.  Authors should discuss both positive and negative outcomes, if any. For instance, authors should discuss a) who may benefit from this research, b) who may be put at disadvantage from this research, c) what are the consequences of failure of the system, and d) whether the task/method leverages biases in the data. If authors believe this is not applicable to them, authors can simply state this.

\begin{ack}
% NIAAA R01AA23187, NIDA P50DA039838, NCI U01CA229437.
Research reported in this paper was supported by National Institute on Alcohol Abuse and Al-coholism (NIAAA) of the National Institutes of Health under award number R01AA23187, Na-tional Institute on Drug Abuse (NIDA) of the National Institutes of Health under award number P50DA039838, National Cancer Institute (NCI) of the National Institutes of Health under award number U01CA229437, and by NIH/NIBIB and OD award number P41EB028242. The content is solely the responsibility of the authors and does not necessarily represent the official views of the National Institutes of Health

This material is based upon work supported by the National Science Foundation Graduate Research Fellowship Program under Grant No. DGE1745303. Any opinions, findings, and conclusions or recommendations expressed in this material are those of the author(s) and do not necessarily reflect the views of the National Science Foundation.

%Use unnumbered first level headings for the acknowledgments. All acknowledgments
%go at the end of the paper before the list of references. Moreover, you are required to declare 
%funding (financial activities supporting the submitted work) and competing interests (related financial activities outside the submitted work). 
%More information about this disclosure can be found at: \url{https://neurips.cc/Conferences/2020/PaperInformation/FundingDisclosure}.
%
%
%Do {\bf not} include this section in the anonymized submission, only in the final paper. You can use the \texttt{ack} environment provided in the style file to autmoatically hide this section in the anonymized submission.
\end{ack}

\bibliography{inference_bandits_bib}
\bibliographystyle{plain}

\appendix

\include{appendix/additional_simulations}

\include{appendix/triangularCLT}

\include{appendix/nonnormality}

\include{appendix/BOLS}

\include{appendix/BOLScontext}

\include{appendix/wdecorrelated}

\end{document}

%% file: appendix/additional_simulations.tex
\section{Simulation Details}
\label{appendix:simulations}

\subsection{W-Decorrelated Estimator}
\label{appendix:wdecorrelatedlambda}
For the $W$-decorrelated estimator \cite{deshpande}, for a batch size of $n$ and for $T$ batches, we set $\lambda$ to be the $\frac{1}{nT}$ quantile of $\lambda_{\min} ( \under{\boldX}^\top \under{\boldX}) / \log (n T)$, where $\lambda_{\min} (\under{\boldX}^\top \under{\boldX})$ denotes the minimum eigenvalue of $\under{\boldX}^\top \under{\boldX}$.
This procedure of choosing $\lambda$ is motivated by the conditions of Theorem 4 of \cite{deshpande} and follows the methods used by \cite{deshpande} in their simulation experiments.
We had to adjust the original procedure for choosing $\lambda$ used by \cite{deshpande} (who set $\lambda$ to the $0.15$ quantile of $\lambda_{\min} (\under{\boldX}^\top \under{\boldX})$), because they only evaluated the W-decorrelated method for when the total number of samples was $nT = 1000$ and valid values of $\lambda$ changes with the sample size.

\subsection{AW-AIPW Estimator}
Since the AW-AIPW test statistic for the treatment effect is not explicitly written in the original paper \cite{athey}, we now write the formulas for the AW-AIPW estimator of the treatment effect: $\hat{ \Delta }^{\TN{AW-AIPW}} := \betahat_1^{\TN{AW-AIPW}} - \betahat_0^{\TN{AW-AIPW}}$.
We use the variance stabilizing weights, equal to the square root of the sampling probabilities, $\sqrt{ \pi_t^{(n)} }$ and $\sqrt{ 1-\pi_t^{(n)} }$. 
Below, $N_{t,1} = \sum_{i=1}^n A_{t,i}$ and $N_{t,0} = \sum_{i=1}^n (1-A_{t,i})$.
\begin{equation*}
Y_{t,1} := \frac{ A_{t,i} }{ \pi_t^{(n)} } R_{t,i} + \bigg( 1 - \frac{ A_{t,i}^{(n)} }{ \pi_t^{(n)} } \bigg) \frac{ \sum_{t'=1}^{t-1} \sum_{i=1}^n A_{t,i}^{(n)} R_{t,i} }{ \sum_{t'=1}^{t-1} N_{t',1} }
\end{equation*}
\begin{equation*}
Y_{t,0} := \frac{ 1-A_{t,i}^{(n)} }{ 1-\pi_t^{(n)} } R_{t,i} + \bigg( 1 - \frac{ 1-A_{t,i} }{ 1 - \pi_t^{(n)} } \bigg) \frac{ \sum_{t'=1}^{t-1} \sum_{i=1}^n (1 - A_{t,i} ) R_{t,i} }{ \sum_{t'=1}^{t-1} N_{t',0} }
\end{equation*}
\begin{equation*}
\betahat_1^{\TN{AW-AIPW}} := \frac{ \sum_{t=1}^T \sum_{i=1}^n \sqrt{ \pi_t^{(n)} } Y_{t,1} }{ \sum_{t=1}^T \sum_{i=1}^n \sqrt{ \pi_t^{(n)} } }
~~~~\TN{ and }~~~~
\betahat_0^{\TN{AW-AIPW}} := \frac{ \sum_{t=1}^T \sum_{i=1}^n \sqrt{ 1 - \pi_t^{(n)} } Y_{t,0} }{ \sum_{t=1}^T \sum_{i=1}^n \sqrt{ 1 - \pi_t^{(n)} } }
\end{equation*}

The variance estimator for $\hat{ \Delta }^{\TN{AW-AIPW}}$ is $\hat{V}_0 + \hat{V}_1 + 2 \hat{C}_{0,1}$ where
\begin{equation*}
\hat{V}_1 := \frac{ \sum_{t=1}^T \sum_{i=1}^n  \pi_t^{(n)} ( Y_{t,1} - \betahat_1^{\TN{AW-AIPW}} )^2 }{ \bigg( \sum_{t=1}^T \sum_{i=1}^n \sqrt{ \pi_t^{(n)} } \bigg)^2 } 
~~~~\TN{ and }~~~~
\hat{V}_0 := \frac{ \sum_{t=1}^T \sum_{i=1}^n ( 1 - \pi_t^{(n)} ) ( Y_{t,0} - \betahat_0^{\TN{AW-AIPW}} )^2 }{ \bigg( \sum_{t=1}^T \sum_{i=1}^n \sqrt{ 1 - \pi_t^{(n)} } \bigg)^2 } 
\end{equation*}
\begin{equation*}
\hat{C}_{0,1} := - \frac{ \sum_{t=1}^T \sum_{i=1}^n \sqrt{ \pi_t^{(n)} ( 1 - \pi_t^{(n)} ) } ( Y_{t,1} - \betahat_1^{\TN{AW-AIPW}} ) ( Y_{t,0} - \betahat_0^{\TN{AW-AIPW}} ) }{ \bigg( \sum_{t=1}^T \sum_{i=1}^n \sqrt{ \pi_t^{(n)} } \bigg) \bigg( \sum_{t=1}^T \sum_{i=1}^n \sqrt{ 1 - \pi_t^{(n)} } \bigg) } 
\end{equation*}

\subsection{Self-Normalized Martingale Bound}
\label{appendix:snb}
By the self-normalized martingale bound of \cite{abbasi2011improved}, specifically Theorem 1 and Lemma 6, we have that in the two arm bandit setting,
%\begin{equation*}
%	\PP \left( \forall T, n \geq 1, \left| \frac{ \sum_{t=1}^T \sum_{i=1}^n A_{t,i} R_{t,i} }{ \sum_{t=1}^T N_{t,1} } - \beta_1 \right| \leq c_{1,T} 
%	\TN{ and } \left| \frac{ \sum_{t=1}^T \sum_{i=1}^n (1-A_{t,i}) R_{t,i} }{ \sum_{t=1}^T N_{t,0} } - \beta_0 \right| \leq c_{0,T} \right) \geq 1-\delta
%\end{equation*}
\begin{equation*}
	\PP \left( \forall T, n \geq 1, \left| \betahat_1^{\OLS} - \beta_1 \right| \leq c_{1,T} 
	\TN{ and } \left| \betahat_0^{\OLS} - \beta_0 \right| \leq c_{0,T} \right) \geq 1-\delta
\end{equation*}
where
\begin{equation*}
	c_{a,T} = \sqrt{ \sigma^2 \frac{1+\sum_{t=1}^T N_{t,a} }{\left( \sum_{t=1}^T N_{t,a} \right)^2} \left( 1 + 2 \log \left( \frac{2 \sqrt{ 1 + \sum_{t=1}^T N_{t,a} } }{ \delta } \right) \right) }
\end{equation*}
We estimate $\sigma^2$ using the procedure stated below for the OLS estimator. We reject the null hypothesis that $\Delta = 0$ whenever either the confidence bounds for the two arms are non-overlapping. Specifically when
\begin{equation*}
	\betahat_1^{\OLS} + c_{1,T} \leq \betahat_0^{\OLS} - c_{0,T}
	~~~\TN{ or }~~~
	\betahat_0^{\OLS} + c_{0,T} \leq \betahat_1^{\OLS} - c_{1,T}
\end{equation*}

\subsection{Estimating Noise Variance}
\label{appendix:noisevar}
\paragraph{OLS Estimator}
Given the OLS estimators for the means of each arm, $\betahat_1^{\OLS}, \betahat_0^{\OLS}$, we estimate the noise variance $\sigma^2$ as follows:
\begin{equation*}
\hat{ \sigma }^2 := \frac{1}{nT-2} \sum_{t=1}^T \sum_{i=1}^n \bigg( R_{t,i} - A_{t,i} \betahat_1^{\OLS} - (1-A_{t,i}) \betahat_0^{\OLS} \bigg)^2.
\end{equation*}
We use a degrees of freedom bias correction by normalizing by $nT-2$ rather than $nT$.
Since the W-decorrelated estimator is a modified version of the OLS estimator, we also use this same noise variance estimator for the W-decorrelated estimator;
we found that this worked well in practice, in terms of Type-1 error control.

\paragraph{Batched OLS}
Given the Batched OLS estimators for the means of each arm for each batch, $\betahat_{t,1}^{\TN{BOLS}}, \betahat_{t,0}^{\TN{BOLS}}$, we estimate the noise variance for each batch $\sigma_t^2$ as follows:
\begin{equation*}
\hat{ \sigma }_t^2 := \frac{1}{n-2} \sum_{i=1}^n \bigg( R_{t,i} - A_{t,i} \betahat_{t,1}^{\TN{BOLS}} - (1-A_{t,i}) \betahat_{t,0}^{\TN{BOLS}} \bigg)^2.
\end{equation*}
Again, we use a degrees of freedom bias correction by normalizing by $n-2$ rather than $n$. We prove the consistency of $\hat{ \sigma }_t^2$ (meaning $\hat{ \sigma }_t^2 \Pto \sigma^2$) in Corollary \ref{corollary:varianceconsistency}.
Using BOLS to test $H_0: \Delta = a$ vs. $H_1: \Delta \neq a$, we use the following test statistic:
\begin{equation*}
	\frac{1}{ \sqrt{T } } \sum_{t=1}^T \sqrt{ \frac{ N_{t,0} N_{t,1} }{ n \hat{\sigma}_t^2 } } ( \hat{\Delta}_t^{\TN{BOLS}} - a ).
\end{equation*}
Above, $N_{t,1} = \sum_{i=1}^n A_{t,i}$ and $N_{t,0} = \sum_{i=1}^n (1-A_{t,i})$.
For this test statistic, we use cutoffs based on the Student-t distribution, i.e., for $Y_t \iidsim t_{n-2}$ we use a cutoff $c_{\alpha/2}$ such that 
\begin{equation*}
	\PP \bigg( \bigg| \frac{1}{ \sqrt{T} } \sum_{t=1}^T Y_t \bigg| > c_{\alpha/2} \bigg) = \alpha.
\end{equation*}
We found $c_{\alpha/2}$ by simulating draws from the Student-t distribution.

\subsection{Non-Stationary Treatment Effect}
\label{appendix:nste}
%Alternatively we may believe that the margin itself varies from batch to batch. 
When we believe that the margin itself varies from batch to batch, we are able to construct a confidence region that contains the true margin $\Delta_t$ for each batch simultaneously with probability $1 - \alpha$. %by Theorem \ref{thm:bols}, 
%%%%%%%%%%%%%%%%%%%%%%%%
\begin{corollary}[Confidence band for margin for non-stationary bandits]
\label{corollary:nonstationary}
Assume the same conditions as Theorem \ref{thm:bols}. Let $z_x$ be $x^{\TN{th}}$ quantile of the standard Normal distribution, i.e., for $Z \sim \N(0, 1)$, $\PP( Z < z_{\alpha} ) = \alpha$. 
For each $t \in [1 \colon T]$, 
we define the interval 
\begin{equation*}
\boldL_t = \hat{\Delta}_t^{\OLS} \pm z_{1-\frac{\alpha}{2T}} \sqrt{ \frac{ \sigma^2 n}{ N_{t,0} N_{t,1} } }.
\end{equation*}
$\lim_{n \to \infty} \PP \big( \forall t \in [1:T], \Delta_t \in \boldL_t \big) \geq 1-\alpha$. Above, $N_{t,1} = \sum_{i=1}^n A_{t,i}$ and $N_{t,0} = \sum_{i=1}^n (1-A_{t,i})$.
\end{corollary}

\paragraph{Proof:}
Note that by Corollary \ref{corollary:BOLSmargin}, 
\begin{equation*}
\PP \big( \TN{exists some } t \in [1:T] ~ s.t. ~ \Delta_t \notin \boldL_t \big) 
\leq \sum_{t=1}^T \PP \big( \Delta_t \notin \boldL_t \big) 
\to \sum_{t=1}^T \frac{\alpha}{T}
= \alpha
\end{equation*}
where the limit is as $n \to \infty$. Since
\begin{equation*}
\PP \big( \forall t \in [1:T], \Delta_t \in \boldL_t  \big)
= 1 - \PP \big( \TN{exists some } t \in [1:T] ~ s.t. ~ \Delta_t \notin \boldL_t \big)
\end{equation*}
Thus,
\begin{equation*}
\lim_{n \to \infty} \PP \big( \forall t \in [1:T], \Delta_t \in \boldL_t \big) \geq 1 - \alpha ~~~ \qed
\end{equation*}

%%%%%%%%%%%%%%%%%%%%%%%% 
% that for at least one batch of students there is a difference in effectiveness of two teaching methods, i.e., 
We can also test the null hypothesis of no margin against the alternative that at least one batch has non-zero margin, i.e., $H_0 \colon \forall t \in [1 \colon T], \Delta_t = 0$ vs. $H_1 \colon \exists t \in [1 \colon T] ~ s.t. ~ \Delta_t \not= 0$.
Note that the global null stated above is of great interest in the mobile health literature \cite{klasnja2015microrandomized,liao2016sample}.
Specifically we use the following test statistic:
\begin{equation*}
	\label{eqn:BOLSNSTE}
	\sum_{t=1}^T \frac{ N_{t,0} N_{t,1} }{ \sigma^2 n} ( \hat{\Delta}_t^{\OLS} - 0 )^2,
\end{equation*}
which by Theorem \ref{thm:bols} converges in distribution to a chi-squared distribution with $T$ degrees of freedom under the null $\Delta_t = 0$ for all $t$. 

%In the non-stationary treatment effect simulations, we test the null that $H_0 \colon \forall t \in [1 \colon T], \beta_{t,1} - \beta_{t,0} = 0$ vs. $H_1 \colon \exists t \in [1 \colon T], \beta_{t,1} - \beta_{t,0} \neq 0$. 
%To test this we use the following test statistic:
%\begin{equation}
%\label{eqn:BOLSNSTE}
%	\frac{1}{ T } \sum_{t=1}^T \frac{ N_{t,0} N_{t,1} }{ n \hat{\sigma}_t^2 } ( \hat{\Delta}_t^{\TN{BOLS}} - 0 )^2.
%\end{equation}
To account for estimating noise variance $\sigma^2$, in our simulations for this test statistic, we use cutoffs based on the Student-t distribution, i.e., for $Y_t \iidsim t_{n-2}$ we use a cutoff $c_{\alpha/2}$ such that 
\begin{equation*}
	\PP \bigg( \frac{1}{ T } \sum_{t=1}^T Y_t^2 > c_{\alpha} \bigg) = \alpha.
\end{equation*}
We found $c_{\alpha}$ by simulating draws from the Student-t distribution.

In the plots below we call the test statistic in \eqref{eqn:BOLSNSTE} ``BOLS Non-Stationary Treatment Effect'' (BOLS NSTE).
BOLS NSTE performs poorly in terms of power compared to other test statistics in the stationary setting; 
however, \textit{in the non-stationary setting, BOLS NSTE significantly outperforms all other test statistics, which tend to have low power when the average treatment effect is close to zero.}
Note that the W-decorrelated estimator performs well in the left plot of Figure \ref{fig:nonstationary}; 
this is because as we show in Appendix \ref{appendix:wdecorrelated}, the W-decorrelated estimator upweights samples from the earlier batches in the study.
So when the treatment effect is large in the beginning of the study, the W-decorrelated estimator has high power and when the treatment effect is small or zero in the beginning of the study, the W-decorrelated estimator has low power.

\bigskip \bigskip

\begin{figure*}[h]
\centering
  \includegraphics[width=0.49\textwidth]{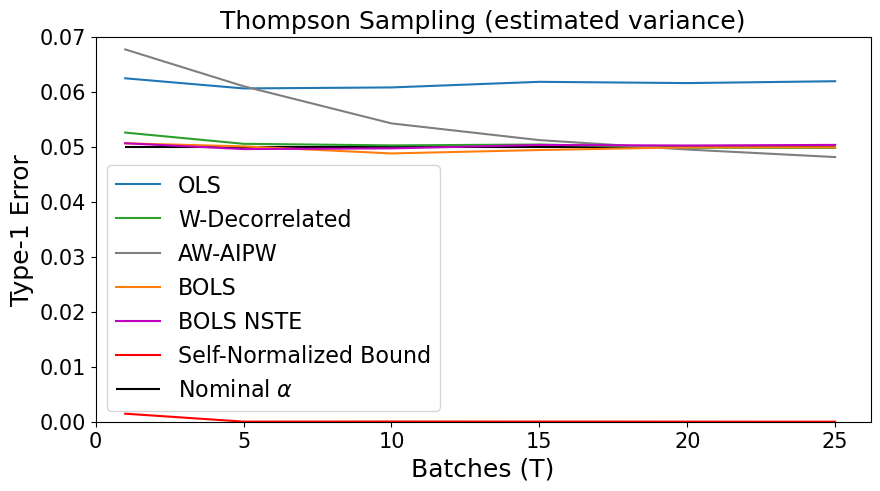}
  ~
  \includegraphics[width=0.49\textwidth]{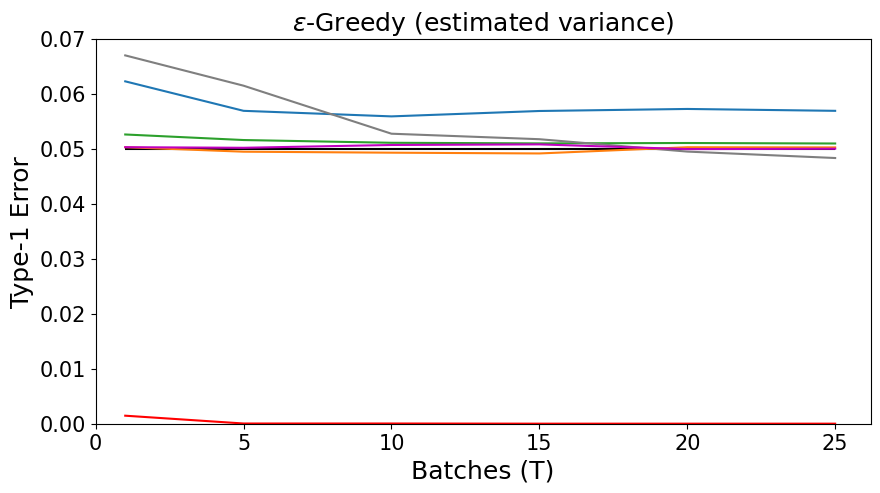}
\caption{ \bo{Stationary Setting:} Type-1 error for a two-sided test of $H_0 \colon \Delta = 0$ vs. $H_1 \colon \Delta \neq 0$ ($\alpha=0.05$). % for estimators of margin 
 We set $\beta_1 = \beta_0 = 0$, $n = 25$, and a clipping constraint of $0.1 \leq \pi_t^{(n)} \leq 0.9$.
 We use 100k Monte Carlo simulations and standard errors are $<0.001$.}
%Type-1 error for estimators of treatment effect for a two-sided test of $H_0 \colon \Delta = 0$ vs. $H_1 \colon \Delta \neq 0$ ($\alpha=0.05$).
% We set $\beta_1 = \beta_0 = 0$ and use $25$ samples per batch.
% We use a fixed clipping constraint of $0.1 \leq \pi_t^{(n)} \leq 0.9$.
% Standard errors for the above simulations are $<0.001$.}
 \label{fig:type1}
\end{figure*}
\begin{figure*}[h]
\centering
  \includegraphics[width=0.49\textwidth]{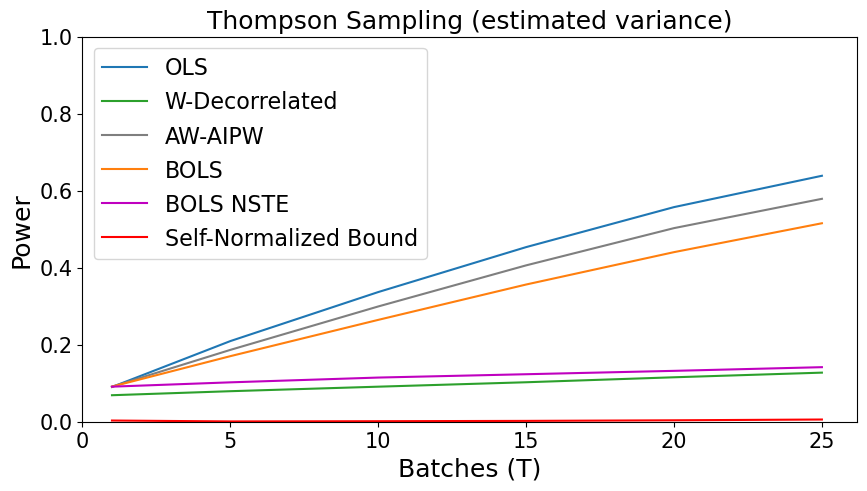}
  ~
  \includegraphics[width=0.49\textwidth]{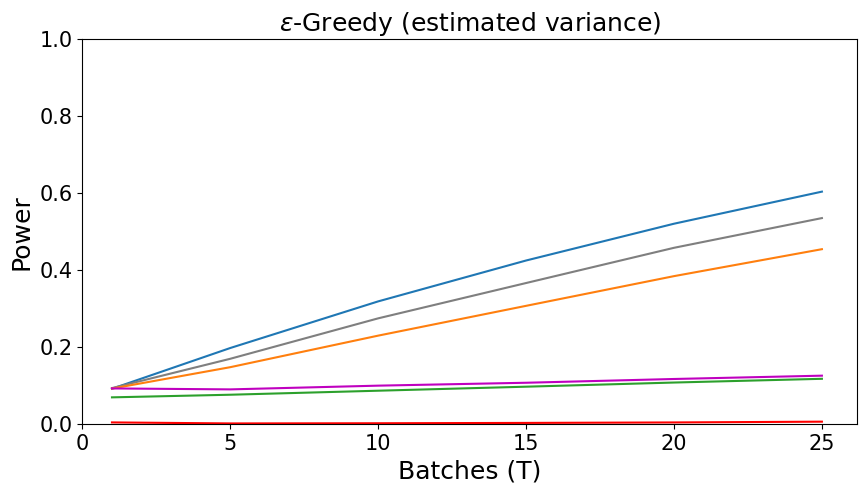}
\caption{ \bo{Stationary Setting:} Power for a two-sided test of $H_0 \colon \Delta = 0$ vs. $H_1 \colon \Delta \neq 0$ ($\alpha=0.05$).
 We set $\beta_1 = 0$, $\beta_0 = 0.25$, $n = 25$, and a clipping constraint of $0.1 \leq \pi_t^{(n)} \leq 0.9$.
 We use 100k Monte Carlo simulations and standard errors are $<0.002$.
 We account for Type-1 error inflation as described in Section \ref{sec:simprocedure}.}
%
%We plot the power for estimators of treatment effect for a two-sided test of $H_0 \colon \Delta = 0$ vs. $H_1 \colon \Delta \neq 0$ ($\alpha=0.05$).
% We set $\beta_1 = 0$, $\beta_0 = 0.25$, and use $25$ samples per batch.
% We use a fixed clipping constraint of $0.1 \leq \pi_t^{(n)} \leq 0.9$.
% All standard errors for the above simulations are $<0.002$.}
 \label{fig:power}
\end{figure*}
\begin{figure*}[h]
\centering
 	\includegraphics[width=0.49\textwidth]{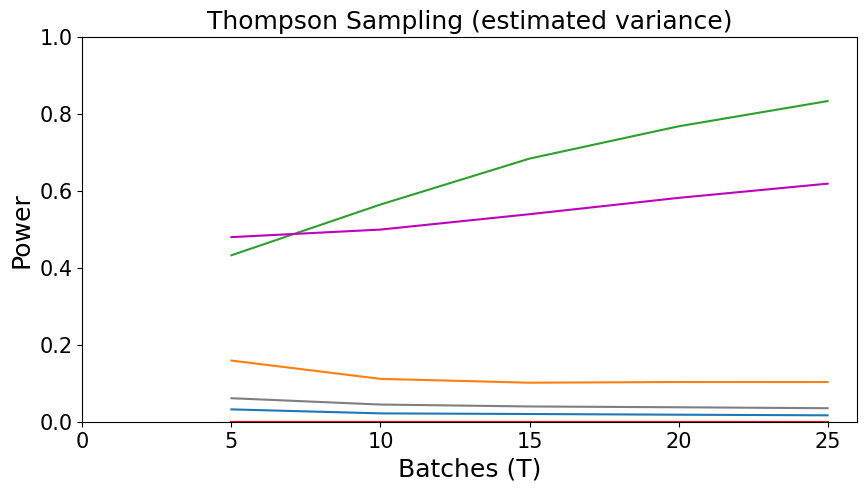}
  ~
  \includegraphics[width=0.49\textwidth]{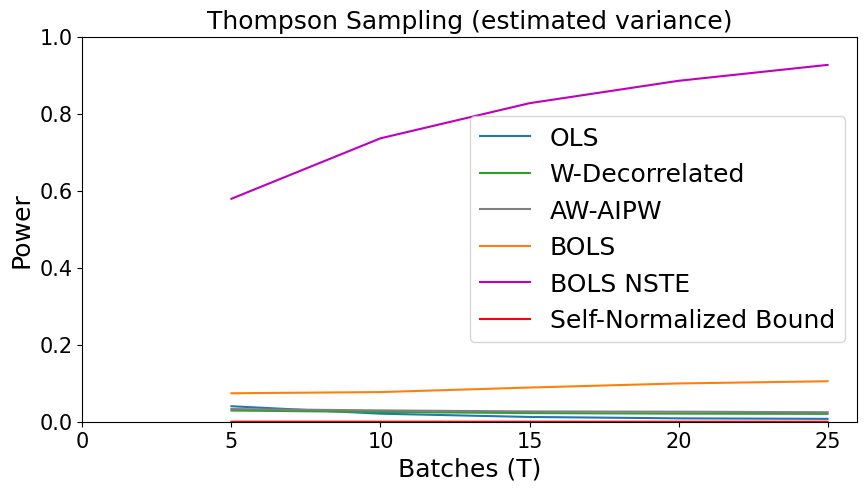}
  \vspace{-2mm}
   \includegraphics[width=0.49\textwidth]{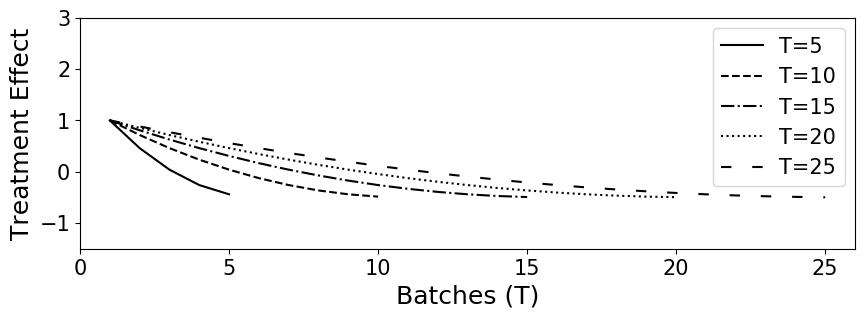}
   ~
   \includegraphics[width=0.49\textwidth]{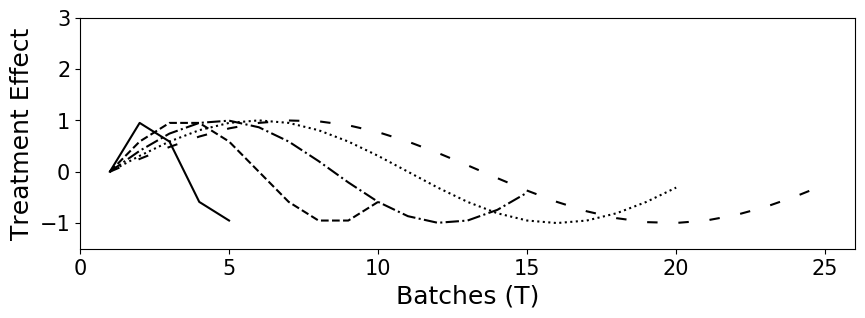}
\caption{ \bo{Nonstationary setting:} The two upper plots display the power of estimators for a two-sided test of $H_0 \colon \forall t \in [1 \colon T], ~ \beta_{t,1} - \beta_{t,0} = 0$ vs. $H_1 \colon \exists t \in [1 \colon T], \beta_{t,1} - \beta_{t,0} \neq 0$ ($\alpha=0.05$).
The two lower plots display two treatment effect trends; the left plot considers a decreasing trend (quadratic function) and the right plot considers a oscillating trend ($\sin$ function).
We set $n = 25$, and a clipping constraint of $0.1 \leq \pi_t^{(n)} \leq 0.9$.
 We use 100k Monte Carlo simulations and standard errors are $<0.002$.}
\label{fig:nonstationary}
\end{figure*}

%% file: appendix/triangularCLT.tex
\section{Asymptotic Normality of the OLS Estimator}
\label{appendix:triangularCLT}

\begin{condition}[Weak moments]
\label{cond:weakmoments}
$\forall t, n, i$, $\E [ \epsilon_{t, i}^2 \big| \G_{t-1}^{(n)} ] = \sigma^2$ and 
for all $\forall t, n, i$, $\E[ \varphi( \epsilon_{t,i}^2 ) | \G_{t-1}^{(n)} ] < M < \infty$ a.s. for some function $\varphi$ where $\lim_{x \to \infty} \frac{ \varphi(x) }{x} \to \infty$.
\end{condition}

\begin{condition}[Stability] 
\label{cond:stability}
There exists a sequence of nonrandom positive-definite symmetric matrices, $\underline{ \boldV }_n$, such that 
\vspace{-3mm}
\begin{enumerate}[label=(\alph*)]
	\item \label{cond:stable} $\underline{ \boldV }_n^{-1} \big( \sum_{t=1}^T \sum_{i=1}^n \boldX_{t,i} \boldX_{t,i}^\top \big)^{\frac{1}{2}}
= \underline{ \boldV }_n^{-1} \big( \under{\boldX}^\top \under{\boldX} \big)^{\frac{1}{2}} \Pto \under{\bo{I}}_p$
	\vspace{-1mm}
	\item \label{cond:UAN}$\max_{i \in [1 \colon n], t \in [1 \colon T]} \| \underline{ \boldV }_n^{-1} \boldX_{t,i} \|_2 \Pto 0$
\end{enumerate}
\end{condition}

\begin{theorem}[Triangular array version of Lai \& Wei (1982), Theorem 3]
\label{thm:triangularCLT}
Let $\boldX_{t,i} \in \real^p$ be non-anticipating with respect to filtration $\{ \G_t^{(n)} \}_{t=1}^T$, so $\boldX_{t,i}$ is $\G_{t-1}^{(n)}$ measurable. 
We assume the following conditional mean model for rewards: %For all $t \in [1 \colon T]$, $i \in [1 \colon n]$,
\begin{equation*}
\E \big[ R_{t,i} \big| \G_{t-1}^{(n)} \big] = \boldX_{t,i}^\top \bs{\beta}.
\end{equation*}
We define $\epsilon_{t,i} := R_{t,i} - \boldX_{t,i}^\top \bs{\beta}$.
Note that $\{ \epsilon_{t,i} \}_{i=1,t=1}^{i=n,t=T}$ is a martingale difference array with respect to filtration $\{ \G_t^{(n)} \}_{t=1}^T$.

Assuming Conditions \ref{cond:weakmoments} and \ref{cond:stability}, as $n \to \infty$,
\begin{equation*}
	( \under{\boldX}^\top \under{\boldX} )^{1/2} ( \bs{\betahat}^{\OLS} - \bs{\beta}) 
	\Dto \N( 0, \sigma^2 \under{\bo{I}}_p)
\end{equation*}
Note, in the body of the paper we state that this theorem holds in the two-arm bandit case assuming Conditions \ref{cond:banditstability} and \ref{cond:moments}.
Note that Condition \ref{cond:moments} is sufficient for Condition \ref{cond:weakmoments} and Condition \ref{cond:banditstability} is sufficient for Condition \ref{cond:stability} in the two-arm bandit case.
\end{theorem}

\paragraph{Proof:}
\begin{equation*}
\bs{\betahat}^{\OLS} = ( ( \underline{\boldX}^\top \underline{\boldX} )^{-1} \underline{\boldX}^{(n), \top} \boldR^{(n)}
= (\underline{\boldX}^\top \underline{\boldX})^{-1} \underline{\boldX}^\top ( \underline{\boldX}\bs{\beta} + \bs{\epsilon})
\end{equation*}
\begin{equation*}
\bs{\betahat}^{\OLS} - \bs{\beta} 
= (\underline{\boldX}^\top \underline{\boldX})^{-1} \underline{\boldX}^\top \bs{\epsilon}
= \bigg( \sum_{t=1}^T \sum_{i=1}^n \boldX_{t,i} \boldX_{t,i}^\top \bigg)^{-1}  \sum_{t=1}^T \sum_{i=1}^n \boldX_{t,i} \epsilon_{t,i}
\end{equation*}

It is sufficient to show that as $n \to \infty$:
\begin{equation*}
(\underline{\boldX}^\top \underline{\boldX})^{-1/2} \sum_{t=1}^T \sum_{i=1}^n \boldX_{t,i} \epsilon_{t,i}
\Dto \N( 0, \sigma^2 \under{\bo{I}}_p)
\end{equation*}
By Slutsky's Theorem and Condition \ref{cond:stability} \ref{cond:stable}, it is also sufficient to show that as $n \to \infty$,
\begin{equation*}
\underline{\boldV}_n^{-1} \sum_{t=1}^T \sum_{i=1}^n \boldX_{t,i} \epsilon_{t,i} \Dto \N(0, \sigma^2 \under{\bo{I}}_p)
\end{equation*}

By Cramer-Wold device, it is sufficient to show multivariate normality if for any fixed $\boldc \in \real^p$ s.t. $\| \boldc \|_2 = 1$, as $n \to \infty$,
\begin{equation*}
\boldc^\top \underline{\boldV}_n^{-1} \sum_{t=1}^T \sum_{i=1}^n \boldX_{t,i} \epsilon_{t,i} \Dto \N(0, \sigma^2 )
\end{equation*}

We will prove this central limit theorem by using a triangular array martingale central limit theorem, specifically Theorem 2.2 of \cite{dvoretzky1972asymptotic}.
We will do this by letting $Y_{t,i} = \boldc^T \underline{\boldV}_n^{-1} \boldX_{t,i} \epsilon_{t,i}$.
The theorem states that as $n \to \infty$, $\sum_{t=1}^T \sum_{i=1}^n Y_{t,i} \Dto \N(0, \sigma^2)$ if the following conditions hold as $n \to \infty$:
\begin{enumerate}[label=(\alph*)]
    \item $\sum_{t=1}^T \sum_{i=1}^n E[ Y_{t,i} | \G_{t-1}^{(n)} ] \Pto 0$
    \item $\sum_{t=1}^T \sum_{i=1}^n E[ Y_{t,i}^2 | \G_{t-1}^{(n)} ] \Pto \sigma^2$
    \item $\forall \delta > 0, \sum_{t=1}^T \sum_{i=1}^n E \big[ Y_{t,i}^2 \II_{( | Y_{t,i} | > \delta )} \big| \G_{t-1}^{(n)} \big] \Pto 0$
\end{enumerate}

%%%%%%%%%%%%%%%%%%%%%%%%%%%%%%%%%%%%
\paragraph{Useful Properties}
%%%%%%%%%%%%%%%%%%%%%%%%%%%%%%%%%%%%

Note that by Cauchy-Schwartz and Condition \ref{cond:stability} \ref{cond:UAN}, as $n \to \infty$,
\begin{equation*}
\max_{i \in [1 \colon n], t \in [1 \colon T]} \big| \boldc^\top \underline{ \boldV }_n^{-1} \boldX_{t,i}  \big| \leq
\max_{i \in [1 \colon n], t \in [1 \colon T]} \| \boldc \|_2 \| \underline{ \boldV }_n^{-1} \boldX_{t,i} \|_2 \Pto 0
\end{equation*}
%%%%%%%%%%%%%%%%%%%%%%%%%%%%%%
By continuous mapping theorem and since the square function on non-negative inputs is order preserving, as $n \to \infty$,
\begin{equation}
	\bigg( \max_{i \in [1 \colon n], t \in [1 \colon T]} \big| \boldc^\top \underline{ \boldV }_n^{-1} \boldX_{t,i} \big| \bigg)^2
    = \max_{i \in [1 \colon n], t \in [1 \colon T]} \big( \boldc^\top \underline{ \boldV }_n^{-1} \boldX_{t,i} \big)^2 \Pto 0
    \label{eqn:max1.5}
\end{equation}
%%%%%%%%%%%%%%%%%%%%%%%%%%%%%%
By Condition \ref{cond:stability} \ref{cond:stable} and continuous mapping theorem,
$\boldc^\top \underline{ \boldV }_n^{-1} ( \boldX_{t,i}^\top \boldX_{t,i} )^{1/2} \Pto \boldc^\top$, so
\begin{equation*}
\boldc^\top \underline{ \boldV }_n^{-1} ( \boldX_{t,i}^\top \boldX_{t,i} )^{1/2} ( \boldX_{t,i}^\top \boldX_{t,i} )^{1/2} \underline{ \boldV }_n^{-1} \boldc 
\Pto \boldc^\top \boldc = 1
\end{equation*}
%%%%%%%%%%%%%%%%%%%%%%%%%%%%%%
Thus,
\begin{equation*}
\boldc^\top \underline{ \boldV }_n^{-1} \bigg( \sum_{t=1}^T \sum_{i=1}^n \boldX_{t,i} \boldX_{t,i}^\top \bigg) \underline{ \boldV }_n^{-1} \boldc 
= \sum_{t=1}^T \sum_{i=1}^n \boldc^\top \underline{ \boldV }_n^{-1} \boldX_{t,i} \boldX_{t,i}^\top \underline{ \boldV }_n^{-1} \boldc  \Pto 1
\end{equation*}
%%%%%%%%%%%%%%%%%%%%%%%%%%%%%%
Since $\boldc^\top \underline{ \boldV }_n^{-1} \boldX_{t,i}$ is a scalar, as $n \to \infty$,
\begin{equation}
    \sum_{t=1}^T \sum_{i=1}^n ( \boldc^\top \underline{ \boldV }_n^{-1} \boldX_{t,i} )^2 \Pto 1 
    \label{eqn:sum1}
\end{equation}

\paragraph{Condition (a): Martingale}
\begin{equation*}
\sum_{t=1}^T \sum_{i=1}^n \E[ \boldc^\top \underline{ \boldV }_n^{-1} \boldX_{t,i} \epsilon_{t,i} | \G_{t-1}^{(n)} ]
= \sum_{t=1}^T \sum_{i=1}^n \boldc^\top \underline{ \boldV }_n^{-1} \boldX_{t,i} \E[ \epsilon_{t,i} | \G_{t-1}^{(n)} ]
= 0
\end{equation*}

\paragraph{Condition (b): Conditional Variance}
\begin{equation*}
\sum_{t=1}^T \sum_{i=1}^n \E[ ( \boldc^\top \underline{ \boldV }_n^{-1} \boldX_{t,i} )^2 \epsilon_{t,i}^2 | \G_{t-1}^{(n)} ] 
= \sum_{t=1}^T \sum_{i=1}^n ( \boldc^\top \underline{ \boldV }_n^{-1} \boldX_{t,i} )^2 \E[ \epsilon_{t,i}^2 | \G_{t-1}^{(n)} ]
= \sigma^2 \sum_{t=1}^T \sum_{i=1}^n ( \boldc^\top \underline{ \boldV }_n^{-1} \boldX_{t,i} )^2 \Pto \sigma^2
\end{equation*}
where the last equality holds by Condition \ref{cond:weakmoments} and the limit holds by \eqref{eqn:sum1} as $n \to \infty$.

\paragraph{Condition (c): Lindeberg Condition}
Let $\delta > 0$. We want to show that as $n \to \infty$,
\begin{equation*}
\sum_{t=1}^T \sum_{i=1}^n Z_{t,i}^2 
\E \bigg[ \epsilon_{t,i}^2 \II_{( Z_{t,i}^2 \epsilon_{t,i}^2 > \delta^2 )} \bigg| \G_{t-1}^{(n)} \bigg] \Pto 0
\end{equation*}
where above, we define $Z_{t,i}^{(n)} := \boldc^\top \underline{ \boldV }_n^{-1} \boldX_{t,i}$.
By Condition \ref{cond:weakmoments}, we have that for all $n \geq 1$,
\begin{equation*}
\max_{t \in [1 \colon T], i \in [1 \colon n]} \E[ \varphi( \epsilon_{t,i}^2 ) | \G_{t-1}^{(n)} ] < M
\end{equation*}
Since we assume that $\lim_{x \to \infty} \frac{ \varphi(x) }{x} = \infty$, for all $m \geq 1$, there exists a $b_m$ s.t. $\varphi(x) \geq m M x$ for all $x \geq b_m$. 
So, for all $n, t, i$,
\begin{equation*}
M \geq \E[ \varphi( \epsilon_{t,i}^2 ) | \G_{t-1}^{(n)} ] \geq \E[ \varphi( \epsilon_{t,i}^2 ) \II_{( \epsilon_{t,i}^2 \geq b_m )} | \G_{t-1}^{(n)} ] 
\geq m M \E[ \epsilon_{t,i}^2 \II_{( \epsilon_{t,i}^2 \geq b_m )} | \G_{t-1}^{(n)} ]
\end{equation*}
Thus, 
\begin{equation*}
\max_{t \in [1 \colon T], i \in [1 \colon n]} \E[ \epsilon_{t,i}^2 \II_{( \epsilon_{t,i}^2 \geq b_m )} | \G_{t-1}^{(n)} ] \leq \frac{1}{m}
\end{equation*}
%%%%%%%%%%%%%%%%%%%%%%%%%%%%%%
So we have that
\begin{equation*}
\sum_{t=1}^T \sum_{i=1}^n Z_{t,i}^2 
\E \big[ \epsilon_{t,i}^2 \II_{( Z_{t,i}^2 \epsilon_{t,i}^2 > \delta^2 )} \big| \G_{t-1}^{(n)} \big]
\end{equation*}
\begin{equation*}
= \sum_{t=1}^T \sum_{i=1}^n Z_{t,i}^2 
\bigg( \E \bigg[ \epsilon_{t,i}^2 \II_{( Z_{t,i}^2 \epsilon_{t,i}^2 > \delta^2 )} \bigg| \G_{t-1}^{(n)} \bigg]  \II_{( Z_{t,i}^2 \leq \delta^2 / b_m )} +
 \E \bigg[ \epsilon_{t,i}^2 \II_{( Z_{t,i}^2 \epsilon_{t,i}^2 > \delta^2 )} \bigg| \G_{t-1}^{(n)} \bigg] \II_{( Z_{t,i}^2 > \delta^2 / b_m )} \bigg)
\end{equation*}
\begin{equation*}
\leq \sum_{t=1}^T \sum_{i=1}^n Z_{t,i}^2 
\bigg( \E \bigg[ \epsilon_{t,i}^2 \II_{( \epsilon_{t,i}^2 > b_m )} \bigg| \G_{t-1}^{(n)} \bigg] +
\sigma^2 \II_{( Z_{t,i}^2 > \delta^2 / b_m )} \bigg)
\end{equation*}
\begin{equation*}
\leq \bigg( \frac{1}{m} + \sigma^2 \II_{( \max_{t' \in [1 \colon T], j \in [1 \colon n]} Z_{t',j}^2 > \delta^2 / b_m )} \bigg) \sum_{t=1}^T \sum_{i=1}^n Z_{t,i}^2
\end{equation*}
%%%%%%%%%%%%%%%%%%%%%%%%%%%%%%
By Slutsky's Theorem and \eqref{eqn:sum1}, it is sufficient to show that as $n \to \infty$,
\begin{equation*}
\frac{1}{m} + \sigma^2 \II_{( \max_{t' \in [1 \colon T], j \in [1 \colon n]} Z_{t',j}^2 > \delta^2 / b_m )} \Pto 0
\end{equation*}
%%%%%%%%%%%%%%%%%%%%%%%%%%%%%%
For any $\epsilon > 0$, 
\begin{equation*}
\PP \bigg( \frac{1}{m} + \sigma^2 \II_{( \max_{t' \in [1 \colon T], j \in [1 \colon n]} Z_{t',j}^2 > \delta^2 / b_m )} > \epsilon \bigg)
\leq \II_{ ( \frac{1}{m} > \frac{ \epsilon }{2} ) } + \PP \bigg( \sigma^2 \II_{( \max_{t' \in [1 \colon T], j \in [1 \colon n]} Z_{t',j}^2 > \delta^2 / b_m )} > \frac{ \epsilon }{2} \bigg)
\end{equation*}
We can choose $m$ such that $\frac{1}{m} \leq \frac{ \epsilon }{2}$, so $\PP \big( \frac{1}{m} > \frac{ \epsilon }{2} \big) = 0$.
For the second term (note that $m$ is now fixed),
\begin{equation*}
 \PP \bigg( \sigma^2 \II_{( \max_{t' \in [1 \colon T], j \in [1 \colon n]} Z_{t',j}^2 > \delta^2 / b_m )} > \frac{ \epsilon }{2} \bigg) \leq 
\PP \bigg( \max_{t' \in [1 \colon T], j \in [1 \colon n]} Z_{t',j}^2 > \delta^2 / b_m \bigg) \to 0
\end{equation*}
where the last limit holds by \eqref{eqn:max1.5} as $n \to \infty$. $\qed$

\subsection{Corollary \ref{corollary:sufficientcond} (Sufficient conditions for Theorem \ref{thm:triangularCLT})}
\begin{itshape}
Under Conditions \ref{cond:moments} and \ref{cond:condiid}, when \bo{the treatment effect is non-zero} data collected in batches using $\epsilon$-greedy, Thompson Sampling, or UCB with a fixed clipping constraint (see Definition \ref{def:clipping}) will satisfy Theorem \ref{thm:triangularCLT} conditions. 
\end{itshape}
\vspace{-2mm}

\paragraph{Proof:}
The only condition of Theorem \ref{thm:triangularCLT} that needs to verified is Condition \ref{cond:banditstability}.
To satisfy Condition \ref{cond:banditstability}, it is sufficient to show that for any given $\Delta$, for some constant $c \in (0, T)$,
\begin{equation*}
	\frac{1}{n} \sum_{t=1}^T \sum_{i=1}^n A_{t,i} = \frac{1}{n} \sum_{t=1}^T N_{t,1} \Pto c.
\end{equation*}

\bo{$\epsilon$-greedy}
We assume without loss of generality that $\Delta > 0$ and $\pi_1^{(n)} = \frac{1}{2}$. 
Recall that for $\epsilon$-greedy, for $a \in [2 \colon T]$,
\begin{equation*}
\pi_a^{(n)} = \begin{cases}
	1 - \frac{\epsilon}{2} & \TN{ if } \frac{ \sum_{t=1}^a \sum_{i=1}^n A_{t,i} R_{t,i} }{ \sum_{t'=1}^a N_{t',1} } > \frac{ \sum_{t=1}^a \sum_{i=1}^n (1-A_{t,i}) R_{t,i} }{ \sum_{t'=1}^a N_{t',0} } \\
	\frac{\epsilon}{2} & \TN{ otherwise } 
\end{cases}
\end{equation*}
Thus to show that $\pi_a^{(n)} \Pto 1 - \frac{\epsilon}{2}$ for all $a \in [2 \colon T]$, it is sufficient to show that 
\begin{equation}
\label{eqn:epsilonprob}
\PP \bigg( \frac{ \sum_{t=1}^a \sum_{i=1}^n A_{t,i} R_{t,i} }{ \sum_{t'=1}^a N_{t',1} } > \frac{ \sum_{t=1}^a \sum_{i=1}^n (1-A_{t,i}) R_{t,i} }{ \sum_{t'=1}^a N_{t',0} } \bigg) \to 1
\end{equation}
To show \eqref{eqn:epsilonprob}, it is equivalent to show that
\begin{equation}
\label{eqn:epsilonprob2}
\PP \bigg( \Delta > \frac{ \sum_{t=1}^a \sum_{i=1}^n (1-A_{t,i}) \epsilon_{t,i} }{ \sum_{t'=1}^a N_{t',0} } - \frac{ \sum_{t=1}^a \sum_{i=1}^n A_{t,i} \epsilon_{t,i} }{ \sum_{t'=1}^a N_{t',1} } \bigg) \to 1
\end{equation}
To show \eqref{eqn:epsilonprob2}, it is sufficient to show that
\begin{equation}
\label{eqn:epsilonprob3}
\frac{ \sum_{t=1}^a \sum_{i=1}^n (1-A_{t,i}) \epsilon_{t,i} }{ \sum_{t'=1}^a N_{t',0} } - \frac{ \sum_{t=1}^a \sum_{i=1}^n A_{t,i} \epsilon_{t,i} }{ \sum_{t'=1}^a N_{t',1} } \Pto 0.
\end{equation}
To show \eqref{eqn:epsilonprob3}, it is equivalent to show that
\begin{equation}
\label{eqn:epsilonprob4}
\sum_{t=1}^a \frac{  \sqrt{ N_{t,0} } }{ \sum_{t'=1}^a N_{t',0} } \frac{ \sum_{i=1}^n (1-A_{t,i}) \epsilon_{t,i} }{ \sqrt{ N_{t,0} } } 
- \sum_{t=1}^a \frac{ \sqrt{ N_{t,1} } }{ \sum_{t'=1}^a N_{t',1} } \frac{ \sum_{i=1}^n A_{t,i} \epsilon_{t,i} }{ \sqrt{ N_{t,1} } } \Pto 0.
\end{equation}

By Lemma \ref{lemma:ratio}, for all $t \in [1 \colon T]$, 
\begin{equation*}
\frac{N_{t,1}}{\pi_t^{(n)} n} \Pto 1
\end{equation*}

Thus by Slutsky's Theorem, to show \eqref{eqn:epsilonprob4}, it is sufficient to show that
\begin{equation}
\label{eqn:epsilonprob5}
\sum_{t=1}^a \frac{  \sqrt{ n (1-\pi_t^{(n)}) } }{ n \sum_{t'=1}^a (1-\pi_{t'}^{(n)}) } \frac{ \sum_{i=1}^n (1-A_{t,i}) \epsilon_{t,i} }{ \sqrt{ N_{t,0} } }
- \sum_{t=1}^a \frac{ \sqrt{ n \pi_t^{(n)}} }{ n \sum_{t'=1}^a \pi_{t'}^{(n)} } \frac{ \sum_{i=1}^n A_{t,i} \epsilon_{t,i} }{ \sqrt{ N_{t,1} } } \Pto 0.
\end{equation}

Since $\pi_t^{(n)} \in [ \frac{\epsilon}{2}, 1-\frac{\epsilon}{2}]$ for all $t, n$, the left hand side of \eqref{eqn:epsilonprob5} equals the following:
\begin{equation*}
\sum_{t=1}^a o_p(1) \frac{ \sum_{i=1}^n (1-A_{t,i}) \epsilon_{t,i} }{ \sqrt{ N_{t,0} } }
- \sum_{t=1}^a o_p(1) \frac{ \sum_{i=1}^n A_{t,i} \epsilon_{t,i} }{ \sqrt{ N_{t,1} } } \Pto 0.
\end{equation*}
The above limit holds because by Thereom \ref{thm:bols}, we have that
\small
\begin{equation}
\label{eqn:bolsresult}
\bigg( \frac{ \sum_{i=1}^n A_{1,i} \epsilon_{1,i} }{ \sqrt{ N_{1,1} } }, \frac{ \sum_{i=1}^n (1-A_{1,i}) \epsilon_{1,i} }{ \sqrt{ N_{1,0} } }, ...,  
\frac{ \sum_{i=1}^n A_{T,i} \epsilon_{T,i} }{ \sqrt{ N_{T,1} } },  \frac{ \sum_{i=1}^n (1-A_{T,i}) \epsilon_{T,i} }{ \sqrt{ N_{T,0} } } \bigg) 
\Dto \N( \bs{0}, \sigma^2 \under{\bs{I}}_{2T} ).
\end{equation}
\normalsize

Thus, by Slutsky's Theorem and Lemma \ref{lemma:ratio}, we have that
\begin{equation*}
\frac{1}{n} \sum_{t=1}^T N_{t,1} \Pto \frac{1}{2} + (T-1) (1 - \frac{\epsilon}{2})
~~~~~~~~~\TN{ and }~~~~~~~~~~
\frac{1}{n} \sum_{t=1}^T N_{t,0} \Pto \frac{1}{2} + (T-1) \frac{\epsilon}{2}
\end{equation*}

\bo{Thompson Sampling}
We assume without loss of generality that $\Delta > 0$ and $\pi_1^{(n)} = \frac{1}{2}$. 
Recall that for Thompson Sampling with independent standard normal priors ($\tilde{\beta}_1, \tilde{\beta}_0 \iidsim \N(0, 1)$) for $a \in [2 \colon T]$,
\begin{equation*}
\pi_a^{(n)} = \pi_{\min} \vee \big[ \pi_{\max} \wedge \PP( \tilde{\beta}_1 > \tilde{\beta}_0 ~|~ H_{a-1}^{(n)} ) \big]
\end{equation*}

Given the independent standard normal priors on $\tilde{\beta}_1, \tilde{\beta}_0$, we have the following posterior distribution:
\begin{multline*}
\tilde{\beta}_1 - \tilde{\beta}_0 ~|~ H_{a-1}^{(n)} 
\sim \N \bigg( \frac{ \sum_{t=1}^{a-1} \sum_{i=1}^n A_{t,i} R_{t,i} }{\sigma^2 + \sum_{t=1}^{a-1} N_{t,1}}
- \frac{ \sum_{t=1}^{a-1} \sum_{i=1}^n (1-A_{t,i}) R_{t,i} }{\sigma^2 + \sum_{t=1}^{a-1} N_{t,0}}, \\
\frac{ \sigma^2 (\sigma^2 + \sum_{t=1}^{a-1} N_{t,1}) + \sigma^2 ( \sigma^2 + \sum_{t=1}^{a-1} N_{t,0}) }{( \sigma^2 + \sum_{t=1}^{a-1} N_{t,0})( \sigma^2 + \sum_{t=1}^{a-1} N_{t,1})} \bigg)
\end{multline*}
\begin{equation*}
=: \N \big( \mu_{a-1}^{(n)}, ( \sigma_{a-1}^{(n)} )^2 \big)
\end{equation*}

Thus to show that $\pi_a^{(n)} \Pto \pi_{\max}$ for all $a \in [2 \colon T]$, it is sufficient to show that $\mu_{a-1}^{(n)} \Pto \Delta$ and $( \sigma_{a-1}^{(n)} )^2 \Pto 0$ for all $a \in [2 \colon T]$.
By Lemma \ref{lemma:ratio}, for all $t \in [1 \colon T]$, 
\begin{equation*}
\frac{N_{t,1}}{\pi_t^{(n)} n} \Pto 1
\end{equation*}

% Sigma
Thus, to show $( \sigma_{a-1}^{(n)} )^2 \Pto 0$, it is sufficient to show that 
\begin{equation*}
\frac{ \sigma^2 (\sigma^2 + n \sum_{t=1}^{a-1} \pi_t^{(n)}) + \sigma^2 ( \sigma^2 + n \sum_{t=1}^{a-1} (1-\pi_t^{(n)}) ) }{( \sigma^2 + n \sum_{t=1}^{a-1} (1- \pi_t^{(n)}) )( \sigma^2 + n \sum_{t=1}^{a-1} \pi_t^{(n)} )} 
\Pto 0
\end{equation*}
The above limit holds because $\pi_t^{(n)} \in [ \pi_{\min}, \pi_{\max} ]$ for $0 < \pi_{\min} \leq \pi_{\max} < 1$ by the clipping condition.

% mu
We now show that $\mu_{a-1}^{(n)} \Pto \Delta$, which is equivalent to showing that the following converges in probability to $\Delta$
\begin{equation*}
\frac{ \sum_{t=1}^{a-1} \sum_{i=1}^n A_{t,i} R_{t,i} }{\sigma^2 + \sum_{t=1}^{a-1} N_{t,1}}
- \frac{ \sum_{t=1}^{a-1} \sum_{i=1}^n (1-A_{t,i}) R_{t,i} }{\sigma^2 + \sum_{t=1}^{a-1} N_{t,0}}
\end{equation*}
\begin{equation*}
= \frac{ \sum_{t=1}^{a-1} N_{t,1} }{\sigma^2 + \sum_{t=1}^{a-1} N_{t,1}} \frac{ \sum_{t=1}^{a-1} \sum_{i=1}^n A_{t,i} R_{t,i} }{ \sum_{t=1}^{a-1} N_{t,1}}
- \frac{ \sum_{t=1}^{a-1} N_{t,0} }{\sigma^2 + \sum_{t=1}^{a-1} N_{t,0}} \frac{ \sum_{t=1}^{a-1} \sum_{i=1}^n (1-A_{t,i}) R_{t,i} }{ \sum_{t=1}^{a-1} N_{t,0}}
\end{equation*}
\begin{multline}
\label{eqn:tsmu}
= \frac{ \sum_{t=1}^{a-1} N_{t,1} }{\sigma^2 + \sum_{t=1}^{a-1} N_{t,1}} \bigg( \beta_1 + \frac{ \sum_{t=1}^{a-1} \sum_{i=1}^n A_{t,i} \epsilon_{t,i} }{ \sum_{t=1}^{a-1} N_{t,1}} \bigg) \\
- \frac{ \sum_{t=1}^{a-1} N_{t,0} }{\sigma^2 + \sum_{t=1}^{a-1} N_{t,0}} \bigg( \beta_0 + \frac{ \sum_{t=1}^{a-1} \sum_{i=1}^n (1-A_{t,i}) \epsilon_{t,i} }{ \sum_{t=1}^{a-1} N_{t,0}} \bigg)
\end{multline}

Note that 
\begin{equation}
\label{eqn:tsdelta}
\frac{ \sum_{t=1}^{a-1} N_{t,1} }{\sigma^2 + \sum_{t=1}^{a-1} N_{t,1}} \beta_1 - \frac{ \sum_{t=1}^{a-1} N_{t,0} }{\sigma^2 + \sum_{t=1}^{a-1} N_{t,0}} \beta_0 \Pto \Delta
\end{equation}
Equation \eqref{eqn:tsdelta} above holds  by Lemma \ref{lemma:ratio}, because
\begin{equation}
\label{eqn:nratio}
\frac{ n \sum_{t=1}^{a-1} \pi_t^{(n)} }{\sigma^2 + n \sum_{t=1}^{a-1} \pi_t^{(n)} } \Pto 1 ~~~~~~~~~~~~~~~~~~~~~~~~~~~~~~~~~~ 
\frac{ n \sum_{t=1}^{a-1} (1-\pi_t^{(n)}) }{\sigma^2 + n \sum_{t=1}^{a-1} (1-\pi_t^{(n)})} \Pto 1
\end{equation}
which hold because $\pi_t^{(n)} \in [ \pi_{\min}, \pi_{\max} ]$ due to our clipping condition.

By Slutsky's Theorem and \eqref{eqn:tsdelta}, to show \eqref{eqn:tsmu}, it is sufficient to show that
\begin{equation}
\label{eqn:tszero} 
\frac{ \sum_{t=1}^{a-1} \sum_{i=1}^n A_{t,i} \epsilon_{t,i} }{ \sigma^2 + \sum_{t=1}^{a-1} N_{t,1} }
- \frac{ \sum_{t=1}^{a-1} \sum_{i=1}^n (1-A_{t,i}) \epsilon_{t,i} }{ \sigma^2 + \sum_{t=1}^{a-1} N_{t,0} } \Pto 0.
\end{equation}

Equation \eqref{eqn:tszero} is equivalent to the following:
\begin{equation}
\label{eqn:tszero2} 
\sum_{t=1}^{a-1} \frac{ \sqrt{ N_{t,1} } }{\sigma^2 + \sum_{t'=1}^{a-1} N_{t',1}} \frac{ \sum_{i=1}^n A_{t,i} \epsilon_{t,i} }{ \sqrt{ N_{t,1} } }
- \sum_{t=1}^{a-1} \frac{ \sqrt{N_{t,0}} }{\sigma^2 + \sum_{t'=1}^{a-1} N_{t',0}} \frac{ \sum_{i=1}^n (1-A_{t,i}) \epsilon_{t,i} }{ \sqrt{N_{t,0}} } \Pto 0
\end{equation}

By Lemma \ref{lemma:ratio}, to show \eqref{eqn:tszero2} it is sufficient to show that
\small
\begin{equation}
\label{eqn:tszero3} 
\sum_{t=1}^{a-1} \frac{ \sqrt{ n \pi_t^{(n)} } }{\sigma^2 + n \sum_{t'=1}^{a-1} \pi_{t'}^{(n)} } \frac{ \sum_{i=1}^n A_{t,i} \epsilon_{t,i} }{ \sqrt{ N_{t,1} } }
- \sum_{t=1}^{a-1} \frac{ \sqrt{ n (1-\pi_t^{(n)}) } }{\sigma^2 + n \sum_{t'=1}^{a-1} (1-\pi_{t'}^{(n)})} \frac{ \sum_{i=1}^n (1-A_{t,i}) \epsilon_{t,i} }{ \sqrt{N_{t,0}} } \Pto 0
\end{equation}
\normalsize

Since $\pi_t^{(n)} \in [ \pi_{\min}, \pi_{\max} ]$ due to our clipping condition, the left hand side of \eqref{eqn:tszero3} equals the following
\begin{equation*}
\sum_{t=1}^{a-1} o_p(1) \frac{ \sum_{i=1}^n A_{t,i} \epsilon_{t,i} }{ \sqrt{ N_{t,1} } }
- \sum_{t=1}^{a-1} o_p(1) \frac{ \sum_{i=1}^n (1-A_{t,i}) \epsilon_{t,i} }{ \sqrt{N_{t,0}} } \Pto 0
\end{equation*}
The above limit holds by \eqref{eqn:bolsresult}.

Thus, by Slutsky's Theorem and Lemma \ref{lemma:ratio}, we have that
\begin{equation*}
\frac{1}{n} \sum_{t=1}^T N_{t,1} \Pto \frac{1}{2} + (T-1) \pi_{\max}
~~~~~~~~~\TN{ and }~~~~~~~~~~
\frac{1}{n} \sum_{t=1}^T N_{t,0} \Pto \frac{1}{2} + (T-1) \pi_{\min} ~~~ \qed
\end{equation*}

\paragraph{UCB}
We assume without loss of generality that $\Delta > 0$ and $\pi_1^{(n)} = \frac{1}{2}$. 
Recall that for UCB, for $a \in [2 \colon T]$,
\begin{equation*}
\pi_a^{(n)} 
= \begin{cases}
		\pi_{\max} 		& \TN{ if } U_{a-1,1} > U_{a-1,0} \\
		1 -  \pi_{\max}		& \TN{ otherwise } \\
\end{cases}
\end{equation*}
where we define the upper confidence bounds $U$ for any confidence level $\delta$ with $0 < \delta < 1$ as follows:
\begin{equation*}
	U_{a-1,1} = \begin{cases} 
			\infty & \TN{ if } \sum_{t=1}^{a-1} N_{t,1} = 0 \\
			\frac{ \sum_{t=1}^{a-1} \sum_{i=1}^n A_{t,i} R_{t,i} }{ \sum_{t=1}^{a-1} N_{t,1} } + \sqrt{ \frac{2 \log 1/\delta}{ \sum_{t=1}^{a-1} N_{t,1} } } & \TN{ otherwise } 
		\end{cases}
\end{equation*}
\begin{equation*}
	U_{a-1,0} = \begin{cases} 
			\infty & \TN{ if } N_{1,0} = 0 \\
			\frac{ \sum_{t=1}^{a-1} \sum_{i=1}^n (1-A_{t,i}) R_{t,i} }{ \sum_{t=1}^{a-1} N_{t,0} } + \sqrt{ \frac{2 \log 1/\delta}{ \sum_{t=1}^{a-1} N_{t,0} } } & \TN{ otherwise } 
		\end{cases}
\end{equation*}

Thus to show that $\pi_a^{(n)} \Pto \pi_{\max}$ for all $a \in [2 \colon T]$, it is sufficient to show that $\II_{ ( U_{a,1} > U_{a,0} ) } \Pto 1$, which is equivalent to showing that the following converges in probability to $1$:
\begin{multline*}
	\II_{ ( \sum_{t=1}^a N_{t,1} > 0, \sum_{t=1}^a N_{t,0} > 0) } \II_{ \big( \frac{ \sum_{t=1}^a \sum_{i=1}^n A_{t,i} R_{t,i} }{ \sum_{t=1}^a N_{t,1} } + \sqrt{ \frac{2 \log 1/\delta}{ \sum_{t=1}^a N_{t,1} } } 
	> \frac{ \sum_{t=1}^a \sum_{i=1}^n (1-A_{t,i}) R_{t,i} }{ \sum_{t=1}^a N_{t,1} } + \sqrt{ \frac{2 \log 1/\delta}{ \sum_{t=1}^a N_{t,0} } } \big) } \\
	+ \II_{ (\sum_{t=1}^a N_{t,1} = 0, \sum_{t=1}^a N_{t,0} > 0) }
\end{multline*}
\begin{equation*}
	= \II_{ \big( (\beta_1 - \beta_0) + \frac{ \sum_{t=1}^a \sum_{i=1}^n A_{t,i} \epsilon_{t,i} }{ \sum_{t=1}^a N_{t,1} } + \sqrt{ \frac{2 \log 1/\delta}{ \sum_{t=1}^a N_{t,1} } } 
	> \frac{ \sum_{t=1}^a \sum_{i=1}^n (1-A_{t,i}) \epsilon_{t,i} }{ \sum_{t=1}^a N_{t,1} } + \sqrt{ \frac{2 \log 1/\delta}{ \sum_{t=1}^a N_{t,0} } } \big) } 
	+ o_p(1)
\end{equation*}
Note that to show that the above converges in probability to $1$, it is sufficient to show that the following:
\begin{equation*}
	\frac{ \sum_{t=1}^a \sum_{i=1}^n (1-A_{t,i}) \epsilon_{t,i} }{ \sum_{t=1}^a N_{t,1} } + \sqrt{ \frac{2 \log 1/\delta}{ \sum_{t=1}^a N_{t,0} } }
	- \frac{ \sum_{t=1}^a \sum_{i=1}^n A_{t,i} \epsilon_{t,i} }{ \sum_{t=1}^a N_{t,1} } - \sqrt{ \frac{2 \log 1/\delta}{ \sum_{t=1}^a N_{t,1} } }  \Pto 0
\end{equation*}
Note that for fixed $\delta$, we have that $\frac{2 \log 1/\delta}{ \sum_{t=1}^a N_{t,0} } \Pto 0$, since $\frac{ N_{t,0} }{ n/2 } \Pto 1$.
Also note that $\frac{ \sum_{t=1}^a \sum_{i=1}^n (1-A_{t,i}) \epsilon_{t,i} }{ \sum_{t=1}^a N_{t,1} } - \frac{ \sum_{t=1}^a \sum_{i=1}^n A_{t,i} \epsilon_{t,i} }{ \sum_{t=1}^a N_{t,1} } \Pto 0$, by the same argument made in the $\epsilon$-greedy case to show \eqref{eqn:epsilonprob3}.

Thus, by Slutsky's Theorem and Lemma \ref{lemma:ratio}, we have that
\begin{equation*}
\frac{1}{n} \sum_{t=1}^T N_{t,1} \Pto \frac{1}{2} + (T-1) \pi_{\max}
~~~~~~~~~\TN{ and }~~~~~~~~~~
\frac{1}{n} \sum_{t=1}^T N_{t,0} \Pto \frac{1}{2} + (T-1) (1-\pi_{\max})
\end{equation*}

%% file: appendix/nonnormality.tex
\section{Non-uniform convergence of the OLS Estimator}
\label{appendix:nonnormality}

\begin{mydef}[Non-concentration of a sequence of random variables]
For a sequence of random variables $\{ Y_i \}_{i=1}^n$ on probability space $( \Omega, \F, \mathbb{P} )$, we say $Y_n$ \textit{does not concentrate} if for each $a \in \real$ there exists an $\epsilon_a > 0$ with
\begin{equation*}
P \big( \big\{ \omega \in \Omega : \big| Y_n ( \omega ) - a \big| > \epsilon_a \big\} \big) \not\to 0.
\end{equation*}
\end{mydef}

\subsection{Thompson Sampling}

\begin{proposition}[Non-concentration of sampling probabilities under Thompson Sampling]
\label{prop:nonconcentrationTS}
Under the assumptions of Theorem \ref{thm:nonnormTS}, the posterior distribution that arm $1$ is better than arm $0$ converges as follows:
\begin{equation*}
\PP \big( \tilde{\beta}_1 > \tilde{\beta}_0 ~\big|~ H_1^{(n)} \big)
\Dto
\begin{cases}
	1 & \TN{ if } \Delta > 0 \\
	0 & \TN{ if } \Delta < 0 \\
	\TN{Uniform}[0, 1] & \TN{ if } \Delta = 0
\end{cases}
\end{equation*}
Thus, the sampling probabilities $\pi_t^{(n)}$ do not concentrate when $\Delta = 0$.
\end{proposition}

\paragraph{Proof:}
Below, $N_{t,1} = \sum_{i=1}^n A_{t,i}$ and $N_{t,0} = \sum_{i=1}^n (1-A_{t,i})$. Posterior means:
$$\tilde{\beta}_0 | H_1^{(n)} 
\sim \N \bigg( 
\frac{ \sum_{i=1}^n (1-A_{1,i}) R_{1,i} }{\sigma_a^2 + N_{1,0} }, 
\frac{ \sigma^2}{\sigma_a^2 + N_{0, 1}} \bigg)$$
$$\tilde{\beta}_1  | H_1^{(n)} 
\sim \N \bigg( 
\frac{ \sum_{i=1}^n A_{1,i} R_{1,i} }{\sigma_a^2 + N_{1,1} }, 
\frac{ \sigma_a^2}{\sigma_a^2 + N_{1,1} } \bigg)$$
%%%%%%%%%5
$$\tilde{\beta}_1 - \tilde{\beta}_0 ~|~ H_1^{(n)} 
\sim \N( \mu_n, \sigma^2_n )$$
for $\mu_n := \frac{ \sum_{i=1}^n A_{1,i} R_{1,i} }{\sigma_a^2 + N_{1,1}}
- \frac{ \sum_{i=1}^n (1-A_{1,i}) R_{1,i} }{\sigma_a^2 + N_{1,0}}$ and $\sigma^2_n := \frac{ \sigma_a^2 (\sigma_a^2 + N_{1,1}) + \sigma_a^2 ( \sigma_a^2 + N_{1,0}) }{( \sigma_a^2 + N_{1,0})( \sigma_a^2 + N_{1,1})}$.
%\N \bigg( \frac{ \sum_{i=1}^n A_{1,i} R_{1,i} }{\sigma_a^2 + N_{1,1}}
%- \frac{ \sum_{i=1}^n (1-A_{1,i}) R_{1,i} }{\sigma_a^2 + N_{1,0}},
%\frac{ \sigma_a^2 (\sigma_a^2 + N_{1,1}) + \sigma_a^2 ( \sigma_a^2 + N_{1,0}) }{( \sigma_a^2 + N_{1,0})( \sigma_a^2 + N_{1,1})} \bigg)
%=: \N( \mu_n, \sigma^2_n )$$
%\normalsize
$$P(\tilde{\beta}_1 > \tilde{\beta}_0 ~|~ H_1^{(n)} )
= P( \tilde{\beta}_1 - \tilde{\beta}_0 > 0 ~|~ H_1^{(n)} )
= P \bigg( \frac{ \tilde{\beta}_1 - \tilde{\beta}_0 - \mu_n }{ \sigma_n } > - \frac{ \mu_n }{ \sigma_n } ~\bigg|~ H_1^{(n)} \bigg)$$
For $Z \sim \N(0, 1)$ independent of $\mu_n, \sigma_n$.
$$= P \bigg( Z > - \frac{ \mu_n }{ \sigma_n } ~\bigg|~ H_1^{(n)} \bigg)
= P \bigg( Z < \frac{ \mu_n }{ \sigma_n } ~\bigg|~ H_1^{(n)} \bigg)
= \Phi \bigg( \frac{ \mu_n }{ \sigma_n } ~\bigg|~ H_1^{(n)} \bigg)$$
%%%%%%%%%%%%%%%%%%%%%%%%%%%%%%%%%%%%%%%%%%%%%%%%%%%%%%
$$\frac{ \mu_n }{ \sigma_n }
= \bigg( \frac{ \sum_{i=1}^n A_{1,i} R_{1,i} }{\sigma_a^2 + N_{1,1}} - \frac{ \sum_{i=1}^n (1-A_{1,i}) R_{1,i} }{\sigma_a^2 + N_{1,0}} \bigg) 
\sqrt{ \frac{ ( \sigma_a^2 + N_{1,0} )( \sigma_a^2 + N_{1,1}) }{ 2 \sigma_a^4 + \sigma_a^2 n } }$$
$$= \bigg( \frac{ \beta_1 N_{1,1} + \sum_{i=1}^n A_{1,i} \epsilon_{1,i} }{\sigma_a^2 + N_{1,1}} 
- \frac{ \beta_0 N_{1,0} + \sum_{i=1}^n (1-A_{1,i}) \epsilon_{1,i} }{\sigma_a^2 + N_{1,0}} \bigg) 
\sqrt{ \frac{ ( \sigma_a^2 + N_{1,0} )( \sigma_a^2 + N_{1,1}) }{ 2 \sigma_a^4 + \sigma_a^2 n } }$$
$$= \frac{ \sum_{i=1}^n A_{1,i} \epsilon_{1,i} }{ \sqrt{N_{1,1} } } 
\sqrt{ \frac{ N_{1,1} ( \sigma_a^2 + N_{1,0} ) }{ ( 2 \sigma_a^4 + \sigma_a^2 n )( \sigma_a^2 + N_{1,1}) } }
- \frac{ \sum_{i=1}^n (1-A_{1,i}) \epsilon_{1,i} }{ \sqrt{ N_{1,0} } } 
\sqrt{ \frac{ N_{1,0} ( \sigma_a^2 + N_{1,1}) }{ ( 2 \sigma_a^4 + \sigma_a^2 n ) ( \sigma_a^2 + N_{1,0} ) } }$$
$$+ \bigg( \beta_1 \frac{  N_{1,1} }{\sigma_a^2 + N_{1,1}} - \beta_0 \frac{ N_{1,0} }{\sigma_a^2 + N_{1,0}} \bigg) 
\sqrt{ \frac{ ( \sigma_a^2 + N_{1,0} )( \sigma_a^2 + N_{1,1}) }{ 2 \sigma_a^4 + \sigma_a^2 n } } =: B_n + C_n$$
%%%%%%%%%%%%%%%%%%%%%%%%%%%
Let's first examine $C_n$.
Note that $\beta_1 = \beta_0 + \Delta$, so $\beta_1 \frac{  N_{1,1} }{\sigma_a^2 + N_{1,1}} - \beta_0 \frac{ N_{1,0} }{\sigma_a^2 + N_{1,0}}$ equals
$$= (\beta_0 + \Delta) \frac{  N_{1,1} }{\sigma_a^2 + N_{1,1}} - \beta_0 \frac{ N_{1,0} }{\sigma_a^2 + N_{1,0}}
= \Delta \frac{  N_{1,1} }{\sigma_a^2 + N_{1,1}} + \beta_0 \bigg( \frac{  N_{1,1} }{\sigma_a^2 + N_{1,1}} - \frac{ N_{1,0} }{\sigma_a^2 + N_{1,0}} \bigg)$$
$$= \Delta \frac{ N_{1,1} / n }{ (\sigma_a^2 + N_{1,1} ) / n } + 
\beta_0 \bigg( \frac{  N_{1,1} (\sigma_a^2 + N_{1,0}) - N_{1,0} (\sigma_a^2 + N_{1,1}) }{ ( \sigma_a^2 + N_{1,1} ) (\sigma_a^2 + N_{1,1}) } \bigg)$$
$$= \Delta \frac{ \frac{1}{2} + o(1) }{ \frac{1}{2} + o(1) } + 
\beta_0  \sigma_a^2  \bigg( \frac{  N_{1,1}- N_{1,0} }{ ( \sigma_a^2 + N_{1,1} ) (\sigma_a^2 + N_{1,1}) } \bigg)
=  \Delta [ 1 + o(1) ] + o \bigg( \frac{1}{n} \bigg)$$
where the last equality holds by the Strong Law of Large Numbers because 
$$\frac{  \frac{1}{n^2} ( N_{1,1}- N_{1,0} ) }{ \frac{1}{n^2} ( \sigma_a^2 + N_{1,1} ) (\sigma_a^2 + N_{1,1}) }
= \frac{  \frac{1}{n} [ \frac{1}{2} - \frac{1}{2} + o(1) ] }{ [ \frac{1}{2} + o(1) ] [ \frac{1}{2} + o(1) ] }
= \frac{  \frac{1}{n} o(1) }{ \frac{1}{4} + o(1) } = o \bigg( \frac{1}{n} \bigg)$$
%%%%%%%%%%%%%%%%%%%%%%%%%%%
Thus, 
$$C_n = \bigg[ \Delta [ 1 + o(1) ] + o \bigg( \frac{1}{n} \bigg) \bigg] \sqrt{ \frac{ ( \sigma_a^2 + N_{1,0} )( \sigma_a^2 + N_{1,1}) }{ 2 \sigma_a^4 + \sigma_a^2 n } }$$
$$= \bigg[ \Delta [ 1 + o(1) ] + o \bigg( \frac{1}{n} \bigg) \bigg] \sqrt{ \frac{ n [ \frac{1}{2} + o(1) ] [ \frac{1}{2} + o(1) ]  }{ o(1) + \sigma_a^2 } }
=  \sqrt{n} \Delta \big[ 1/ (2 \sigma_a) + o(1) \big] + o \bigg( \frac{1}{ \sqrt{n} } \bigg)$$
%%%%%%%%%%%%%%%%%%%%%%%%%%%
Let's now examine $B_n$. 
$$\sqrt{ \frac{ N_{1,1} ( \sigma_a^2 + N_{1,0} ) }{ ( 2 \sigma_a^4 + \sigma_a^2 n )( \sigma_a^2 + N_{1,1}) } } 
= \sqrt{ \frac{ [ \frac{1}{2} + o(1) ] [ \frac{1}{2} + o(1) ] }{ [ \sigma_a^2 + o(1) ][ \frac{1}{2} + o(1) ] } }
= \sqrt{ \frac{ 1 }{2  \sigma_a^2} } + o(1)$$
$$\sqrt{ \frac{ N_{1,0} ( \sigma_a^2 + N_{1,1}) }{ ( 2 \sigma_a^4 + \sigma_a^2 n ) ( \sigma_a^2 + N_{1,0} ) } }
= \sqrt{ \frac{ [ \frac{1}{2} + o(1) ] [ \frac{1}{2} + o(1) ] }{ [ \sigma_a^2 + o(1) ][ \frac{1}{2} + o(1) ] } }
= \sqrt{ \frac{ 1 }{2 \sigma_a^2 } } + o(1)$$
Note that by Theorem \ref{thm:bols}, $\bigg[ \frac{1}{\sqrt{ N_{1,1} }} \sum_{i=1}^{n} \epsilon_{1,i} A_{1,i}, \frac{1}{\sqrt{ N_{1,0} }} \sum_{i=1}^{n} \epsilon_{1,i} (1-A_{1,i}) \bigg] \Dto \N ( \bs{0}, \under{\bo{I}}_2 )$.
Thus by Slutky's Theorem,
$$\begin{bmatrix}
\frac{ \sum_{i=1}^n A_{1,i} \epsilon_{1,i} }{ \sqrt{N_{1,1} } } 
	\sqrt{ \frac{ N_{1,1} ( \sigma_a^2 + N_{1,0} ) }{ ( 2 \sigma_a^4 + \sigma_a^2 n )( \sigma_a^2 + N_{1,1}) } } \\
\frac{ \sum_{i=1}^n (1-A_{1,i}) \epsilon_{1,i} }{ \sqrt{ N_{1,0} } } 
	\sqrt{ \frac{ N_{1,0} ( \sigma_a^2 + N_{1,1}) }{ ( 2 \sigma_a^4 + \sigma_a^2 n ) ( \sigma_a^2 + N_{1,0} ) } }
\end{bmatrix}
= \begin{bmatrix} 
\frac{ \sum_{i=1}^n A_{1,i} \epsilon_{1,i} }{ \sqrt{N_{1,1} } } \big[ \sqrt{ \frac{ 1 }{2 \sigma_a^2 } } + o(1) \big] \\
\frac{ \sum_{i=1}^n (1-A_{1,i}) \epsilon_{1,i} }{ \sqrt{ N_{1,0} } } \big[ \sqrt{ \frac{ 1 }{2 \sigma_a^2 } } + o(1) \big]
\end{bmatrix}
\Dto \N \bigg( \bs{0}, \frac{1}{2  \sigma_a^2 } \under{\bo{I}}_2 \bigg)$$
Thus, we have that, $B_n \Dto \N \big( 0, \frac{1}{ \sigma_a^2} \big)$. 
Since we assume that the algorithm's variance is correctly specified, so $\sigma_a^2 = 1$, 
$$B_n + C_n \Dto 
\begin{cases}
	\infty & \TN{ if } \Delta > 0 \\
	-\infty & \TN{ if } \Delta < 0 \\
	\N(0, 1) & \TN{ if } \Delta = 0
\end{cases}$$
Thus, by continuous mapping theorem,
$$\PP \big( \tilde{\beta}_1 > \tilde{\beta}_0 \big| H_1^{(n)} \big)
= \Phi \bigg( \frac{ \mu_n }{ \sigma_n } \bigg) = \Phi( B_n + C_n ) \Dto 
\begin{cases}
	1 & \TN{ if } \Delta > 0 \\
	0 & \TN{ if } \Delta < 0 \\
	\TN{Uniform}[0, 1] & \TN{ if } \Delta = 0
\end{cases} \qed$$

\paragraph{Proof of Theorem \ref{thm:nonnormTS} (Non-uniform convergence of the OLS estimator of the treatment effect for Thompson Sampling):}
The normalized errors of the OLS estimator for $\Delta$, which are asymptotically normal under i.i.d. sampling are as follows:
$$\sqrt{ \frac{ (N_{1,1} + N_{2,1}) (N_{1,0} + N_{2,0}) }{ 2 n } } \bigg( \betahat_1^{\OLS} - \betahat_0^{\OLS} - \Delta \bigg)$$
$$= \sqrt{ \frac{ (N_{1,1} + N_{2,1}) (N_{1,0} + N_{2,0}) }{ 2 n } } \bigg( \frac{ \sum_{t=1}^2 \sum_{i=1}^n A_{t,i} R_{t,i}  }{ N_{1,1} + N_{2,1} } 
- \frac{ \sum_{t=1}^2 \sum_{i=1}^n (1-A_{t,i}) R_{t,i}}{ N_{1,0} + N_{2,0} } - \Delta \bigg)$$
$$= \sqrt{ \frac{ (N_{1,1} + N_{2,1}) (N_{1,0} + N_{2,0}) }{ 2 n } } 
\bigg( (\beta_1 - \beta_0 ) - \Delta + \frac{ \sum_{t=1}^2 \sum_{i=1}^n A_{t,i} \epsilon_{t,i}  }{ N_{1,1} + N_{2,1} } - \frac{ \sum_{t=1}^2 \sum_{i=1}^n (1-A_{t,i}) \epsilon_{t,i}}{ N_{1,0} + N_{2,0} } \bigg)$$
$$= \sqrt{ \frac{ N_{1,0} + N_{2,0} }{ 2 n } } \frac{ \sum_{t=1}^2 \sum_{i=1}^n A_{t,i} \epsilon_{t,i}  }{ \sqrt{ N_{1,1} + N_{2,0} } } 
- \sqrt{ \frac{ N_{1,1} + N_{2,1} }{ 2 n } } \frac{ \sum_{t=1}^2 \sum_{i=1}^n (1-A_{t,i}) \epsilon_{t,i}}{ \sqrt{ N_{1,0} + N_{2,0} } }$$
\begin{equation*}
= [1, -1, 1, -1] \begin{bmatrix}
	\vspace{1mm}
	\sqrt{ \frac{ N_{1,0} + N_{2,0} }{ 2 n } }  \frac{ \sum_{i=1}^n A_{1,i} \epsilon_{1,i} }{ \sqrt{ N_{1,1} + N_{2,1} } } \\
	\vspace{1mm}
	\sqrt{ \frac{ N_{1,1} + N_{2,1} }{ 2 n } }\frac{ \sum_{i=1}^n (1-A_{1,i}) \epsilon_{1,i} }{ \sqrt{ N_{1,0} + N_{2,0} } } \\
	\vspace{1mm}
	\sqrt{ \frac{ N_{1,0} + N_{2,0} }{ 2 n } } \frac{ \sum_{i=1}^n A_{2,i} \epsilon_{2,i} }{ \sqrt{ N_{1,1} + N_{2,1} } } \\
	\sqrt{ \frac{ N_{1,1} + N_{2,1} }{ 2 n } } \frac{ \sum_{i=1}^n (1-A_{2,i}) \epsilon_{2,i} }{ \sqrt{ N_{1,0} + N_{2,0} } }
\end{bmatrix}
\end{equation*}
\begin{equation}
\label{eqn:normalizederrors}
= [1, -1, 1, -1] \begin{bmatrix}
	\vspace{1mm}
	\sqrt{ \frac{ N_{1,0} + N_{2,0} }{ 2 ( N_{1,1} + N_{2,1} ) } } \sqrt{ \frac{ N_{1,1} }{n} } \frac{ \sum_{i=1}^n A_{1,i} \epsilon_{1,i} }{ \sqrt{ N_{1,1} } } \\
	\vspace{1mm}
	\sqrt{ \frac{ N_{1,1} + N_{2,1} }{ 2 ( N_{1,0} + N_{2,0} ) } } \sqrt{ \frac{ N_{1,0} }{n} }\frac{ \sum_{i=1}^n (1-A_{1,i}) \epsilon_{1,i} }{ \sqrt{ N_{1,0} } } \\
	\vspace{1mm}
	\sqrt{ \frac{ N_{1,0} + N_{2,0} }{ 2 ( N_{1,1} + N_{2,1} ) } } \sqrt{ \frac{ N_{2,1} }{n} } \frac{ \sum_{i=1}^n A_{2,i} \epsilon_{2,i} }{ \sqrt{N_{2,1} } } \\
	\sqrt{ \frac{ N_{1,1} + N_{2,1} }{ 2 ( N_{1,0} + N_{2,0} ) } } \sqrt{ \frac{ N_{2,0} }{n} } \frac{ \sum_{i=1}^n (1-A_{2,i}) \epsilon_{2,i} }{ \sqrt{ N_{2,0} } }
\end{bmatrix}
\end{equation}
%%%%%%%%%%%%%%%%%%%%%%%%%%%%%%%%%%
By Theorem \ref{thm:bols},
$\bigg( \frac{ \sum_{i=1}^n A_{1,i} \epsilon_{1,i} }{ \sqrt{ N_{1,1} } }, 
\frac{ \sum_{i=1}^n (1-A_{1,i}) \epsilon_{1,i} }{ \sqrt{ N_{1,0} } }, 
\frac{ \sum_{i=1}^n A_{2,i} \epsilon_{2,i} }{ \sqrt{N_{2,1} } },
\frac{ \sum_{i=1}^n (1-A_{2,i}) \epsilon_{2,i} }{ \sqrt{N_{2,0} } }
\bigg) \Dto \N( \bs{0}, \under{\bo{I}}_4 )$.
%%%%%%%%%%%%%%%%%%%%%%%%%%%%%%%%%%
By Lemma \ref{lemma:ratio} and Slutsky's Theorem, 
$\sqrt{ \frac{ 2 n ( N_{1,1} + N_{2,1} ) }{  N_{1,1} ( N_{1,0} + N_{2,0} ) } } \sqrt{ \frac{ \frac{1}{2} ( \frac{1}{2}  + [1-\pi_2] ) }{ 2 (  \frac{1}{2}  + \pi_2 ) } } = 1 + o_p(1)$,
thus, 
$$\sqrt{ \frac{ N_{1,0} + N_{2,0} }{ 2 ( N_{1,1} + N_{2,1} ) } } \sqrt{ \frac{ N_{1,1} }{n} } \frac{ \sum_{i=1}^n A_{1,i} \epsilon_{1,i} }{ \sqrt{ N_{1,1} } }$$
$$= \bigg( \sqrt{ \frac{ 2 n ( N_{1,1} + N_{2,1} ) }{  N_{1,1} ( N_{1,0} + N_{2,0} ) } } \sqrt{ \frac{ \frac{1}{2} ( \frac{1}{2}  + [1-\pi_2] ) }{ 2 (  \frac{1}{2}  + \pi_2 ) } } + o_p(1) \bigg) 
\sqrt{ \frac{ N_{1,0} + N_{2,0} }{ 2 ( N_{1,1} + N_{2,1} ) } } \sqrt{ \frac{ N_{1,1} }{n} } \frac{ \sum_{i=1}^n A_{1,i} \epsilon_{1,i} }{ \sqrt{ N_{1,1} } }$$
$$= \sqrt{ \frac{ \frac{1}{2} ( \frac{1}{2}  + [1-\pi_2] ) }{ 2 (  \frac{1}{2}  + \pi_2 ) } }  \frac{ \sum_{i=1}^n A_{1,i} \epsilon_{1,i} }{ \sqrt{ N_{1,1} } }
+ o_p(1) \sqrt{ \frac{ N_{1,0} + N_{2,0} }{ 2 ( N_{1,1} + N_{2,1} ) } } \sqrt{ \frac{ N_{1,1} }{n} } \frac{ \sum_{i=1}^n A_{1,i} \epsilon_{1,i} }{ \sqrt{ N_{1,1} } }$$
%%%%%%%%%%%%%%%%%%%%%%%%%%%%%%%%%%
Note that $\sqrt{ \frac{ N_{1,0} + N_{2,0} }{ 2 ( N_{1,1} + N_{2,1} ) } }$ is stochastically bounded because for any $K > 2$,
$$\PP \bigg( \frac{ N_{1,0} + N_{2,0} }{ 2 ( N_{1,1} + N_{2,1} ) } > K \bigg)
\leq \PP \bigg( \frac{ n }{ N_{1,1} } > K \bigg)
= \PP \bigg( \frac{1}{K} > \frac{ N_{1,1} }{ n } \bigg) \to 0$$
where the limit holds by the law of large numbers since $N_{1,1}^{(n)} \sim \TN{Binomial}( n, \frac{1}{2} )$.
Thus, since $\frac{ N_{1,1} }{n} \leq 1$ and $\frac{ \sum_{i=1}^n A_{1,i} \epsilon_{1,i} }{ \sqrt{ N_{1,1} } } \Dto \N( 0, 1 )$,
$$o_p(1) \sqrt{ \frac{ N_{1,0} + N_{2,0} }{ 2 ( N_{1,1} + N_{2,1} ) } } \sqrt{ \frac{ N_{1,1} }{n} } \frac{ \sum_{i=1}^n A_{1,i} \epsilon_{1,i} }{ \sqrt{ N_{1,1} } } = o_p(1)$$
We can perform the above procedure on the other three terms.
Thus, equation \eqref{eqn:normalizederrors} is equal to the following:
$$[1, -1, 1, -1] \begin{bmatrix}
	\vspace{1mm}
	\sqrt{ \frac{ 1/2 + 1 - \pi_2 }{ 4 ( 1/2 + \pi_2 ) } } \frac{ \sum_{i=1}^n A_{1,i} \epsilon_{1,i} }{ \sqrt{ N_{1,1} } } \\
	\vspace{1mm}
	\sqrt{ \frac{ 1/2 + \pi_2 }{ 4 ( 1/2 + 1 - \pi_2 ) } } \frac{ \sum_{i=1}^n (1-A_{1,i}) \epsilon_{1,i} }{ \sqrt{ N_{1,0} } } \\
	\vspace{1mm}
	\sqrt{ \frac{ ( 1/2 + 1 - \pi_2 ) \pi_2 }{ 2 ( 1/2 + \pi_2) } } \frac{ \sum_{i=1}^n A_{2,i} \epsilon_{2,i} }{ \sqrt{N_{2,1} } } \\
	\sqrt{ \frac{ ( 1/2 + \pi_2 ) (1 - \pi_2 ) }{ 2 ( 1/2 + 1 - \pi_2 ) } } \frac{ \sum_{i=1}^n (1-A_{2,i}) \epsilon_{2,i} }{ \sqrt{ N_{2,0} } }
\end{bmatrix} + o_p(1)$$
%%%%%%%%%%%%%%%%%%%%%%%%%%%%%%%%%%
Recall that we showed earlier in Proposition \ref{prop:nonconcentrationTS} that
$$\pi_2^{(n)} = \pi_{\min} \vee \bigg[ \pi_{\max} \wedge \Phi \bigg( \frac{ \mu_n }{ \sigma_n } \bigg) \bigg]
= \pi_{\min} \vee \bigg[ \pi_{\max} \wedge \Phi \bigg(  B_n + C_n \bigg) \bigg]$$
$$= \pi_{\min} \vee \bigg[ \pi_{\max} \wedge \Phi \bigg( 
\frac{ \sum_{i=1}^n A_{1,i} \epsilon_{1,i} }{ \sqrt{2 N_{1,1} } } 
- \frac{ \sum_{i=1}^n (1-A_{1,i}) \epsilon_{1,i} }{ \sqrt{ 2 N_{1,0} } } 
+ \sqrt{n} \Delta \bigg[ \frac{1}{2} + o(1) \bigg] + o(1)
 \bigg) \bigg]$$
 
When $\Delta > 0$, $\pi_2^{(n)} \Pto \pi_{\max}$ and when $\Delta < 0$, $\pi_2^{(n)} \Pto \pi_{\min}$.
We now consider the $\Delta = 0$ case.
$$\pi_2^{(n)} = \pi_{\min} \vee \bigg[ \pi_{\max} \wedge \Phi \bigg( \frac{1}{ \sqrt{2} }
\bigg[ \frac{ \sum_{i=1}^n A_{1,i} \epsilon_{1,i} }{ \sqrt{ N_{1,1} } } 
- \frac{ \sum_{i=1}^n (1-A_{1,i}) \epsilon_{1,i} }{ \sqrt{ N_{1,0} } } \bigg] + o(1) \bigg) \bigg]$$
 $$= \pi_{\min} \vee \bigg[ \pi_{\max} \wedge \Phi \bigg( \frac{1}{ \sqrt{2} }
\bigg[ \frac{ \sum_{i=1}^n A_{1,i} \epsilon_{1,i} }{ \sqrt{ N_{1,1} } } 
- \frac{ \sum_{i=1}^n (1-A_{1,i}) \epsilon_{1,i} }{ \sqrt{ N_{1,0} } } \bigg] \bigg) \bigg] + o(1)$$

By Slutsky's Theorem, for $Z_1, Z_2, Z_3, Z_4 \iidsim \N(0, 1)$,
\small
$$[1, -1, 1, -1] \begin{bmatrix}
	\vspace{1mm}
	\sqrt{ \frac{ 1/2 + 1 - \pi_2 }{ 4 ( 1/2 + \pi_2 ) } } \frac{ \sum_{i=1}^n A_{1,i} \epsilon_{1,i} }{ \sqrt{ N_{1,1} } } \\
	\vspace{1mm}
	\sqrt{ \frac{ 1/2 + \pi_2 }{ 4 ( 1/2 + 1 - \pi_2 ) } } \frac{ \sum_{i=1}^n (1-A_{1,i}) \epsilon_{1,i} }{ \sqrt{ N_{1,0} } } \\
	\vspace{1mm}
	\sqrt{ \frac{ ( 1/2 + 1 - \pi_2 ) \pi_2 }{ 2 ( 1/2 + \pi_2) } } \frac{ \sum_{i=1}^n A_{2,i} \epsilon_{2,i} }{ \sqrt{N_{2,1} } } \\
	\sqrt{ \frac{ ( 1/2 + \pi_2 ) (1 - \pi_2 ) }{ 2 ( 1/2 + 1 - \pi_2 ) } } \frac{ \sum_{i=1}^n (1-A_{2,i}) \epsilon_{2,i} }{ \sqrt{ N_{2,0} } }
\end{bmatrix} + o_p(1)
\Dto 
[1, -1, 1, -1] \begin{bmatrix}
	\vspace{1mm}
	\sqrt{ \frac{ 1/2 + 1 - \pi_* }{ 4 ( 1/2 + \pi_* ) } } Z_1 \\
	\vspace{1mm}
	\sqrt{ \frac{ 1/2 + \pi_* }{ 4 ( 1/2 + 1 - \pi_* ) } } Z_2 \\
	\vspace{1mm}
	\sqrt{ \frac{ ( 1/2 + 1 - \pi_* ) \pi_* }{ 2 ( 1/2 + \pi_* ) } } Z_3  \\
	\sqrt{ \frac{ ( 1/2 + \pi_* ) (1 - \pi_* ) }{ 2 ( 1/2 + 1 - \pi_* ) } } Z_4 
\end{bmatrix}$$
\normalsize
$$= \sqrt{ \frac{ 1/2 + 1-\pi_* }{ 2 (1/2 + \pi_*) } }
\bigg( \sqrt{1/2} Z_1 + \sqrt{ \pi_* } Z_3 \bigg)
- \sqrt{ \frac{ 1/2 + \pi_* }{ 2 (1/2 + 1-\pi_* ) } }
\bigg( \sqrt{ 1/2 } Z_2 + \sqrt{ 1-\pi_* } Z_4 \bigg)$$
where
$\pi_* = \begin{cases}
	\pi_{\max} & \TN{ if } \Delta > 0 \\
	\pi_{\min} & \TN{ if } \Delta < 0 \\
	\pi_{\min} \vee ( \pi_{\max} \wedge \Phi [ \sqrt{ 1/2 } ( Z_1 - Z_2 ) ] ) & \TN{ if } \Delta = 0 ~~~ \square
\end{cases}$

%\medskip \medskip \medskip 
\clearpage
\subsection{$\epsilon$-Greedy}

\begin{proposition}[Non-concentration of the sampling probabilities under zero treatment effect for $\epsilon$-greedy]
\label{prop:epsilonnonconcentration}
Let $T=2$ and $\pi_1^{(n)} = \frac{1}{2}$ for all $n$.
We assume that $\{ \epsilon_{t,i} \}_{i=1}^n \iidsim \N(0, 1)$, and %$\beta_1 = b \in \real$, 
$$\pi_2^{(n)} 
= \begin{cases}
		1-\frac{\epsilon}{2} 		& \TN{ if } \frac{ \sum_{i=1}^n A_{1,i} R_{1,i} }{ N_{1,1} } > \frac{ \sum_{i=1}^n (1-A_{1,i}) R_{1,i} }{ N_{1,0} } \\
		\frac{\epsilon}{2} 			& \TN{ otherwise } \\
\end{cases}$$
%we perform batch $\epsilon$-greedy, so we assume that for the second batch that our action selection probabilities are determined as follows:
Thus, the sampling probability $\pi_2^{(n)}$ does not concentrate when $\beta_1 = \beta_0$.
\end{proposition}

\paragraph{Proof:}
We define $M_n := \II_{ \big( \frac{ \sum_{i=1}^n A_{1,i} R_{1,i} }{ N_{1,1} } > \frac{ \sum_{i=1}^n (1-A_{1,i}) R_{1,i} }{ N_{1,0} } \big) }
= \II_{ \big( (\beta_1 - \beta_0) + \frac{ \sum_{i=1}^n A_{1,i} \epsilon_{1,i} }{ N_{1,1} } > \frac{ \sum_{i=1}^n (1-A_{1,i}) \epsilon_{1,i} }{ N_{1,0} } \big) }$.
Note that when $M_n = 1$, $\pi_2^{(n)} = 1 - \frac{ \epsilon }{2}$ and when $M_n = 0$, $\pi_2^{(n)} = \frac{ \epsilon }{2}$. 

When the margin is zero, $M_n$ does not concentrate because for all $N_{1,1}, N_{1,0}$, since $\epsilon_{1,i} \iidsim \N(0,1)$,
$$\PP \bigg(  \frac{ \sum_{i=1}^n A_{1,i} \epsilon_{1,i} }{ N_{1,1} } > \frac{ \sum_{i=1}^n (1-A_{1,i}) \epsilon_{1,i} }{ N_{1,0} } \bigg) 
= \PP \bigg( \frac{1}{ \sqrt{ N_{1,1} } } Z_1 - \frac{1}{ \sqrt{ N_{1,0} }  } Z_2 > 0 \bigg) = \frac{1}{2}$$
for $Z_1, Z_2 \iidsim \N(0,1)$.
Thus, we have shown that $\pi_2^{(n)}$ does not concentrate when $\beta_1 - \beta_0 = 0$.  $\square$ \\

\begin{theorem}[Non-uniform convergence of the OLS estimator of the treatment effect for $\epsilon$-greedy]
%The OLS estimator of $\beta_1$ is as follows:
%$$\betahat_1^{\OLS} := \frac{ \sum_{t=1}^2 \sum_{i=1}^n A_{t,i} R_{t,i} }{ N_{1,1} + N_{2,1} }$$that we are under the null, $H_0 \colon \beta_1 = b$
Assuming the setup and conditions of Proposition \ref{prop:epsilonnonconcentration}, and that $\beta_1 = b$, we show that the normalized errors of the OLS estimator converges in distribution as follows: 
$$\sqrt{ N_{1,1} + N_{2,1} } \big( \betahat_1^{\OLS} - b \big)
\Dto Y$$
$$Y = \begin{cases} 
%%%%%%%%%%%%%%%%%%%%%%%%%%%%%%%%%%%%%%%%%%%%%%%%%%%%%%%
Z_1 & \TN{ if } \beta_1 - \beta_0 \neq 0 \\
 %%%%%%%%%%%%%%%%%%%%%%%%%%%%%%%%%%%%%%%%%%%%%%%%%%%%%%%
\sqrt{ \frac{ 1 }{ 3 - \epsilon } }  \big( Z_1 - \sqrt{ 2 - \epsilon } Z_3 \big) \II_{ (Z_1 > Z_2) }
+ \sqrt{ \frac{ 1 }{ 1 + \epsilon } } \big( Z_1 - \sqrt{ \epsilon } Z_3 \big) \II_{ (Z_1 < Z_2) } 
& \TN{ if } \beta_1 - \beta_0 = 0
\end{cases}$$
for $Z_1, Z_2, Z_3 \iidsim N(0, 1)$. 
Note the $\beta_1 - \beta_0 = 0$ case, $Y$ is non-normal.
\end{theorem}

\paragraph{Proof:}
The normalized errors of the OLS estimator for $\beta_1$ are
$$\sqrt{ N_{1,1} + N_{2,1} } \bigg( \frac{ \sum_{t=1}^2 \sum_{i=1}^n A_{t,i} R_{t,i} }{ N_{1,1} + N_{2,1} } - b \bigg)
= \frac{ \sum_{t=1}^2 \sum_{i=1}^n A_{t,i} \epsilon_{t,i} }{ \sqrt{ N_{1,1} + N_{2,1} } }$$
%= \frac{ \sum_{t=1}^2 \sum_{i=1}^n A_{t,i} \epsilon_{t,i} }{ \sqrt{ N_{1,1} + N_{2,1} } } + \sqrt{ N_{1,1} + N_{2,1} } ( \beta_1 - b )
$$= [1, 1] \begin{bmatrix}
	\frac{ \sum_{i=1}^n A_{1,i} \epsilon_{1,i} }{ \sqrt{ N_{1,1} + N_{2,1} } } \\
	\frac{ \sum_{i=1}^n A_{2,i} \epsilon_{2,i} }{ \sqrt{ N_{1,1} + N_{2,1} } }
\end{bmatrix}
= [1, 1] \begin{bmatrix}
	 \sqrt{ \frac{ N_{1,1} }{ N_{1,1} + N_{2,1} } } \frac{ \sum_{i=1}^n A_{1,i} \epsilon_{1,i} }{ \sqrt{ N_{1,1} } } \\
	 \sqrt{ \frac{ N_{2,1} }{ N_{1,1} + N_{2,1} } } \frac{ \sum_{i=1}^n A_{2,i} \epsilon_{2,i} }{ \sqrt{ N_{2,1} } }
\end{bmatrix}$$
%%%%%%%%%%%%%%%%%%%%%%%%%%%%%%%%%%%%%%%%%%%
By Slutsky's Theorem and Lemma \ref{lemma:ratio}, 
$\bigg( \sqrt{ \frac{ 1/2 }{ 1/2 + \pi_2^{(n)}} } \sqrt{ \frac{ N_{1,1} + N_{2,1} }{ N_{1,1} } },
	 \sqrt{ \frac{ \pi_2^{(n)} }{ 1/2 + \pi_2^{(n)}} } \sqrt{ \frac{ N_{1,1} + N_{2,1} }{ N_{2,1} } } 
\bigg) \Pto ( 1, 1 )$, so
$$= [1, 1] \begin{bmatrix}
	 \bigg( \sqrt{ \frac{ 1/2 }{ 1/2 + \pi_2^{(n)}} } \sqrt{ \frac{ N_{1,1} + N_{2,1} }{ N_{1,1} } } + o_p(1) \bigg) 
	 \sqrt{ \frac{ N_{1,1} }{ N_{1,1} + N_{2,1} } } \frac{ \sum_{i=1}^n A_{1,i} \epsilon_{1,i} }{ \sqrt{ N_{1,1} } } \\
	 \bigg( \sqrt{ \frac{ \pi_2^{(n)} }{ 1/2 + \pi_2^{(n)}} } \sqrt{ \frac{ N_{1,1} + N_{2,1} }{ N_{2,1} } } + o_p(1) \bigg)
	 \sqrt{ \frac{ N_{2,1} }{ N_{1,1} + N_{2,1} } } \frac{ \sum_{i=1}^n A_{2,i} \epsilon_{2,i} }{ \sqrt{ N_{2,1} } }
\end{bmatrix}$$
%$$= [1, 1] \begin{bmatrix}
%	 \sqrt{ \frac{ 1/2 }{ 1/2 + \pi_2^{(n)}} } \frac{ \sum_{i=1}^n A_{1,i} \epsilon_{1,i} }{ \sqrt{ N_{1,1} } } + o_p(1) \sqrt{ \frac{ N_{1,1} }{ N_{1,1} + N_{2,1} } } \frac{ \sum_{i=1}^n A_{1,i} \epsilon_{1,i} }{ \sqrt{ N_{1,1} } } \\
%	 \sqrt{ \frac{ \pi_2^{(n)} }{ 1/2 + \pi_2^{(n)}} }  \frac{ \sum_{i=1}^n A_{2,i} \epsilon_{2,i} }{ \sqrt{ N_{2,1} } } + 
%	 o_p(1) \sqrt{ \frac{ N_{2,1} }{ N_{1,1} + N_{2,1} } } \frac{ \sum_{i=1}^n A_{2,i} \epsilon_{2,i} }{ \sqrt{ N_{2,1} } }
%\end{bmatrix}$$
$$= [1, 1] \begin{bmatrix}
	 \sqrt{ \frac{ 1/2 }{ 1/2 + \pi_2^{(n)}} } \frac{ \sum_{i=1}^n A_{1,i} \epsilon_{1,i} }{ \sqrt{ N_{1,1} } } + o_p(1) \\
	 \sqrt{ \frac{ \pi_2^{(n)} }{ 1/2 + \pi_2^{(n)}} }  \frac{ \sum_{i=1}^n A_{2,i} \epsilon_{2,i} }{ \sqrt{ N_{2,1} } } + o_p(1) 
\end{bmatrix}$$
The last equality holds because by Theorem \ref{thm:bols}, 
$\big( \frac{ \sum_{i=1}^n A_{1,i} \epsilon_{1,i} }{ \sqrt{ N_{1,1} } }, 
\frac{ \sum_{i=1}^n A_{2,i} \epsilon_{2,i} }{ \sqrt{N_{2,1} } }
\big) \Dto \N( \bs{0}, \under{\bo{I}}_2 )$. \\

Let's define $M_n := \II_{ \big( \frac{ \sum_{i=1}^n A_{1,i} R_{1,i} }{ N_{1,1} } > \frac{ \sum_{i=1}^n (1-A_{1,i}) R_{1,i} }{ N_{1,0} } \big) }
= \II_{ \big( (\beta_1 - \beta_0) + \frac{ \sum_{i=1}^n A_{1,i} \epsilon_{1,i} }{ N_{1,1} } > \frac{ \sum_{i=1}^n (1-A_{1,i}) \epsilon_{1,i} }{ N_{1,0} } \big) }$.
Note that when $M_n = 1$, $\pi_2^{(n)} = 1 - \frac{ \epsilon }{2}$ and when $M_n = 0$, $\pi_2^{(n)} = \frac{ \epsilon }{2}$. 
$$M_n = \II_{ \big( (\beta_1 - \beta_0) + \frac{ \sum_{i=1}^n A_{1,i} \epsilon_{1,i} }{ N_{1,1} } > \frac{ \sum_{i=1}^n (1-A_{1,i}) \epsilon_{1,i} }{ N_{1,0} } \big) }
= \II_{ \big( \sqrt{ N_{1,0} } (\beta_1 - \beta_0) + \sqrt{ \frac{ N_{1,0} }{ N_{1,1} } } \frac{ \sum_{i=1}^n A_{1,i} \epsilon_{1,i} }{ \sqrt{ N_{1,1} } } > \frac{ \sum_{i=1}^n (1-A_{1,i}) \epsilon_{1,i} }{ \sqrt{ N_{1,0} } } \big) }$$
$$= \II_{ \big( \sqrt{ N_{1,0} } (\beta_1 - \beta_0) + [1 + o_p(1)] \frac{ \sum_{i=1}^n A_{1,i} \epsilon_{1,i} }{ \sqrt{ N_{1,1} } } > \frac{ \sum_{i=1}^n (1-A_{1,i}) \epsilon_{1,i} }{ \sqrt{ N_{1,0} } } \big) }$$
where the last equality holds because $\sqrt{ \frac{ N_{1,0} }{ N_{1,1} } }  \Pto 1$ by Lemma \ref{lemma:ratio}, Slutsky's Theorem, and continuous mapping theorem.
Thus, by Proposition \ref{prop:epsilonnonconcentration}, 
$$M^{(n)} \Pto \begin{cases} 
%%%%%%%%%%%%%%%%%%%%%%%%%%%%%%%%%%%%%%%%%%%%%%%%%%%%%%%
1 & \TN{ if } \beta_1 - \beta_0 > 0 \\
%%%%%%%%%%%%%%%%%%%%%%%%%%%%%%%%%%%%%%%%%%%%%%%%%%%%%%%
%%%%%%%%%%%%%%%%%%%%%%%%%%%%%%%%%%%%%%%%%%%%%%%%%%%%%%%
0 & \TN{ if } \beta_1 - \beta_0 < 0 \\
 %%%%%%%%%%%%%%%%%%%%%%%%%%%%%%%%%%%%%%%%%%%%%%%%%%%%%%%
\TN{does not concentrate }
& \TN{ if } \beta_1 - \beta_0 = 0
\end{cases}$$

%We will now show the asymptotic non-normality result.
Note that
$$\begin{bmatrix}
	 \sqrt{ \frac{ \frac{1}{2} }{ \frac{1}{2} + \pi_2^{(n)}} } \frac{ \sum_{i=1}^n A_{1,i} \epsilon_{1,i} }{ \sqrt{ N_{1,1} } } + o_p(1) \\
	 \sqrt{ \frac{ \pi_2^{(n)} }{ \frac{1}{2} + \pi_2^{(n)}} } \frac{ \sum_{i=1}^n A_{2,i} \epsilon_{2,i} }{ \sqrt{ N_{2,1} } } + o_p(1)
\end{bmatrix}$$
$$= \begin{bmatrix}
	 \sqrt{ \frac{ \frac{1}{2} }{ \frac{1}{2} + 1 - \frac{ \epsilon }{2} } } \frac{ \sum_{i=1}^n A_{1,i} \epsilon_{1,i} }{ \sqrt{ N_{1,1} } } + o_p(1) \\
	 \sqrt{ \frac{ 1 - \epsilon/2 }{ \frac{1}{2} + 1 - \frac{ \epsilon }{2} } } \frac{ \sum_{i=1}^n A_{2,i} \epsilon_{2,i} }{ \sqrt{ N_{2,1} } } + o_p(1)
\end{bmatrix} M_n
+ \begin{bmatrix}
	 \sqrt{ \frac{ \frac{1}{2} }{ \frac{1}{2} + \frac{ \epsilon }{2} } } \frac{ \sum_{i=1}^n A_{1,i} \epsilon_{1,i} }{ \sqrt{ N_{1,1} } } + o_p(1) \\
	 \sqrt{ \frac{ \frac{ \epsilon }{2} }{ \frac{1}{2} + \frac{ \epsilon }{2} } } \frac{ \sum_{i=1}^n A_{2,i} \epsilon_{2,i} }{ \sqrt{ N_{2,1} } } + o_p(1)
\end{bmatrix} ( 1 - M_n ) $$

Also note that by Theorem \ref{thm:bols}, 
$\bigg( \frac{ \sum_{i=1}^n A_{1,i} \epsilon_{1,i} }{ \sqrt{ N_{1,1} } }, 
\frac{ \sum_{i=1}^n (1-A_{1,i}) \epsilon_{1,i} }{ \sqrt{ N_{1,0} } }, 
\frac{ \sum_{i=1}^n A_{2,i} \epsilon_{2,i} }{ \sqrt{N_{2,1} } },
\frac{ \sum_{i=1}^n (1-A_{2,i}) \epsilon_{2,i} }{ \sqrt{N_{2,1} } }
\bigg) \Dto \N( \bs{0}, \under{\bo{I}}_4 )$.

When $\beta_1 > \beta_0$, $M_n \Pto 1$ and when $\beta_1 < \beta_0$, $M_n \Pto 0$; in both these cases the normalized errors are asymptotically normal. 
We now focus on the case that $\beta_1 = \beta_0$.
By continuous mapping theorem and Slutsky's theorem for $Z_1, Z_2, Z_3, Z_4 \iidsim \N(0, 1)$,
$$= [1, 1] \begin{bmatrix}
	 \sqrt{ \frac{ \frac{1}{2} }{ \frac{1}{2} + 1 - \frac{ \epsilon }{2} } } \frac{ \sum_{i=1}^n A_{1,i} \epsilon_{1,i} }{ \sqrt{ N_{1,1} } } + o_p(1) \\
	 \sqrt{ \frac{ 1 - \epsilon/2 }{ \frac{1}{2} + 1 - \frac{ \epsilon }{2} } } \frac{ \sum_{i=1}^n A_{2,i} \epsilon_{2,i} }{ \sqrt{ N_{2,1} } } + o_p(1)
\end{bmatrix} \II_{ \big( [1 + o(1)] \frac{ \sum_{i=1}^n A_{1,i} \epsilon_{1,i} }{ \sqrt{ N_{1,1} } } > \frac{ \sum_{i=1}^n (1-A_{1,i}) \epsilon_{1,i} }{ \sqrt{ N_{1,0} } } \big) }$$
$$+ [1, 1] \begin{bmatrix}
	 \sqrt{ \frac{ \frac{1}{2} }{ \frac{1}{2} + \frac{ \epsilon }{2} } } \frac{ \sum_{i=1}^n A_{1,i} \epsilon_{1,i} }{ \sqrt{ N_{1,1} } } + o_p(1) \\
	 \sqrt{ \frac{ \frac{ \epsilon }{2} }{ \frac{1}{2} + \frac{ \epsilon }{2} } } \frac{ \sum_{i=1}^n A_{2,i} \epsilon_{2,i} }{ \sqrt{ N_{2,1} } } + o_p(1)
\end{bmatrix} \bigg( 1 - \II_{ \big( [1 + o(1)] \frac{ \sum_{i=1}^n A_{1,i} \epsilon_{1,i} }{ \sqrt{ N_{1,1} } } > \frac{ \sum_{i=1}^n (1-A_{1,i}) \epsilon_{1,i} }{ \sqrt{ N_{1,0} } } \big) } \bigg) $$
$$\Dto [1, 1] \begin{bmatrix}
	 \sqrt{ \frac{ 1/2 }{ 1/2 + 1 - \epsilon/2 } } Z_1 \\
	 \sqrt{ \frac{ 1 - \epsilon/2 }{ 1/2 + 1 - \epsilon/2 } } Z_3
\end{bmatrix} \II_{ (Z_1 > Z_2) }
+ [1, 1] \begin{bmatrix}
	 \sqrt{ \frac{ 1/2 }{ 1/2 + \epsilon/2 } } Z_1 \\
	 \sqrt{ \frac{ \epsilon/2 }{ 1/2 + \epsilon/2 } } Z_3
\end{bmatrix} \II_{ (Z_1 < Z_2) }$$
$$= \bigg( \sqrt{ \frac{ 1 }{ 3 - \epsilon } } Z_1 + \sqrt{ \frac{ 2 - \epsilon }{ 3 - \epsilon } } Z_3 \bigg) \II_{ (Z_1 > Z_2) }
+ \bigg( \sqrt{ \frac{ 1 }{ 1 + \epsilon } } Z_1 + \sqrt{ \frac{ \epsilon }{ 1 + \epsilon } } Z_3 \bigg) \II_{ (Z_1 < Z_2) }$$ 

Thus,
%\small
$$\frac{ \sum_{t=1}^2 \sum_{i=1}^n A_{t,i} \epsilon_{t,i} }{ \sqrt{ N_{1,1} + N_{2,1} } } \Dto Y$$
$$Y := \begin{cases} 
%%%%%%%%%%%%%%%%%%%%%%%%%%%%%%%%%%%%%%%%%%%%%%%%%%%%%%%
\sqrt{ \frac{ 1 }{ 3 - \epsilon } }  \big( Z_1 - \sqrt{ 2 - \epsilon } Z_3 \big)
& \TN{ if } \beta_1 - \beta_0 > 0 \\
%%%%%%%%%%%%%%%%%%%%%%%%%%%%%%%%%%%%%%%%%%%%%%%%%%%%%%%
%%%%%%%%%%%%%%%%%%%%%%%%%%%%%%%%%%%%%%%%%%%%%%%%%%%%%%%
\sqrt{ \frac{ 1 }{ 1 + \epsilon } } \big( Z_1 - \sqrt{ \epsilon } Z_3 \big)
 & \TN{ if } \beta_1 - \beta_0 < 0 \\
 %%%%%%%%%%%%%%%%%%%%%%%%%%%%%%%%%%%%%%%%%%%%%%%%%%%%%%%
\sqrt{ \frac{ 1 }{ 3 - \epsilon } }  \big( Z_1 - \sqrt{ 2 - \epsilon } Z_3 \big) \II_{( Z_1 > Z_2 )}
+ \sqrt{ \frac{ 1 }{ 1 + \epsilon } } \big( Z_1 - \sqrt{ \epsilon } Z_3 \big) \II_{ (Z_1 < Z_2 )}
& \TN{ if } \beta_1 - \beta_0 = 0 ~~ \square
\end{cases}$$

\clearpage
\subsection{UCB}

\begin{theorem}[Asymptotic non-Normality under zero treatment effect for clipped UCB]
Let $T=2$ and $\pi_1^{(n)} = \frac{1}{2}$ for all $n$.
We assume that $\{ \epsilon_{t,i} \}_{i=1}^n \iidsim \N(0, 1)$, and %$\beta_1 = b \in \real$, 
\begin{equation*}
\pi_2^{(n)} 
= \begin{cases}
		\pi_{\max} 		& \TN{ if } U_1 > U_0 \\
		1 -  \pi_{\max}		& \TN{ otherwise } \\
\end{cases}
\end{equation*}
where we define the upper confidence bounds $U$ for any confidence level $\delta$ with $0 < \delta < 1$ as follows:
\begin{equation*}
	U_1 = \begin{cases} 
			\infty & \TN{ if } N_{1,1} = 0 \\
			\frac{ \sum_{i=1}^n A_{1,i} R_{1,i} }{ N_{1,1} } + \sqrt{ \frac{2 \log 1/\delta}{ N_{1,1} } } & \TN{ otherwise } 
		\end{cases}
\end{equation*}
\begin{equation*}
	U_0 = \begin{cases} 
			\infty & \TN{ if } N_{1,0} = 0 \\
			\frac{ \sum_{i=1}^n (1-A_{1,i}) R_{1,i} }{ N_{1,1} } + \sqrt{ \frac{2 \log 1/\delta}{ N_{1,0} } } & \TN{ otherwise } 
		\end{cases}
\end{equation*}
Assuming above conditions, and that $\beta_1 = b$, we show that the normalized errors of the OLS estimator converges in distribution as follows: 
\begin{equation*}
	\sqrt{ N_{1,1} + N_{2,1} } \big( \betahat_1^{\OLS} - b \big) \Dto Y
\end{equation*}
\footnotesize
\begin{equation*}
	Y = \begin{cases} 
	%%%%%%%%%%%%%%%%%%%%%%%%%%%%%%%%%%%%%%%%%%%%%%%%%%%%%%%
	Z_1 & \TN{ if } \Delta = 0 \\
 	%%%%%%%%%%%%%%%%%%%%%%%%%%%%%%%%%%%%%%%%%%%%%%%%%%%%%%%
	\big( \sqrt{ \frac{ \frac{1}{2} }{ \frac{1}{2} + \pi_{\max} } } Z_1 + \sqrt{ \frac{ \pi_{\max} }{ \frac{1}{2} + \pi_{\max} } } Z_3 \big) \II_{ (Z_1 > Z_2) }
+ \big(  \sqrt{ \frac{ \frac{1}{2} }{ \frac{3}{2} - \pi_{\max} } } Z_1 + \sqrt{ \frac{ 1 - \pi_{\max} }{ \frac{3}{2} - \pi_{\max} } } Z_3 \big) \II_{ (Z_1 < Z_2) }
	& \TN{ if } \Delta = 0
	\end{cases}
\end{equation*}
\normalsize
for $Z_1, Z_2, Z_3 \iidsim N(0, 1)$. 
Note the $\Delta := \beta_1 - \beta_0 = 0$ case, $Y$ is non-normal.
\end{theorem}

\paragraph{Proof:}
The proof is very similar to that of asymptotic non-normality result for $\epsilon$-Greedy. By the same arguments made as in the $\epsilon$-Greedy case, we have that
\begin{equation*}
	\sqrt{ N_{1,1} + N_{2,1} } \bigg( \frac{ \sum_{t=1}^2 \sum_{i=1}^n A_{t,i} R_{t,i} }{ N_{1,1} + N_{2,1} } - b \bigg)
	= [1, 1] \begin{bmatrix}
	 	\sqrt{ \frac{ 1/2 }{ 1/2 + \pi_2^{(n)}} } \frac{ \sum_{i=1}^n A_{1,i} \epsilon_{1,i} }{ \sqrt{ N_{1,1} } } + o_p(1) \\
	 	\sqrt{ \frac{ \pi_2^{(n)} }{ 1/2 + \pi_2^{(n)}} }  \frac{ \sum_{i=1}^n A_{2,i} \epsilon_{2,i} }{ \sqrt{ N_{2,1} } } + o_p(1) 
	\end{bmatrix}
\end{equation*}

Assuming $n \geq 1$, we then define 
\begin{equation*}
M_n := \II_{ ( U_1 > U_0 ) }
\end{equation*}
\begin{equation*}
= \II_{ (N_{1,1} > 0, N_{1,0} > 0) } \II_{ \big( \frac{ \sum_{i=1}^n A_{1,i} R_{1,i} }{ N_{1,1} } + \sqrt{ \frac{2 \log 1/\delta}{ N_{1,1} } } > \frac{ \sum_{i=1}^n (1-A_{1,i}) R_{1,i} }{ N_{1,1} } + \sqrt{ \frac{2 \log 1/\delta}{ N_{1,0} } } \big) } 
+ \II_{ (N_{1,1} = 0, N_{1,0} > 0) }
\end{equation*}
\begin{equation*}
= \II_{ (N_{1,1} > 0, N_{1,0} > 0) } \II_{ \big( (\beta_1 - \beta_0) + \frac{ \sum_{i=1}^n A_{1,i} \epsilon_{1,i} }{ N_{1,1} } 
+ \sqrt{ \frac{2 \log 1/\delta}{ N_{1,1} } } > \frac{ \sum_{i=1}^n (1-A_{1,i}) \epsilon_{1,i} }{ N_{1,1} } + \sqrt{ \frac{2 \log 1/\delta}{ N_{1,0} } } \big) } 
+ \II_{ (N_{1,1} = 0, N_{1,0} > 0) }
\end{equation*}
\begin{multline*}
= \II_{ (N_{1,1} > 0, N_{1,0} > 0) } \II_{ \big( \sqrt{ N_{1,0} } (\beta_1 - \beta_0) + \sqrt{ \frac{ N_{1,0} }{ N_{1,1} } } \big[ \frac{ \sum_{i=1}^n A_{1,i} \epsilon_{1,i} }{ \sqrt{ N_{1,1} } } + \sqrt{ 2 \log 1/\delta } \big]
> \frac{ \sum_{i=1}^n (1-A_{1,i}) \epsilon_{1,i} }{ \sqrt{ N_{1,1} } } + \sqrt{ 2 \log 1/\delta } \big) }  \\
+ \II_{ (N_{1,1} = 0, N_{1,0} > 0) }
\end{multline*}
Note that $\frac{ N_{1,0} }{ N_{1,1} }  \Pto 1$ by Lemma \ref{lemma:ratio}. Thus by Slutsky's Theorem and continuous mapping theorem,
\begin{equation}
\label{eqn:ucbnonconcentration}
= \II_{ \big( \sqrt{ N_{1,0} } (\beta_1 - \beta_0) + [1 + o_p(1)] \frac{ \sum_{i=1}^n A_{1,i} \epsilon_{1,i} }{ \sqrt{ N_{1,1} } } + o_p(1)
> \frac{ \sum_{i=1}^n (1-A_{1,i}) \epsilon_{1,i} }{ \sqrt{ N_{1,1} } } \big) }  
+ o_p(1)
\end{equation}
Note that 
\begin{equation*}
	\begin{bmatrix}
	 \sqrt{ \frac{ \frac{1}{2} }{ \frac{1}{2} + \pi_2^{(n)}} } \frac{ \sum_{i=1}^n A_{1,i} \epsilon_{1,i} }{ \sqrt{ N_{1,1} } } + o_p(1) \\
	 \sqrt{ \frac{ \pi_2^{(n)} }{ \frac{1}{2} + \pi_2^{(n)}} } \frac{ \sum_{i=1}^n A_{2,i} \epsilon_{2,i} }{ \sqrt{ N_{2,1} } } + o_p(1)
\end{bmatrix}
\end{equation*}
\begin{equation*}
= \begin{bmatrix}
	 \sqrt{ \frac{ \frac{1}{2} }{ \frac{1}{2} + \pi_{\max} } } \frac{ \sum_{i=1}^n A_{1,i} \epsilon_{1,i} }{ \sqrt{ N_{1,1} } } + o_p(1) \\
	 \sqrt{ \frac{ \pi_{\max} }{ \frac{1}{2} + \pi_{\max} } } \frac{ \sum_{i=1}^n A_{2,i} \epsilon_{2,i} }{ \sqrt{ N_{2,1} } } + o_p(1)
\end{bmatrix} M_n
+ \begin{bmatrix}
	 \sqrt{ \frac{ \frac{1}{2} }{ \frac{1}{2} + 1- \pi_{\max} } } \frac{ \sum_{i=1}^n A_{1,i} \epsilon_{1,i} }{ \sqrt{ N_{1,1} } } + o_p(1) \\
	 \sqrt{ \frac{ 1 - \pi_{\max} }{ \frac{1}{2} + 1 - \pi_{\max} } } \frac{ \sum_{i=1}^n A_{2,i} \epsilon_{2,i} }{ \sqrt{ N_{2,1} } } + o_p(1)
\end{bmatrix} ( 1 - M_n )
\end{equation*}
Let $(Z_1^{(n)}, Z_2^{(n)}, Z_3^{(n)}, Z_4^{(n)}) := \bigg( \frac{ \sum_{i=1}^n A_{1,i} \epsilon_{1,i} }{ \sqrt{ N_{1,1} } }, 
\frac{ \sum_{i=1}^n (1-A_{1,i}) \epsilon_{1,i} }{ \sqrt{ N_{1,0} } }, 
\frac{ \sum_{i=1}^n A_{2,i} \epsilon_{2,i} }{ \sqrt{N_{2,1} } },
\frac{ \sum_{i=1}^n (1-A_{2,i}) \epsilon_{2,i} }{ \sqrt{N_{2,1} } }
\bigg)$.
Note that by Theorem \ref{thm:bols}, 
$(Z_1^{(n)}, Z_2^{(n)}, Z_3^{(n)}, Z_4^{(n)}) \Dto \N( \bs{0}, \under{\bo{I}}_4 )$. 

When $\beta_1 > \beta_0$, $M_n \Pto 1$ and when $\beta_1 < \beta_0$, $M_n \Pto 0$; in both these cases the normalized errors are asymptotically normal. 
We now focus on the case that $\beta_1 = \beta_0$.
By continuous mapping theorem and Slutsky's theorem, %for $Z_1, Z_2, Z_3, Z_4 \iidsim \N(0, 1)$,
\begin{multline*}
	= [1, 1] \begin{bmatrix}
	 	\sqrt{ \frac{ \frac{1}{2} }{ \frac{1}{2} + \pi_{\max} } } Z_1^{(n)} + o_p(1) \\
		 \sqrt{ \frac{ \pi_{\max} }{ \frac{1}{2} + \pi_{\max} } } Z_3^{(n)} + o_p(1)
	\end{bmatrix} 
	\bigg[ \II_{ \big( [1 + o_p(1)] Z_1^{(n)} + o_p(1)> Z_2^{(n)} \big) } + o_p(1) \bigg] \\
+ [1, 1] \begin{bmatrix}
	 \sqrt{ \frac{ \frac{1}{2} }{ \frac{1}{2} + 1- \pi_{\max} } } Z_1^{(n)} + o_p(1) \\
	 \sqrt{ \frac{ 1 - \pi_{\max} }{ \frac{1}{2} + 1 - \pi_{\max} } } Z_3^{(n)} + o_p(1)
\end{bmatrix} \bigg[ 1 - \II_{ \big( [1 + o_p(1)] Z_1^{(n)} + o_p(1)> Z_2^{(n)} \big) } + o_p(1) \bigg] \bigg)
\end{multline*}
\begin{equation*}
	\Dto [1, 1] \begin{bmatrix}
	 	\sqrt{ \frac{ \frac{1}{2} }{ \frac{1}{2} + \pi_{\max} } } Z_1 \\
		 \sqrt{ \frac{ \pi_{\max} }{ \frac{1}{2} + \pi_{\max} } } Z_3
	\end{bmatrix}  \II_{ (Z_1 > Z_2) }
	+ [1, 1] \begin{bmatrix}
	 \sqrt{ \frac{ \frac{1}{2} }{ \frac{1}{2} + 1- \pi_{\max} } } Z_1 \\
	 \sqrt{ \frac{ 1 - \pi_{\max} }{ \frac{1}{2} + 1 - \pi_{\max} } } Z_3
\end{bmatrix} \II_{ (Z_1 < Z_2) }
\end{equation*}
\small
\begin{equation*}
= \bigg( \sqrt{ \frac{ \frac{1}{2} }{ \frac{1}{2} + \pi_{\max} } } Z_1 + \sqrt{ \frac{ \pi_{\max} }{ \frac{1}{2} + \pi_{\max} } } Z_3 \bigg) \II_{ (Z_1 > Z_2) }
+ \bigg(  \sqrt{ \frac{ \frac{1}{2} }{ \frac{3}{2} - \pi_{\max} } } Z_1 + \sqrt{ \frac{ 1 - \pi_{\max} }{ \frac{3}{2} - \pi_{\max} } } Z_3 \bigg) \II_{ (Z_1 < Z_2) }. 
\qed
\end{equation*}
\normalsize

Note that \eqref{eqn:ucbnonconcentration} implies that if $\beta_1 = \beta_0$, that $\pi_2^{(n)}$ will not concentrate.

%% file: appendix/BOLS.tex
\section{Asymptotic Normality of the Batched OLS Estimator: Multi-Arm Bandits}
\label{appendix:BOLS}

\paragraph{Theorem \ref{thm:bols}} (Asymptotic normality of Batched OLS estimator for multi-arm bandits)
\begin{itshape}
Assuming Conditions \ref{cond:weakmoments} (weak moments) and \ref{cond:condiid} (conditionally i.i.d. actions), and a clipping rate of $f(n) = \omega(\frac{1}{n})$ (Definition \ref{def:clipping}),
\begin{equation*}
\begin{bmatrix} 
	\begin{bmatrix} N_{1,0} & 0 \\ 0 & N_{1,1} \end{bmatrix}^{1/2} ( \bs{\hat{\beta}}_1^{\BOLS} - \bs{\beta}_1 ) \\
	\begin{bmatrix} N_{2,0} & 0 \\ 0 & N_{2,1} \end{bmatrix}^{1/2} ( \bs{\hat{\beta}}_2^{\BOLS} - \bs{\beta}_2 )  \\
	\vdots \\
	\begin{bmatrix} N_{T,0} & 0 \\ 0 & N_{T,1} \end{bmatrix}^{1/2} ( \bs{\hat{\beta}}_T^{\BOLS} - \bs{\beta}_T )  \\
\end{bmatrix}
\Dto \N(0, \sigma^2 \under{\bo{I}}_{2T} )
\end{equation*}
where $\bs{\beta}_t = ( \beta_{t,0}, \beta_{t,1} )$, $N_{t,1} = \sum_{i=1}^n A_{t,i}$, and $N_{t,0} = \sum_{i=1}^n (1-A_{t,i})$.
Note in the body of this paper, we state Theorem \ref{thm:bols} with conditions that are are sufficient for the weaker conditions we use here.
\end{itshape}

\smallskip
\begin{lemma}
\label{lemma:ratio}
Assuming the conditions of Theorem \ref{thm:bols}, for any batch $t \in [1 \colon T]$,
\begin{equation*}
\frac{ N_{t,1} }{ n \pi_t^{(n)} }
= \frac{ \sum_{i=1}^n A_{t,i} }{ n \pi_t^{(n)} } \Pto 1
~~~~ \TN{ and } ~~~~
\frac{ N_{t,0} }{ n (1 - \pi_t^{(n)} ) }
= \frac{ \sum_{i=1}^n (1 - A_{t,i} ) }{ n (1-\pi_t^{(n)}) } \Pto 1
\end{equation*}
\end{lemma}

\paragraph{Proof of Lemma \ref{lemma:ratio}:}
To prove that $\frac{ N_{t,1} }{ n \pi_t^{(n)} }\Pto 1$, it is equivalent to show that
$\frac{1}{ n \pi_t^{(n)} } \sum_{i=1}^{n} ( A_{t,i} - \pi_t^{(n)} ) \Pto 0$. Let $\epsilon > 0$.
\small
\begin{equation*}
\PP \bigg( \bigg| \frac{1}{ n \pi_t^{(n)} } \sum_{i=1}^{n} ( A_{t,i} - \pi_t^{(n)} ) 
 \bigg| > \epsilon \bigg)
 \end{equation*}
\begin{equation*}
 = \PP \bigg( \bigg| \frac{1}{ n \pi_t^{(n)} } \sum_{i=1}^{n} ( A_{t,i} - \pi_t^{(n)} ) \bigg| \bigg[ \II_{( \pi_t^{(n)} \in [f(n), 1-f(n)] )} + \II_{( \pi_t^{(n)} \not\in [f(n), 1-f(n)] )} \bigg] > \epsilon \bigg)
\end{equation*}
\begin{multline*}
 \leq \PP \bigg( \bigg| \frac{1}{ n \pi_t^{(n)} } \sum_{i=1}^{n} ( A_{t,i} - \pi_t^{(n)} ) \bigg| \II_{( \pi_t^{(n)} \in [f(n), 1-f(n)] )} > \frac{\epsilon}{2} \bigg) \\
 + \PP \bigg( \bigg| \frac{1}{ n \pi_t^{(n)} } \sum_{i=1}^{n} ( A_{t,i} - \pi_t^{(n)} ) \bigg| \II_{( \pi_t^{(n)} \not\in [f(n), 1-f(n)] )} > \frac{\epsilon}{2} \bigg)
\end{multline*}
\normalsize
Since by our clipping assumption, $\II_{( \pi_t^{(n)} \in [f(n), 1-f(n)] )} \Pto 1$, the second probability in the summation above converges to $0$ as $n \to \infty$.
We will now show that the first probability in the summation above also goes to zero. Note that 
$\E \big[ \frac{1}{ n \pi_t^{(n)} } \sum_{i=1}^{n} ( A_{t,i} - \pi_t^{(n)} ) \big] 
= \E \big[ \frac{1}{ n \pi_t^{(n)} } \sum_{i=1}^{n} ( \E[ A_{t,i} | H_{t-1}^{(n)} ] - \pi_t^{(n)} ) \big] 
= 0$.
So by Chebychev inequality, for any $\epsilon > 0$,
\begin{equation*}
\PP \bigg( \bigg| \frac{1}{ n \pi_t^{(n)} } \sum_{i=1}^{n} ( A_{t,i} - \pi_t^{(n)} ) \bigg| \II_{( \pi_t^{(n)} \in [f(n), 1-f(n)] )} > \epsilon \bigg)
\end{equation*}
\begin{equation*}
 \leq \frac{1}{\epsilon^2 n^2} \E \bigg[ \frac{1}{ ( \pi_t^{(n)} )^2} \bigg(  \sum_{i=1}^{n} ( A_{t,i} - \pi_t^{(n)} ) \bigg)^2 \II_{( \pi_t^{(n)} \in [f(n), 1-f(n)] )} \bigg]
\end{equation*}
\begin{equation*}
\leq \frac{1}{\epsilon^2 n^2 } \sum_{i=1}^{n} \sum_{j=1}^{n} \E \bigg[ \frac{1}{ ( \pi_t^{(n)} )^2} ( A_{t,i} - \pi_t^{(n)} ) ( A_{t,j} - \pi_t^{(n)} ) \II_{( \pi_t^{(n)} \in [f(n), 1-f(n)] )} \bigg]
\end{equation*}
  %%%%%%%%%%%%%%%%%%%%%%%%%%%%%%%%%%%
\begin{equation*}
= \frac{1}{\epsilon^2 n^2 } 
  \sum_{i=1}^{n} \sum_{j=1}^{n} \E \bigg[ \frac{1}{ ( \pi_t^{(n)} )^2} \II_{( \pi_t^{(n)} \in [f(n), 1-f(n)] )}
  \E \big[ A_{t,i} A_{t,j} - \pi_t^{(n)} (A_{t,i} + A_{t,j}) + ( \pi_t^{(n)} )^2 \big| H_{t-1}^{(n)} \big] \bigg]
\end{equation*}
\begin{equation}
\label{eqn:1}
= \frac{1}{\epsilon^2 n^2 } 
  \sum_{i=1}^{n} \sum_{j=1}^{n} \E \bigg[ \frac{1}{ ( \pi_t^{(n)} )^2} \II_{( \pi_t^{(n)} \in [f(n), 1-f(n)] )}
 \bigg( \E \big[ A_{t,i} A_{t,j} \big| H_{t-1}^{(n)} \big] - ( \pi_t^{(n)} )^2 \bigg) \bigg]
 \end{equation}
  %%%%%%%%%%%%%%%%%%%%%%%%%%%%%%%%%%%
Note that if $i \neq j$, since $A_{t,i} \iidsim \TN{Bernoulli}( \pi_t^{(n)} )$,
$\E [ A_{t,i} A_{t,j} | H_{t-1}^{(n)} ] = \E [ A_{t,i} | H_{t-1}^{(n)} ] E[ A_{t,j} | H_{t-1}^{(n)} ] = ( \pi_t^{(n)} )^2$, so \eqref{eqn:1} above equals the following
\begin{equation*}
= \frac{1}{\epsilon^2 n^2 } 
  \sum_{i=1}^{n} \E \bigg[ \frac{1}{ ( \pi_t^{(n)} )^2} \II_{( \pi_t^{(n)} \in [f(n), 1-f(n)] )}
 \bigg( \E \big[ A_{t,i} \big| H_{t-1}^{(n)} \big] - ( \pi_t^{(n)} )^2 \bigg) \bigg]
\end{equation*}
\small
\begin{equation*}
 = \frac{1}{\epsilon^2 n^2 } 
  \sum_{i=1}^{n} \E \bigg[ \frac{1 - \pi_t^{(n)} }{ \pi_t^{(n)} } \II_{( \pi_t^{(n)} \in [f(n), 1-f(n)] )} \bigg]
   = \frac{1}{\epsilon^2 n } \E \bigg[ \frac{1 - \pi_t^{(n)} }{ \pi_t^{(n)} } \II_{( \pi_t^{(n)} \in [f(n), 1-f(n)] )} \bigg]
   \leq \frac{1}{\epsilon^2 n } \frac{1 }{ f(n) } \to 0
\end{equation*}
\normalsize
where the limit holds because we assume $f(n) = \omega(\frac{1}{n})$ so $f(n) n \to \infty$.
We can make a very similar argument for $\frac{ N_{t,0} }{ n (1-\pi_t^{(n)}) }\Pto 1$. $\qed$
%%%%%%%%%%%%%%%%%%%%%%%%%%%%
% END Lemma 1
%%%%%%%%%%%%%%%%%%%%%%%%%%%%

\paragraph{Proof for Theorem \ref{thm:bols} (Asymptotic normality of Batched OLS estimator for multi-arm bandits):}
For readability, for this proof we drop the $(n)$ superscript on $\pi_t^{(n)}$.
Note that
\begin{equation*}
\begin{bmatrix} N_{t,0} & 0 \\ 0 & N_{t,1} \end{bmatrix}^{1/2} ( \bs{\hat{\beta}}_t^{\BOLS} - \bs{\beta}_t ) 
= \begin{bmatrix} N_{t,0} & 0 \\ 0 & N_{t,1} \end{bmatrix}^{-1/2} \sum_{i=1}^{n} 
	\begin{bmatrix} 1-A_{t,i} \\ A_{t,i} \end{bmatrix} \epsilon_{t,i}.
\end{equation*}
%=  \begin{bmatrix} N_{t,0} & 0 \\ 0 & N_{t,1} \end{bmatrix}^{1/2} \begin{bmatrix}
%		\frac{ \sum_{i=1}^{n}  (1-A_{t,i} ) \epsilon_{t,i} }{ N_{t,0} } \\[10pt]
%		\frac{ \sum_{i=1}^{n}  A_{t,i} \epsilon_{t,i} }{ N_{t,1} }
%	\end{bmatrix}
%\end{equation*}
%\begin{equation*}
%%%%%%%%%%%%%%%%%%%%%%%%%%%%%%%%%%%
We want to show that
\begin{equation*}
\begin{bmatrix} \begin{bmatrix} N_{0,1} & 0 \\ 0 & N_{1,1} \end{bmatrix}^{-1/2} \sum_{i=1}^n
\begin{bmatrix} 1-A_{1,i} \\ A_{1,i} \end{bmatrix} \epsilon_{1,i} \\
\begin{bmatrix} N_{0,2} & 0 \\ 0 & N_{1,2} \end{bmatrix}^{-1/2} \sum_{i=1}^n \begin{bmatrix} 1-A_{2,i} \\ A_{2,i} \end{bmatrix} \epsilon_{2,i} \\
\vdots \\
\begin{bmatrix} N_{t,0} & 0 \\ 0 & N_{t,1} \end{bmatrix}^{-1/2} \sum_{i=1}^n \begin{bmatrix} 1-A_{T,i} \\ A_{T,i} \end{bmatrix} \epsilon_{T,i}
\end{bmatrix}
= \begin{bmatrix}
N_{0,1}^{-1/2} \sum_{i=1}^n (1-A_{1,i}) \epsilon_{1,i} \\
N_{1,1}^{-1/2} \sum_{i=1}^n \ A_{1,i} \epsilon_{1,i} \\
N_{0,2}^{-1/2} \sum_{i=1}^n (1-A_{2,i}) \epsilon_{2,i} \\
N_{1,2}^{-1/2} \sum_{i=1}^n \ A_{2,i} \epsilon_{2,i} \\
\vdots \\
N_{t,0}^{-1/2} \sum_{i=1}^n (1-A_{T,i}) \epsilon_{T,i} \\
N_{t,1}^{-1/2} \sum_{i=1}^n \ A_{T,i} \epsilon_{T,i} \\
\end{bmatrix}
\Dto \N( 0 , \sigma^2 \under{\bo{I}}_{2T} ).
\end{equation*}
%%%%%%%%%%%%%%%%%%%%%%%%%%%%%%%%%%%
By Lemma \ref{lemma:ratio} and Slutsky's Theorem it is sufficient to show that as $n \to \infty$,
\small
\begin{equation*}
\begin{bmatrix}
\frac{1}{ \sqrt{n (1-\pi_1)} } \sum_{i=1}^n (1-A_{1,i}) \epsilon_{1,i} \\
\frac{1}{ \sqrt{n \pi_1} }  \sum_{i=1}^n \ A_{1,i} \epsilon_{1,i} \\
\frac{1}{ \sqrt{n (1-\pi_2)} }  \sum_{i=1}^n (1-A_{2,i}) \epsilon_{2,i} \\
\frac{1}{ \sqrt{n \pi_2} } \sum_{i=1}^n \ A_{2,i} \epsilon_{2,i} \\
\vdots \\
\frac{1}{ \sqrt{n (1-\pi_T)} } \sum_{i=1}^n (1-A_{T,i}) \epsilon_{T,i} \\
\frac{1}{ \sqrt{n \pi_T} } \sum_{i=1}^n \ A_{T,i} \epsilon_{T,i} \\
\end{bmatrix}
= \begin{bmatrix} \frac{1}{ \sqrt{n} } \begin{bmatrix} 1 - \pi_{1,1} & 0 \\ 0 & \pi_{1,1} \end{bmatrix}^{-1/2} \sum_{i=1}^n \begin{bmatrix} 1-A_{1,i} \\ A_{1,i} \end{bmatrix} \epsilon_{1,i} \\
\frac{1}{ \sqrt{n} } \begin{bmatrix} 1 - \pi_2^{(n)} & 0 \\ 0 & \pi_2^{(n)} \end{bmatrix}^{-1/2} \sum_{i=1}^n \begin{bmatrix} 1-A_{2,i} \\ A_{2,i} \end{bmatrix} \epsilon_{2,i} \\
\vdots \\
\frac{1}{ \sqrt{n} } \begin{bmatrix} 1 - \pi_t^{(n)} & 0 \\ 0 & \pi_t^{(n)} \end{bmatrix}^{-1/2} \sum_{i=1}^n \begin{bmatrix} 1-A_{T,i} \\ A_{T,i} \end{bmatrix} \epsilon_{T,i}
\end{bmatrix}
\Dto \N( 0, \sigma^2 \under{\bo{I}}_{2T} )
\end{equation*}
\normalsize
%%%%%%%%%%%%%%%%%%%%%%%%%%%%%%%%%%%
By Cramer-Wold device, it is sufficient to show that for any fixed vector $\boldc \in \real^{2T}$ s.t. $\| \boldc \|_2 = 1$ that as $n \to \infty$,
\begin{equation*}
\boldc^\top \begin{bmatrix} 
n^{-1/2} \begin{bmatrix} 1 - \pi_{1,1} & 0 \\ 0 & \pi_{1,1} \end{bmatrix}^{-1/2} \sum_{i=1}^n \begin{bmatrix} 1-A_{1,i} \\ A_{1,i} \end{bmatrix} \epsilon_{1,i} \\
n^{-1/2} \begin{bmatrix} 1 - \pi_2^{(n)} & 0 \\ 0 & \pi_2^{(n)} \end{bmatrix}^{-1/2} \sum_{i=1}^n \begin{bmatrix} 1-A_{2,i} \\ A_{2,i} \end{bmatrix} \epsilon_{2,i} \\
\vdots \\
n^{-1/2} \begin{bmatrix} 1 - \pi_t^{(n)} & 0 \\ 0 & \pi_t^{(n)} \end{bmatrix}^{-1/2} \sum_{i=1}^n \begin{bmatrix} 1-A_{T,i} \\ A_{T,i} \end{bmatrix} \epsilon_{T,i}
\end{bmatrix} \Dto \N( 0 , \sigma^2 )
\end{equation*}
%%%%%%%%%%%%%%%%%%%%%%%%%%%%%%%%%%%
Let us break up $\boldc$ so that $\boldc = [\boldc_1, \boldc_2, ..., \boldc_T]^\top \in \real^{2T}$ with $\boldc_t \in \real^2$ for $t \in [1 \colon T]$.
The above is equivalent to
\begin{equation*}
\sum_{t=1}^T n^{-1/2} \boldc_t^\top \begin{bmatrix} 1 - \pi_t^{(n)} & 0 \\ 0 & \pi_t^{(n)} \end{bmatrix}^{-1/2} 
\sum_{i=1}^{n} \begin{bmatrix} 1-A_{t,i} \\ A_{t,i} \end{bmatrix} \epsilon_{t,i}
\Dto \N( 0 , \sigma^2 )
\end{equation*}
%%%%%%%%%%%%%%%%%%%%%%%%%%%%%%%%%%%
Let us define $Y_{t,i} := n^{-1/2} \boldc_t^\top
\begin{bmatrix} 1 - \pi_{t,i} & 0 \\ 0 & \pi_{t,i} \end{bmatrix}^{-1/2} \begin{bmatrix} 1-A_{t,i} \\ A_{t,i} \end{bmatrix} \epsilon_{t,i}$. \\
The sequence $\{ Y_{1,1}, Y_{1,2}, ..., Y_{1,n}, ..., Y_{T,1}, Y_{T,2}, ..., Y_{T,n} \}$
is a martingale with respect to sequence of histories $\{ H_{t}^{(n)} \}_{t=1}^T$, since 
\begin{equation*}
\E[ Y_{t,i} | H_{t-1}^{(n)}] = n^{-1/2} \boldc_t^\top 
\begin{bmatrix} 1 - \pi_t^{(n)} & 0 \\ 0 & \pi_t^{(n)} \end{bmatrix}^{-1/2} 
 \E \bigg[ \begin{bmatrix} 1-A_{t,i} \\ A_{t,i} \end{bmatrix} \epsilon_{t,i} \bigg| H_{t-1}^{(n)} \bigg]
\end{equation*}
\begin{equation*}
= n^{-1/2} \boldc_t^\top 
\begin{bmatrix} 1 - \pi_t^{(n)} & 0 \\ 0 & \pi_t^{(n)} \end{bmatrix}^{-1/2} 
 \E \bigg[ \begin{bmatrix} (1 - \pi_t^{(n)} ) E[ \epsilon_{t,i} | H_{t-1}^{(n)}, A_{t,i} = 0 ] \\
\pi_{t,i} E[ \epsilon_{t,i} | H_{t-1}^{(n)}, A_{t,i} = 1] \end{bmatrix} \bigg| H_{t-1}^{(n)} \bigg] = 0
\end{equation*}
for all $i \in [1 \colon n]$ and all $t \in [1 \colon T]$. 
We then apply \cite{dvoretzky1972asymptotic} martingale central limit theorem to $Y_{t,i}$ to show the desired result (see the proof of Theorem \ref{thm:triangularCLT} in Appendix \ref{appendix:triangularCLT} for the statement of the martingale CLT conditions).

%Note that the have already satisfied the first condition because $\{ Y_{t,1}^{(n)}, Y_{t21}^{(n)}, ..., Y_{t,n}^{(n)} \}_{t=1}^T$ is a martingale difference sequence.
%Note that $\big\{ \epsilon_{t,i}^{(n)} : i \in [1 \colon n] \big\}_{t=1}^T$ are also a martingale difference array with respect to filtration $\{ \HH_{t}^{(n)} \}_{t=1}^T$, where $\HH_t^{(n)} = \sigma( H_t )$ because
%$$\E[ \epsilon_{t,i}^{(n)} | \HH_{t-1}^{(n)} ] = \E \bigg[ \E \big[ \epsilon_{t,i}^{(n)} \big| \HH_{t-1}^{(n)}, A_{t,i}^{(n)} \big] \bigg| \HH_{t-1}^{(n)} \bigg] = 0$$
%Note that the actions $A_{t,i}$ are \textit{non-anticipating} with respect to filtration $\G_t^{(n)}$, that is, $A_{t,i}^{(n)} \in \G_{t-1}^{(n)}$; however, $A_{t,i}^{(n)} \notin \HH_{t-1}^{(n)}$.  

\paragraph{Condition(a): Martingale Condition}
The first condition holds because $\E[ Y_{t,i} | H_{t-1}^{(n)} ] = 0$ for all $i \in [1 \colon n]$ and all $t \in [1 \colon T]$.

\paragraph{Condition(b): Conditional Variance}
\begin{equation*}
\sum_{t=1}^T \sum_{i=1}^{n} E[ Y_{n,t,i}^2 | H_{t-1}^{(n)}]
= \sum_{t=1}^T \sum_{i=1}^{n} 
\E \bigg[ \bigg( \frac{1}{ \sqrt{n}} \boldc_t^\top 
\begin{bmatrix} 1 - \pi_t^{(n)} & 0 \\ 0 & \pi_t^{(n)} \end{bmatrix}^{-1/2} 
\begin{bmatrix} 1-A_{t,i} \\ A_{t,i} \end{bmatrix} \epsilon_{t,i} \bigg)^2 
\bigg| H_{t-1}^{(n)} \bigg]
\end{equation*}
%%%%%%%%%%%%%%%%%%%%%%%%%%%%%%%%%%%
\begin{equation*}
= \sum_{t=1}^T \sum_{i=1}^{n}
\E \bigg[ \frac{1}{n} \boldc_t^\top 
\begin{bmatrix} 1 - \pi_t^{(n)} & 0 \\ 0 & \pi_t^{(n)} \end{bmatrix}^{-1/2} 
\begin{bmatrix} 1-A_{t,i} & 0 \\ 0 & A_{t,i} \end{bmatrix} 
\begin{bmatrix} 1 - \pi_t^{(n)} & 0 \\ 0 & \pi_t^{(n)} \end{bmatrix}^{-1/2} 
\boldc_t \epsilon_{t,i}^2 \bigg| H_{t-1}^{(n)} \bigg]
\end{equation*}
\small
\begin{equation*}
= \sum_{t=1}^T \sum_{i=1}^{n}
\frac{1}{n} \boldc_t^\top 
\begin{bmatrix} 1 - \pi_t^{(n)} & 0 \\ 0 & \pi_t^{(n)} \end{bmatrix}^{-1/2} 
\begin{bmatrix} \E[ (1-A_{t,i}) \epsilon_{t,i}^2 | H_{t-1}^{(n)} ] & 0 \\
 0 & \E[ A_{t,i} \epsilon_{t,i}^2 | H_{t-1}^{(n)} ] \end{bmatrix}
\begin{bmatrix} 1 - \pi_t^{(n)} & 0 \\ 0 & \pi_t^{(n)} \end{bmatrix}^{-1/2} 
\boldc_t
\end{equation*}
\normalsize
%%%%%%%%%%%%%%%%%%%%%%%%%%%%%%%%%%%
Since $\E[ A_{t,i} \epsilon_{t,i}^2 | H_{t-1}^{(n)} ] 
= \pi_t^{(n)} \E[ \epsilon_{t,i}^2 | H_{t-1}^{(n)}, A_{t,i} = 1 ] 
= \sigma^2  \pi_t$
and 
$\E[ (1-A_{t,i}) \epsilon_{t,i}^2 | H_{t-1}^{(n)} ] 
= (1-\pi_t) \E[ \epsilon_{t,i}^2 | H_{t-1}^{(n)}, A_{t,i} = 0 ]
= \sigma^2 (1-\pi_t)$,
\begin{equation*}
= \sum_{t=1}^T \sum_{i=1}^{n}
n^{-1} \boldc_t^\top 
\boldc_t \sigma^2
= \sum_{t=1}^T \boldc_t^\top 
\boldc_t \sigma^2
= \sigma^2
\end{equation*}

\paragraph{Condition(c): Lindeberg Condition}
Let $\delta > 0$.
\footnotesize
\begin{equation*}
\sum_{t=1}^T \sum_{i=1}^{n} E \big[ Y_{t,i}^2  \II_{( Y_{t,i}^{2} > \delta^2 )} \big| H_{t-1}^{(n)} \big]
= \sum_{t=1}^T \sum_{i=1}^{n} 
\E \bigg[ \bigg( n^{-1/2} \boldc_t^\top 
\begin{bmatrix} 1 - \pi_t^{(n)} & 0 \\ 0 & \pi_t^{(n)} \end{bmatrix}^{-1/2} 
\begin{bmatrix} 1-A_{t,i} \\ A_{t,i} \end{bmatrix} \epsilon_{t,i} \bigg)^2 
\II_{( Y_{t,i}^{2} > \delta^2 )} \bigg| H_{t-1}^{(n)} \bigg]
\end{equation*}
\begin{equation*}
= \sum_{t=1}^T \frac{1}{n} \sum_{i=1}^{n} 
\E \bigg[ \boldc_t^\top 
\begin{bmatrix} 1 - \pi_t^{(n)} & 0 \\ 0 & \pi_t^{(n)} \end{bmatrix}^{-1/2} 
\begin{bmatrix} 1-A_{t,i} & 0 \\ 0 & A_{t,i} \end{bmatrix} 
\begin{bmatrix} 1 - \pi_t^{(n)} & 0 \\ 0 & \pi_t^{(n)} \end{bmatrix}^{-1/2} \boldc_t \epsilon_{t,i}^2 
\II_{( Y_{t,i}^{2} > \delta^2 )} \bigg| H_{t-1}^{(n)} \bigg]
\end{equation*}
\begin{multline*}
= \sum_{t=1}^T \frac{1}{n} \sum_{i=1}^n \boldc_t^\top 
\begin{bmatrix} 1 - \pi_t^{(n)} & 0 \\ 0 & \pi_t^{(n)} \end{bmatrix}^{- \frac{1}{2} } \\
\begin{bmatrix} \E[ (1-A_{t,i}) \epsilon_{t,i}^2 \II_{( Y_{t,i}^{2} > \delta^2 )} | H_{t-1}^{(n)} ] & 0 \\
 0 & \E[ A_{t,i} \epsilon_{t,i}^2 \II_{( Y_{t,i}^{2} > \delta^2 )} | H_{t-1}^{(n)} ] \end{bmatrix} \\
\begin{bmatrix} 1 - \pi_t^{(n)} & 0 \\ 0 & \pi_t^{(n)} \end{bmatrix}^{- \frac{1}{2} } \boldc_t
\end{multline*}
\normalsize
Note that for $\boldc_t = [ c_{t,0}, c_{t,1} ]^\top$,
$\E \big[ (1-A_{t,i}) \epsilon_{t,i}^2 \II_{( Y_{t,i}^{2} > \delta^2 )} \big| H_{t-1}^{(n)} \big]
= \E \bigg[ \epsilon_{t,i}^2 \II_{ \big( \frac{ c_{t,0}^2 }{ 1-\pi_t^{(n)} } \epsilon_{t,i}^2 > n \delta^2 \big)} \bigg| H_{t-1}^{(n)}, A_{t,i} = 0 \bigg] (1-\pi_t)$
and $\E \big[ A_{t,i} \epsilon_{t,i}^2 \II_{( Y_{t,i}^{2} > \delta^2 )} \big| H_{t-1}^{(n)} \big]
= \E \bigg[ \epsilon_{t,i}^2 \II_{ \big( \frac{ c_{t,1}^2 }{ \pi_t^{(n)} } \epsilon_{t,i}^2 > n \delta^2 \big)} \bigg| H_{t-1}^{(n)}, A_{t,i} = 1 \bigg] \pi_t$.
Thus, we have that
\small
\begin{equation*}
= \sum_{t=1}^T \frac{1}{n} \sum_{i=1}^n c_{t,0}^2 \E \bigg[ \epsilon_{t,i}^2 \II_{ \big( \epsilon_{t,i}^2 > \frac{ n \delta^2  (1-\pi_t) }{ c_{t,0}^2 } \big) } \bigg| H_{t-1}^{(n)}, A_{t,i} = 0 \bigg] 
 + c_{t,1}^2 \E \bigg[ \epsilon_{t,i}^2 \II_{ \big( \epsilon_{t,i}^2 > \frac{ n \delta^2 \pi_t^{(n)} }{ c_{t,1}^2 } \big) } \bigg| H_{t-1}^{(n)}, A_{t,i} = 1 \bigg]
 \end{equation*}
\begin{equation*}
\leq \sum_{t=1}^T \max_{i \in [1 \colon n]} \bigg\{ c_{t,0}^2 \E \bigg[ \epsilon_{t,i}^2 \II_{ \big( \epsilon_{t,i}^2 > \frac{ n \delta^2  (1-\pi_t) }{ c_{t,0}^2 } \big) } \bigg| H_{t-1}^{(n)}, A_{t,i} = 0 \bigg] 
 + c_{t,1}^2 \E \bigg[ \epsilon_{t,i}^2 \II_{ \big( \epsilon_{t,i}^2 > \frac{ n \delta^2 \pi_t^{(n)} }{ c_{t,1}^2 } \big) } \bigg| H_{t-1}^{(n)}, A_{t,i} = 1 \bigg] \bigg\}
 \end{equation*}
 \normalsize
 %%%%%%%%%%%%%%%%%%%%%%%%%%%%%%%%%%%
 
Note that for any $t \in [1 \colon T]$ and $i \in [1 \colon n]$,
\begin{equation*}
\E \bigg[ \epsilon_{t,i}^2 \II_{ \big( \epsilon_{t,i}^2 > \frac{ n \delta^2 \pi_t^{(n)} }{ c_{t,1}^2 } \big) } \bigg| H_{t-1}^{(n)}, A_{t,i} = 1 \bigg]
\end{equation*}
\begin{equation*}
= \E \bigg[ \epsilon_{t,i}^2 \II_{ \big( \epsilon_{t,i}^2 > \frac{ n \delta^2 \pi_t^{(n)} }{ c_{t,1}^2 } \big) } \bigg| H_{t-1}^{(n)}, A_{t,i} = 1 \bigg] 
\bigg( \II_{ ( \pi_t^{(n)} \in [f(n), 1-f(n)] ) } + \II_{ ( \pi_t^{(n)} \not\in [f(n), 1-f(n)] ) } \bigg)
\end{equation*}
\begin{equation*}
\leq \E \bigg[ \epsilon_{t,i}^2 \II_{ \big( \epsilon_{t,i}^2 > \frac{ n \delta^2 f(n) }{ c_{t,1}^2 } \big) } \bigg| H_{t-1}^{(n)}, A_{t,i} = 1 \bigg] 
+ \sigma^2 \II_{ ( \pi_t^{(n)} \not\in [f(n), 1-f(n)] ) } 
\end{equation*}
The second term converges in probability to zero as $n \to \infty$ by our clipping assumption.
We now show how the first term goes to zero in probability.
Since we assume $f(n) = \omega(\frac{1}{n})$, $n f(n) \to \infty$.
So, it is sufficient to show that for all $t, n$,
\begin{equation*}
\lim_{m \to \infty} \max_{i \in [1 \colon n]} \bigg\{ \E \bigg[ \epsilon_{t,i}^2 \II_{ ( \epsilon_{t,i}^2 > m ) } \bigg| H_{t-1}^{(n)}, A_{t,i} = 1 \bigg] \bigg\} = 0
\end{equation*}
%%%%%%%%%%%%%%%%%%%%%%%%%%%%%%%%%%%
By Condition \ref{cond:weakmoments}, we have that for all $n \geq 1$,
\begin{equation*}
\max_{t \in [1 \colon T], i \in [1 \colon n]} \E[ \varphi( \epsilon_{t,i}^2 ) | H_{t-1}^{(n)}, A_{t,i} = 1 ] < M
\end{equation*}
Since we assume that $\lim_{x \to \infty} \frac{ \varphi(x) }{x} = \infty$, for all $m$, there exists a $b_m$ s.t. $\varphi(x) \geq m M x$ for all $x \geq b_m$. 
So, for all $n, t, i$,
\begin{equation*}
M \geq \E[ \varphi( \epsilon_{t,i}^2 ) | H_{t-1}^{(n)}, A_{t,i} = 1] \geq \E[ \varphi( \epsilon_{t,i}^2 ) \II_{( \epsilon_{t,i}^2 \geq b_m )} | H_{t-1}^{(n)}, A_{t,i} = 1 ] 
\end{equation*}
\begin{equation*}
\geq m M \E[ \epsilon_{t,i}^2 \II_{( \epsilon_{t,i}^2 \geq b_m )} | H_{t-1}^{(n)}, A_{t,i} = 1 ]
\end{equation*}
Thus, 
\begin{equation*}
\max_{t \in [1 \colon T], i \in [1 \colon n]} \E[ \epsilon_{t,i}^2 \II_{( \epsilon_{t,i}^2 \geq b_m )} | H_{t-1}^{(n)}, A_{t,i} =1 ] \leq \frac{1}{m}
\end{equation*}
We can make a very similar argument that for all $t \in [1 \colon T]$,  as $n \to \infty$,
\begin{equation*}
\max_{i \in [1 \colon n] } \E \bigg[ \epsilon_{t,i}^2 \II_{ \big( \epsilon_{t,i}^2 > \frac{ n \delta^2  (1-\pi_t) }{ c_{t,0}^2 } \big) } \bigg| H_{t-1}^{(n)}, A_{t,i} = 0 \bigg] \Pto 0 ~~~~ \qed
\end{equation*}

%\clearpage
\begin{corollary}[Asymptotic Normality of the Batched OLS Estimator of Margin; two-arm bandit setting]
\label{corollary:BOLSmargin}
Assume the same conditions as Theorem \ref{thm:bols}.
For each $t \in [1 \colon T]$, we have the BOLS estimator of the margin $\beta_1 - \beta_0$:
\begin{equation*}
\hat{\Delta}_t^{\BOLS} = \frac{ \sum_{i=1}^{n} (1-A_{t,i} ) R_{t,i} }{ N_{t,0} } - \frac{ \sum_{i=1}^{n} A_{t,i} R_{t,i} }{ N_{t,1} }
\end{equation*}
We show that as $n \to \infty$,
\begin{equation*}
\begin{bmatrix} 
\sqrt{ \frac{ N_{1,0} N_{1,1} }{ n } } ( \hat{\Delta}_1^{\BOLS} - \Delta_1 )  \\
\sqrt{ \frac{ N_{2,0} N_{2,1} }{ n } } ( \hat{\Delta}_2^{\BOLS} - \Delta_2 )   \\
\vdots \\
\sqrt{ \frac{ N_{T,0} N_{T,1} }{ n } } ( \hat{\Delta}_T^{\BOLS} - \Delta_T )   \\
 \end{bmatrix}
\Dto \N(0, \sigma^2 \under{\bo{I}}_T )
\end{equation*}
\end{corollary}

\paragraph{Proof:}
\begin{equation*}
\sqrt{ \frac{ N_{t,0} N_{t,1} }{ n } } ( \hat{\Delta}_t^{\BOLS} - \Delta_t ) 
= \sqrt{ \frac{ N_{t,0} N_{t,1} }{ n } } \bigg( \frac{ \sum_{i=1}^{n} (1-A_{t,i} ) \epsilon_{t,i} }{ N_{t,0} } - \frac{ \sum_{i=1}^{n} A_{t,i} \epsilon_{t,i} }{ N_{t,1} }\bigg)
\end{equation*}
\begin{equation*}
= \sqrt{ \frac{ N_{t,1} }{ n } } \frac{ \sum_{i=1}^{n} (1-A_{t,i} ) \epsilon_{t,i} }{ \sqrt{ N_{t,0} } } - 
\sqrt{ \frac{ N_{t,0} }{ n } } \frac{ \sum_{i=1}^{n} A_{t,i} \epsilon_{t,i} }{ \sqrt{ N_{t,1} } }
\end{equation*}
\begin{equation*}
= \begin{bmatrix} \sqrt{ \frac{ N_{t,1} }{ n } } & - \sqrt{ \frac{ N_{t,0} }{ n } } \end{bmatrix} 
	\begin{bmatrix} N_{t,0} & 0 \\ 0 & N_{t,1} \end{bmatrix}^{-1/2} \sum_{i=1}^{n} 
	\begin{bmatrix} 1-A_{t,i} \\ A_{t,i} \end{bmatrix} \epsilon_{t,i}
\end{equation*}
%%%%%%%%%%%%%%%%%%%%%%%%%%%%%%%%%%%
 By Slutsky's Theorem and Lemma \ref{lemma:ratio}, it is sufficient to show that as $n \to \infty$,
 \small
\begin{equation*}
\begin{bmatrix} 
   \frac{1}{ \sqrt{n} } \begin{bmatrix} \sqrt{ \pi_1^{(n)} } & - \sqrt{ 1-\pi_1^{(n)} } ~ \end{bmatrix} 
	\begin{bmatrix} 1-\pi_1^{(n)} & 0 \\ 0 & \pi_1^{(n)} \end{bmatrix}^{-1/2} \sum_{i=1}^{n} 
	\begin{bmatrix} 1-A_{1,i} \\ A_{1,i} \end{bmatrix} \epsilon_{1,i}  \\
  \frac{1}{ \sqrt{n} } \begin{bmatrix} \sqrt{ \pi_2^{(n)} } & - \sqrt{ 1-\pi_2^{(n)} } ~ \end{bmatrix} 
	\begin{bmatrix} 1-\pi_2^{(n)} & 0 \\ 0 & \pi_2^{(n)} \end{bmatrix}^{-1/2} \sum_{i=1}^{n} 
	\begin{bmatrix} 1-A_{2,i} \\ A_{2,i} \end{bmatrix} \epsilon_{2,i}   \\
\vdots \\
  \frac{1}{ \sqrt{n} } \begin{bmatrix} \sqrt{ \pi_t^{(n)} } & - \sqrt{ 1-\pi_t^{(n)} } ~ \end{bmatrix} 
	\begin{bmatrix} 1-\pi_t^{(n)} & 0 \\ 0 & \pi_t^{(n)} \end{bmatrix}^{-1/2} \sum_{i=1}^{n} 
	\begin{bmatrix} 1-A_{T,i} \\ A_{T,i} \end{bmatrix} \epsilon_{T,i}   \\
 \end{bmatrix} \Dto \N( 0 , \sigma^2 \under{\bo{I}}_T )
 \end{equation*}
 \normalsize
%%%%%%%%%%%%%%%%%%%%%%%%%%%%%%%%%%%
By Cramer-Wold device, it is sufficient to show that for any fixed vector $\boldd \in \real^{T}$ s.t. $\| \boldd \|_2 = 1$ that
\small
\begin{equation*}
\boldd^\top \begin{bmatrix} 
   \frac{1}{ \sqrt{n} }\begin{bmatrix} \sqrt{ \pi_1^{(n)} } & - \sqrt{ 1-\pi_1^{(n)} } ~ \end{bmatrix} 
	\begin{bmatrix} 1-\pi_1^{(n)} & 0 \\ 0 & \pi_1^{(n)} \end{bmatrix}^{-1/2} \sum_{i=1}^{n} 
	\begin{bmatrix} 1-A_{1,i} \\ A_{1,i} \end{bmatrix} \epsilon_{1,i}  \\
  \frac{1}{ \sqrt{n} } \begin{bmatrix} \sqrt{ \pi_2^{(n)} } & - \sqrt{ 1-\pi_2^{(n)} } ~ \end{bmatrix} 
	\begin{bmatrix} 1-\pi_2^{(n)} & 0 \\ 0 & \pi_2^{(n)} \end{bmatrix}^{-1/2} \sum_{i=1}^{n} 
	\begin{bmatrix} 1-A_{2,i} \\ A_{2,i} \end{bmatrix} \epsilon_{2,i}   \\
\vdots \\
  \frac{1}{ \sqrt{n} } \begin{bmatrix} \sqrt{ \pi_t^{(n)} } & - \sqrt{ 1-\pi_t^{(n)} } ~ \end{bmatrix} 
	\begin{bmatrix} 1-\pi_t^{(n)} & 0 \\ 0 & \pi_t^{(n)} \end{bmatrix}^{-1/2} \sum_{i=1}^{n} 
	\begin{bmatrix} 1-A_{T,i} \\ A_{T,i} \end{bmatrix} \epsilon_{T,i}   \\
 \end{bmatrix} \Dto \N( 0 , \sigma^2 )
 \end{equation*}
 \normalsize
%%%%%%%%%%%%%%%%%%%%%%%%%%%%%%%%%%%
Let $[d_1, d_2, ..., d_T]^\top := \boldd \in \real^{T}$.
The above is equivalent to
\begin{equation*}
\sum_{t=1}^T \frac{1}{ \sqrt{n} } d_t \begin{bmatrix} \sqrt{ \pi_t^{(n)} } & - \sqrt{ 1-\pi_t^{(n)} } ~ \end{bmatrix} 
	\begin{bmatrix} 1-\pi_t^{(n)} & 0 \\ 0 & \pi_t^{(n)} \end{bmatrix}^{-1/2} \sum_{i=1}^{n} 
	\begin{bmatrix} 1-A_{t,i} \\ A_{t,i} \end{bmatrix} \epsilon_{t,i}
\Dto \N( 0 , \sigma^2 )
\end{equation*}
%%%%%%%%%%%%%%%%%%%%%%%%%%%%%%%%%%%
Define $Y_{t,i} := \frac{1}{ \sqrt{n} } d_t \begin{bmatrix} \sqrt{ \pi_t^{(n)} } & - \sqrt{ 1-\pi_t^{(n)} } ~ \end{bmatrix} 
\begin{bmatrix} 1 - \pi_t^{(n)} & 0 \\ 0 & \pi_t^{(n)} \end{bmatrix}^{-1/2} \begin{bmatrix} 1-A_{t,i} \\ A_{t,i} \end{bmatrix} \epsilon_{t,i}$. \\
$\{ Y_{1,1}, Y_{1,2}, ..., Y_{1,n}, ..., Y_{T,1}, Y_{T,2}, ..., Y_{T,n} \}$ is a martingale difference array with respect to the sequence of histories $\{ H_t^{(n)} \}_{t=1}^T$ because for all $i \in [1 \colon n]$ and $t \in [1 \colon T]$,
 \footnotesize
\begin{equation*}
\E[ Y_{t,i} | H_{t-1}^{(n)}] = \frac{1}{ \sqrt{n} } d_t \begin{bmatrix} \sqrt{ \pi_t^{(n)} } & - \sqrt{ 1-\pi_t^{(n)} } ~ \end{bmatrix} 
\begin{bmatrix} 1 - \pi_t^{(n)} & 0 \\ 0 & \pi_t^{(n)} \end{bmatrix}^{-1/2} 
 \E \bigg[ \begin{bmatrix} 1-A_{t,i} \\ A_{t,i} \end{bmatrix} \epsilon_{t,i} \bigg| H_{t-1}^{(n)} \bigg]
 \end{equation*}
  \footnotesize
\begin{equation*}
= \frac{d_t }{ \sqrt{n} } \begin{bmatrix} \sqrt{ \pi_t^{(n)} } & - \sqrt{ 1-\pi_t^{(n)} } \end{bmatrix} 
\begin{bmatrix} 1 - \pi_t^{(n)} & 0 \\ 0 & \pi_t^{(n)} \end{bmatrix}^{-\frac{1}{2}} 
 \E \bigg[  \begin{bmatrix} (1 - \pi_t^{(n)} ) \E[ \epsilon_{t,i} | H_{t-1}^{(n)}, A_{t,i} = 0 ] \\
\pi_{t,i} \E[ \epsilon_{t,i} | H_{t-1}^{(n)}, A_{t,i} = 1] \end{bmatrix} \bigg| H_{t-1}^{(n)} \bigg] = 0
\end{equation*} 
\normalsize
We now apply \cite{dvoretzky1972asymptotic} martingale central limit theorem to $Y_{t,i}$ to show the desired result.
Verifying the conditions for the martingale CLT is equivalent to what we did to verify the conditions in the conditions in the proof of Theorem \ref{thm:bols}---the only difference is that we replace $\boldc_t^\top$ in the Theorem \ref{thm:bols} proof with $d_t \begin{bmatrix} \sqrt{ 1-\pi_t^{(n)} } & - \sqrt{ \pi_t^{(n)} } ~ \end{bmatrix}$ in this proof.
Even though $\boldc_t$ is a constant vector and $d_t \begin{bmatrix} \sqrt{ 1-\pi_t^{(n)} } & - \sqrt{ \pi_t^{(n)} } ~ \end{bmatrix}$ is a random vector, the proof still goes through with this adjusted $\boldc_t$ vector, since 
(i) $d_t \begin{bmatrix} \sqrt{ 1-\pi_t^{(n)} } & - \sqrt{ \pi_t^{(n)} } ~ \end{bmatrix} \in H_{t-1}^{(n)}$, 
(ii) $\| \begin{bmatrix} \sqrt{ 1-\pi_t^{(n)} } & - \sqrt{ \pi_t^{(n)} } ~ \end{bmatrix} \|_2 = 1$, and 
(iii) $\frac{ n \delta^2 \pi_t^{(n)} }{ c_{t,1}^2 } = \frac{ n \delta^2 \pi_t^{(n)} }{ d_t^2 \pi_t^{(n)} } \to \infty$ and $\frac{ n \delta^2 (1-\pi_t) }{ c_{t,0}^2 } = \frac{ n \delta^2 (1-\pi_t) }{ d_t^2 (1-\pi_t) } \to \infty$. $\qed$

%\medskip \medskip
%\paragraph{Proof of Corollary \ref{corollary:nonstationary} (Confidence interval for treatment effect for non-stationary bandits)}  
%~~ \\
%Note that by Corollary \ref{corollary:BOLSmargin}, 
%\begin{equation*}
%\PP \big( \TN{exists some } t \in [1:T] ~ s.t. ~ \Delta_t \notin \boldL_t \big) 
%\leq \sum_{t=1}^T \PP \big( \Delta_t \notin \boldL_t \big) 
%\to \sum_{t=1}^T \frac{\alpha}{T}
%= \alpha
%\end{equation*}
%where the limit is as $n \to \infty$. Since
%\begin{equation*}
%\PP \big( \forall t \in [1:T], \Delta_t \in \boldL_t  \big)
%= 1 - \PP \big( \TN{exists some } t \in [1:T] ~ s.t. ~ \Delta_t \notin \boldL_t \big)
%\end{equation*}
%Thus,
%\begin{equation*}
%\lim_{n \to \infty} \PP \big( \forall t \in [1:T], \Delta_t \in \boldL_t \big) \geq 1 - \alpha ~~~ \qed
%\end{equation*}'

\clearpage
\begin{corollary}[Consistency of BOLS Variance Estimator]
\label{corollary:varianceconsistency}
Assuming Conditions \ref{cond:moments} (moments) and \ref{cond:condiid} (conditionally i.i.d. actions), and a clipping rate of $f(n) = \omega(\frac{1}{n})$ (Definition \ref{def:clipping}), for all $t \in [1 \colon T]$, as $n \to \infty$,
\begin{equation*}
	\hat{ \sigma }_t^2 = \frac{1}{n-2} \sum_{i=1}^n \bigg( R_{t,i} - A_{t,i} \betahat_{t,1}^{\TN{BOLS}} - (1-A_{t,i}) \betahat_{t,0}^{\TN{BOLS}} \bigg)^2 \Pto \sigma^2
\end{equation*}
%for $\hat{ \sigma }_t^2$ for BOLS as defined in Appendix \ref{appendix:noisevar}.
\end{corollary}

\paragraph{Proof:}
\begin{equation*}
	\hat{ \sigma }_t^2 = \frac{1}{n-2} \sum_{i=1}^n \bigg( R_{t,i} - A_{t,i} \betahat_{t,1}^{\TN{BOLS}} - (1-A_{t,i}) \betahat_{t,0}^{\TN{BOLS}} \bigg)^2
\end{equation*}
\begin{equation*}
	= \frac{1}{n-2} \sum_{i=1}^n \bigg( \bigg[ A_{t,i} \beta_{t,1} + (1-A_{t,i}) \beta_{t,0} + \epsilon_{t,i} \bigg] 
		- A_{t,i} \bigg[ \beta_{t,1} + \frac{ \sum_{i=1}^n A_{t,i} \epsilon_{t,i} }{ N_{t,1} } \bigg] 
		- (1-A_{t,i}) \bigg[ \beta_{t,0} + \frac{ \sum_{i=1}^n (1-A_{t,i}) \epsilon_{t,i} }{ N_{t,0} } \bigg]  \bigg)^2
\end{equation*}
\begin{equation*}
	= \frac{1}{n-2} \sum_{i=1}^n \bigg( \epsilon_{t,i}
		- A_{t,i} \frac{ \sum_{i=1}^n A_{t,i} \epsilon_{t,i} }{ N_{t,1} } 
		- (1-A_{t,i}) \frac{ \sum_{i=1}^n (1-A_{t,i}) \epsilon_{t,i} }{ N_{t,0} } \bigg)^2
\end{equation*}
\begin{multline*}
	= \frac{1}{n-2} \sum_{i=1}^n \bigg( \epsilon_{t,i}^2
		- 2 A_{t,i} \epsilon_{t,i} \frac{ \sum_{i=1}^n A_{t,i} \epsilon_{t,i} }{ N_{t,1} } 
		- 2 (1-A_{t,i}) \epsilon_{t,i} \frac{ \sum_{i=1}^n (1-A_{t,i}) \epsilon_{t,i} }{ N_{t,0} }  \\
		+ A_{t,i} \bigg[ \frac{ \sum_{i=1}^n A_{t,i} \epsilon_{t,i} }{ N_{t,1} } \bigg]^2
		+ (1-A_{t,i}) \bigg[ \frac{ \sum_{i=1}^n (1-A_{t,i}) \epsilon_{t,i} }{ N_{t,0} } \bigg]^2 \bigg)
\end{multline*}
\begin{multline*}
	= \bigg( \frac{1}{n-2} \sum_{i=1}^n \epsilon_{t,i}^2 \bigg)
		- 2 \frac{ ( \sum_{i=1}^n A_{t,i} \epsilon_{t,i} )^2 }{ (n-2) N_{t,1} } 
		- 2 \frac{ ( \sum_{i=1}^n (1-A_{t,i}) \epsilon_{t,i} )^2 }{ (n-2) N_{t,0} }   \\
		+ \frac{ N_{t,1} }{ n-2 } \bigg[ \frac{ \sum_{i=1}^n A_{t,i} \epsilon_{t,i} }{ N_{t,1} } \bigg]^2
		+ \frac{ N_{t,0} }{ n-2 } \bigg[ \frac{ \sum_{i=1}^n (1-A_{t,i}) \epsilon_{t,i} }{ N_{t,0} } \bigg]^2
\end{multline*}
\begin{equation*}
	= \bigg( \frac{1}{n-2} \sum_{i=1}^n \epsilon_{t,i}^2 \bigg)
		- \frac{ ( \sum_{i=1}^n A_{t,i} \epsilon_{t,i} )^2 }{ (n-2) N_{t,1} } 
		- \frac{ ( \sum_{i=1}^n (1-A_{t,i}) \epsilon_{t,i} )^2 }{ (n-2) N_{t,0} } 
\end{equation*}
Note that $\frac{1}{n-2} \sum_{i=1}^n \epsilon_{t,i}^2 \Pto \sigma^2$ because for all $\delta > 0$,
\begin{equation*}
	\PP \bigg( \bigg| \bigg[ \frac{1}{n-2} \sum_{i=1}^n \epsilon_{t,i}^2 \bigg] - \sigma^2 \bigg| > \delta \bigg)
	\leq \PP \bigg( \bigg| \bigg[ \frac{1}{n-2} \sum_{i=1}^n \epsilon_{t,i}^2 \bigg] - \frac{ \sigma^2 (n-2) }{n} \bigg| > \delta/2 \bigg)
	+ \PP \bigg( \bigg| \frac{ \sigma^2 (n-2) }{n} - \sigma^2 \bigg| > \delta/2 \bigg)
\end{equation*}
\begin{equation*}
	= \PP \bigg( \bigg| \frac{1}{n-2} \sum_{i=1}^n ( \epsilon_{t,i}^2 - \sigma^2 ) \bigg| > \delta/2 \bigg)
	+ \PP \bigg( \sigma^2 \bigg| \frac{ -2 }{n} \bigg| > \delta/2 \bigg)
\end{equation*}
Since the second term in the summation above goes to zero for sufficiently large $n$, we now focus on the first term in the summation above. By Chebychev inequality,
\begin{equation*}
	\PP \bigg( \bigg| \frac{1}{n-2} \sum_{i=1}^n ( \epsilon_{t,i}^2 - \sigma^2 ) \bigg| > \delta/2 \bigg)
	\leq \frac{4}{ \delta^2 (n-2)^2} \E \bigg[ \sum_{i=1}^n \sum_{j=1}^n ( \epsilon_{t,i}^2 - \sigma^2 ) ( \epsilon_{t,j}^2 - \sigma^2 ) \bigg]
	= \frac{4}{ \delta^2 (n-2)^2} \E \bigg[ \sum_{i=1}^n ( \epsilon_{t,i}^2 - \sigma^2 )^2 \bigg]
\end{equation*}
where the equality above holds because for $i \not= j$, $\E[ ( \epsilon_{t,i}^2 - \sigma^2 ) ( \epsilon_{t,j}^2 - \sigma^2 ) ]
= \E\big[ \E[( \epsilon_{t,i}^2 - \sigma^2 ) ( \epsilon_{t,j}^2 - \sigma^2 ) | \HH_{t-1}^{(n)}] \big]
= \E\big[ \E[ \epsilon_{t,i}^2 - \sigma^2 | \HH_{t-1}^{(n)}] \E[ \epsilon_{t,j}^2 - \sigma^2 | \HH_{t-1}^{(n)}] \big] = 0$.
By Condition \ref{cond:moments} $\E[  \epsilon_{t,i}^4 | \HH_{t=1}^{(n)} ] < M < \infty$,
\begin{equation*}
	= \frac{4}{ \delta^2 (n-2)^2} \E \bigg[ \sum_{i=1}^n \E[ ( \epsilon_{t,i}^4 - 2 \epsilon_{t,i}^2 \sigma^2 + \sigma^4) | \HH_{t-1}^{(n)} ] \bigg]
	\leq \frac{4 n (M + \sigma^4)}{ \delta^2 (n-2)^2}  \to 0
\end{equation*}

Thus by Slutsky's Theorem it is sufficient to show that $\frac{ ( \sum_{i=1}^n A_{t,i} \epsilon_{t,i} )^2 }{ (n-2) N_{t,1} } + \frac{ ( \sum_{i=1}^n (1-A_{t,i}) \epsilon_{t,i} )^2 }{ (n-2) N_{t,0} } \Pto 0$.
We will only show that $\frac{ ( \sum_{i=1}^n A_{t,i} \epsilon_{t,i} )^2 }{ (n-2) N_{t,1} } \Pto 0$; $\frac{ ( \sum_{i=1}^n (1-A_{t,i}) \epsilon_{t,i} )^2 }{ (n-2) N_{t,0} } \Pto 0$ holds by a very similar argument.

Note that by Lemma \ref{lemma:ratio}, $\frac{N_{t,1}}{n \pi_t^{(n)} } \Pto 1$.
Thus, to show that $\frac{ ( \sum_{i=1}^n A_{t,i} \epsilon_{t,i} )^2 }{ (n-2) N_{t,1} } \Pto 0$ by Slutsky's Theorem it is sufficient to show that $\frac{ ( \sum_{i=1}^n A_{t,i} \epsilon_{t,i} )^2 }{ (n-2) n \pi_t^{(n)} } \Pto 0$.
Let $\delta > 0$. By Markov inequality,
\begin{equation*}
		\PP \bigg( \bigg| \frac{ ( \sum_{i=1}^n A_{t,i} \epsilon_{t,i} )^2 }{ (n-2) n \pi_t^{(n)} } \bigg| > \delta \bigg) 
		\leq \E \bigg[ \frac{1}{ \delta (n-2) n \pi_t^{(n)} } \bigg( \sum_{i=1}^n A_{t,i} \epsilon_{t,i} \bigg)^2 \bigg]
		= \E \bigg[ \frac{1}{ \delta (n-2) n \pi_t^{(n)} } \sum_{j=1}^n \sum_{i=1}^n A_{t,j} A_{t,i} \epsilon_{t,i} \epsilon_{t,j} \bigg]
\end{equation*}
Since $\pi_t^{(n)} \in \HH_{t-1}^{(n)}$,
\begin{equation*}
	= \E \bigg[ \frac{1}{ \delta (n-2) n \pi_t^{(n)} } \sum_{j=1}^n \sum_{i=1}^n \E[ A_{t,j} A_{t,i} \epsilon_{t,i} \epsilon_{t,j} | \HH_{t-1}^{(n)} ]\bigg]
\end{equation*}
Since for $i \not= j$, $\E[ A_{t,j} A_{t,i} \epsilon_{t,j} \epsilon_{t,i} | \HH_{t-1}^{(n)} ] = \E[ A_{t,j} \epsilon_{t,j} | \HH_{t-1}^{(n)} ] \E[ A_{t,i} \epsilon_{t,i} | \HH_{t-1}^{(n)} ] = 0$,
\begin{equation*}
	= \E \bigg[ \frac{1}{ \delta (n-2) n \pi_t^{(n)} } \sum_{i=1}^n \E[ A_{t,i} \epsilon_{t,i}^2 | \HH_{t-1}^{(n)} ]\bigg]
\end{equation*}
Since $\E[ A_{t,i} \epsilon_{t,i}^2 | \HH_{t-1}^{(n)} ] = \E[ \epsilon_{t,i}^2 | \HH_{t-1}^{(n)}, A_{t,i}=1 ] \pi_t^{(n)} = \sigma^2 \pi_t^{(n)}$,
\begin{equation*}
	= \E \bigg[ \frac{1}{ \delta (n-2) n \pi_t^{(n)} } n \sigma^2 \pi_t^{(n)} \bigg]
	= \frac{ \sigma^2 }{ \delta (n-2) } \to 0 ~~~ \qed
\end{equation*}

%% file: appendix/BOLScontext.tex
%%%%%%%%%%%%%%%%%%%%%%%%%%%%%%%%%%%%%%%%%%
%%%%%%%%%%%%%%%%%%%%%%%%%%%%%%%%%%%%%%%%%%

%%%%%%%%%%%%%%%%%%%%%%%%%%%%%%%%%%%%%%%%%%
%%%%%%%%%%%%%%%%%%%%%%%%%%%%%%%%%%%%%%%%%%

\section{Asymptotic Normality of the Batched OLS Estimator: Contextual Bandits}
\label{appendix:BOLScontext}

\paragraph{Theorem \ref{thm:bolscontext} (Asymptotic Normality of the Batched OLS Statistic)}
\begin{itshape}
For a $K$-armed contextual bandit, we for each $t \in [1 \colon T]$, we have the BOLS estimator:
\begin{equation*}
\bs{\betahat}_t^{\BOLS} = \begin{bmatrix} 
\underline{\boldC}_{t,0} & \bs{0} & \bs{0} & \hdots & \bs{0} \\
\bs{0} & \underline{\boldC}_{t,1} & \bs{0} & \hdots & \bs{0} \\
\bs{0} & \bs{0} & \underline{\boldC}_{t,2} & \hdots & \bs{0} \\
\vdots & \vdots & \vdots & \ddots & \vdots \\
\bs{0} & \bs{0} & \bs{0} & \hdots & \underline{\boldC}_{t,K-1} \\
\end{bmatrix}^{-1}
\sum_{i=1}^n \begin{bmatrix}
	\II_{ A_{t,i} = 0 } \boldC_{t,i} \\
	\II_{ A_{t,i} = 1 } \boldC_{t,i} \\
	\vdots \\
	\II_{ A_{t,i} = K-1 } \boldC_{t,i} \\
\end{bmatrix} R_{t,i}
\in \real^{Kd}
\end{equation*}
where $\underline{\boldC}_{t,k} := \sum_{i=1}^n \II_{ A_{t,i}^{(n)} = k } \boldC_{t,i} ( \boldC_{t,i} )^\top \in \real^{d \by d}$.
Assuming Conditions \ref{cond:weakmoments} (weak moments), \ref{cond:condiid} (conditionally i.i.d. actions), \ref{cond:iidcontext} (conditionally i.i.d. contexts), and \ref{cond:boundedcontext} (bounded contexts), and a conditional clipping rate $f(n) = c$ for some $0 \leq c < \frac{1}{2}$ (see Definition \ref{def:condclipping}), we show that as $n \to \infty$, 
\begin{equation*}
\begin{bmatrix} 
\TN{Diagonal} \big[ \underline{\boldC}_{1,0}, \underline{\boldC}_{1,1}, ..., \underline{\boldC}_{1,K-1} \big]^{1/2} ( \bs{\betahat}_1^{\BOLS} - \bs{\beta}_1 ) \\
\TN{Diagonal} \big[ \underline{\boldC}_{2,0}, \underline{\boldC}_{2,1}, ..., \underline{\boldC}_{2,K-1} \big]^{1/2} ( \bs{\betahat}_2^{\BOLS} - \bs{\beta}_2 )  \\
\vdots \\
\TN{Diagonal} \big[ \underline{\boldC}_{T,0}, \underline{\boldC}_{T,1}, ..., \underline{\boldC}_{T,K-1} \big]^{1/2} ( \bs{\betahat}_T^{\BOLS} - \bs{\beta}_T )  \\
 \end{bmatrix}
\Dto \N(0, \sigma^2 \under{\bo{I}}_{TKd} )
\end{equation*}
\end{itshape}

\begin{lemma}
\label{lemma:contextratio}
Assuming the conditions of Theorem \ref{thm:bolscontext}, for any batch $t \in [1 \colon T]$ and any arm $k \in [0 \colon K-1]$, as $n \to \infty$, 
\begin{equation}
	\label{eqn:lemma3a}
		\bigg[ \sum_{i=1}^n \II_{ A_{t,i} = k } \boldC_{t,i} \boldC_{t,i}^\top \bigg] \big[ n \underline{\boldZ}_{t,k} P_{t,k} \big]^{-1} \Pto \under{\bo{I}}_d
\end{equation}
\begin{equation}
		\label{eqn:lemma3b}
		\bigg[ \sum_{i=1}^n \II_{ A_{t,i} = k } \boldC_{t,i} \boldC_{t,i}^\top \bigg]^{1/2} \big[ n \underline{\boldZ}_{t,k} P_{t,k} \big]^{-1/2} \Pto \under{\bo{I}}_d
\end{equation}
%where $\boldC_{t,i} i.i.d.$ and $A_t \sim \TN{Categorical} \big( \bs{\pi}_{t,i}^{(n)} \big) - 1$; 
where $P_{t,k} := \PP( A_{t,i} = k | H_{t-1}^{(n)} )$ and $\underline{\boldZ}_{t,k} := \E \big[ \boldC_{t,i} \boldC_{t,i}^\top \big| H_{t-1}^{(n)}, A_{t,i} = k \big]$. 
\end{lemma}

\paragraph{Proof of Lemma \ref{lemma:contextratio}:}
We first show that as $n \to \infty$, 
$\frac{1}{n} \sum_{i=1}^n \big( \II_{ A_{t,i} = k } \boldC_{t,i} \boldC_{t,i}^\top - \underline{\boldZ}_{t,k} P_{t,k} \big) \Pto \underline{ \bs{0} }$.
It is sufficient to show that convergence holds entry-wise so for any $r, s \in [0 \colon d-1]$,  as $n \to \infty$, 
$\frac{1}{n} \sum_{i=1}^n \big( \II_{ A_{t,i} = k } \boldC_{t,i} \boldC_{t,i}^\top (r, s) - P_{t,k} \underline{\boldZ}_{t,k} (r,s) \big) \Pto 0$.
%%%%%%%%%%%%%%%%%%%%%%%%%%%%%%%
Note that 
\small
\begin{equation*}
\E \bigg[ \II_{ A_{t,i} = k } \boldC_{t,i} \boldC_{t,i}^\top (r,s) - P_{t,k} \underline{\boldZ}_{t,k} (r,s)\bigg]
= \E \bigg[ \E \big[ \boldC_{t,i} \boldC_{t,i}^\top (r,s) \big| H_{t-1}, A_{t,i} = k \big] P_{t,k} - P_{t,k} \underline{\boldZ}_{t,k} (r,s) \bigg] = 0
\end{equation*}
\normalsize
%%%%%%%%%%%%%%%%%%%%%%%%%%%%%%%
By Chebychev inequality, for any $\epsilon > 0$,
\small
\begin{equation*}
\PP \bigg( \bigg| \frac{1}{ n } \sum_{i=1}^n \II_{ A_{t,i} = k } \boldC_{t,i} \boldC_{t,i}^\top (r,s) - P_{t,k} \underline{\boldZ}_{t,k} (r,s) \bigg| > \epsilon \bigg)
\leq \frac{1}{\epsilon^2 n^2} \E \bigg[ \bigg( \sum_{i=1}^{n} \II_{ A_{t,i} = k } \boldC_{t,i} \boldC_{t,i}^\top - P_{t,k} \underline{\boldZ}_{t,k} (r,s) \bigg)^2 \bigg]
\end{equation*}
\begin{equation}
\label{eqn:one}
= \frac{1}{\epsilon^2 n^2 } \sum_{i=1}^{n} \sum_{j=1}^{n} \E \bigg[ \big[ \II_{ A_{t,i} = k } \boldC_{t,i} \boldC_{t,i}^\top (r,s) - P_{t,k} \underline{\boldZ}_{t,k} (r,s) \big] 
\big[ \II_{ A_{t,i} = k } \boldC_{t,j} \boldC_{t,j}^\top (r,s) - P_{t,k} \underline{\boldZ}_{t,k}(r,s) \big] \bigg]
\end{equation}
\normalsize
%%%%%%%%%%%%%%%%%%%%%%%%%%%%%%%
By conditional independence and by law of iterated expectations (conditioning on $H_{t-1}^{(n)}$), for $i \neq j$, 
$\E \big[ \big( \II_{ A_{t,i} = k } \boldC_{t,i} \boldC_{t,i}^\top (r,s) - P_{t,k} \underline{\boldZ}_{t,k} (r,s) \big) 
\big( \II_{ A_{t,j} = k } \boldC_{t,j} \boldC_{t,j}^\top (r,s) - P_{t,k} \underline{\boldZ}_{t,k} (r,s) \big) \big] = 0$.
Thus, \eqref{eqn:one} above equals the following:
%Thus by the law of iterated expectations (conditioning on $H^{(n)}$) we get that \eqref{eqn:one} above equals the following:
\begin{equation*}
= \frac{1}{\epsilon^2 n^2 } \sum_{i=1}^{n} \E \bigg[ \bigg( \II_{ A_{t,i} = k } \boldC_{t,i} \boldC_{t,i}^\top (r,s) - P_{t,k} \underline{\boldZ}_{t,k} (r,s)  \bigg)^2 \bigg]
\end{equation*}
\begin{equation*}
= \frac{1}{\epsilon^2 n^2 } \sum_{i=1}^{n} \E \bigg[ \II_{ A_{t,i} = k } \big[ \boldC_{t,i} \boldC_{t,i}^\top (r,s) \big]^2 
- 2 \II_{ A_{t,i} = k } \boldC_{t,i} \boldC_{t,i}^\top (r,s) P_{t,k} \underline{\boldZ}_{t,k} (r,s) 
+ P_{t,k}^2 \big[ \underline{\boldZ}_{t,k} (r,s) \big]^2 \bigg]
\end{equation*}
\begin{equation*}
= \frac{1}{\epsilon^2 n^2 } \sum_{i=1}^{n} \E \bigg[ \II_{ A_{t,i} = k } \big[ \boldC_{t,i} \boldC_{t,i}^\top (r,s) \big]^2 - P_{t,k}^2 \big[ \underline{\boldZ}_{t,k} (r,s) \big]^2 \bigg]
\end{equation*}
\begin{equation*}
= \frac{1}{\epsilon^2 n } \E \bigg[ \II_{ A_{t,i} = k } \big[ \boldC_{t,i} \boldC_{t,i}^\top (r,s) \big]^2 - P_{t,k}^2 \big[ \underline{\boldZ}_{t,k} (r,s) \big]^2 \bigg] 
\leq \frac{2 d \max(u^2, 1) }{\epsilon^2 n } \to 0
\end{equation*}
as $n \to \infty$. The last inequality above holds by Condition \ref{cond:boundedcontext}.

%%%%%%%%%%%%%%%%%%%%%%%%%%%%%%%
\bo{Proving Equation \eqref{eqn:lemma3a}: } 
%
%It is sufficient to show that for some $K > 0$, $\PP \big( \| n \boldZ_{t,k}^{-1} P_{t,k} \|_{op} \geq K \big) \to 0$. 
%
%
%
%\todo[inline]{working1}
%
%Recall that $P_{t,k} = \PP( A_{t,i} = k ~|~ H_{t-1}^{(n)} )$, so $P_{t,k} \geq f(n)$ by our conditional clipping assumption.
%
%By our conditional clipping condition and Bayes rule we have that for all $\boldc \in [-u,u]^d$, 
%\begin{equation*}
%\PP( \boldC_{t,i} = \boldc | A_{t,i} = k, H_{t-1}^{(n)} ) = \frac{ \PP( A_{t,i} = k | \boldC_{t,i} = \boldc, H_{t-1}^{(n)} ) \PP( \boldC_{t,i} = \boldc | H_{t-1}^{(n)} ) }{ \PP( A_{t,i} = k | H_{t-1}^{(n)} ) } 
%\geq \frac{ f(n) ~ \PP( \boldC_{t,i} = \boldc | H_{t-1}^{(n)} ) }{ 1 }.
%\end{equation*}
%Thus, we have that 
%\begin{equation*}
%\underline{\boldZ}_{t,k} = \E \big[ \boldC_{t,i} \boldC_{t,i}^\top \big| H_{t-1}^{(n)}, A_{t,i} = k \big] 
%\succcurlyeq f(n) \E \big[ \boldC_{t,i} \boldC_{t,i}^\top \big| H_{t-1}^{(n)} \big] 
%= f(n) \underline{ \bs{\Sigma} }_t^{(n)},
%\end{equation*}
%which implies
%\begin{equation}
%\label{eqn:eigenvaluebound}
%\lambda_{\max} ( \underline{\boldZ}_{t,k} ^{-1} ) \leq \frac{1}{ f(n) } \lambda_{\max} \bigg( \big( \underline{ \bs{\Sigma} }_t^{(n)} \big)^{-1} \bigg) \leq \frac{1}{ l ~ f(n) }
%\end{equation}
%where the last inequality above holds for constant $l$ by Condition \ref{cond:boundedcontext}.
%
%
%\todo[inline]{working2}

It is sufficient to show that 
\begin{equation}
\label{eqn:lemma1}
\bigg\| \frac{2 \max(d u^2, 1) }{\epsilon^2 n } \big[ n \underline{\boldZ}_{t,k} P_{t,k} \big]^{-1} \bigg\|_{op}
= \bigg\| \frac{2 \max(d u^2, 1) }{\epsilon^2 n^2 P_{t,k}  } \underline{\boldZ}_{t,k}^{-1} \bigg\|_{op} \Pto 0
\end{equation}

We define random variable $M_t^{(n)} = \II_{( \forall ~\boldc \in \real^d, ~ \MC{A}_t( H_{t-1}^{(n)}, \boldc) \in [ f(n), 1-f(n) ]^K )}$, representing whether the conditional clipping condition is satisfied.
Note that by our conditional clipping assumption, $M_t^{(n)} \Pto 1$ as $n \to \infty$.
The left hand side of \eqref{eqn:lemma1} is equal to the following
\begin{equation}
\label{eqn:lemma2}
\bigg\| \frac{2 \max(d u^2, 1) }{\epsilon^2 n^2 P_{t,k}  } \underline{\boldZ}_{t,k}^{-1} ( M_t^{(n)} + (1-M_t^{(n)}) ) \bigg\|_{op}
= \bigg\| \frac{2 \max(d u^2, 1) }{\epsilon^2 n^2 P_{t,k}  } \underline{\boldZ}_{t,k}^{-1} M_t^{(n)} \bigg\|_{op} + o_p(1)
\end{equation}

By our conditional clipping condition and Bayes rule we have that for all $\boldc \in [-u,u]^d$, 
\begin{equation*}
\PP( \boldC_{t,i} = \boldc | A_{t,i} = k, H_{t-1}^{(n)}, M_t^{(n)}=1 )
\end{equation*}
\begin{equation*}
	= \frac{ \PP( A_{t,i} = k | \boldC_{t,i} = \boldc, H_{t-1}^{(n)}, M_t^{(n)}=1 ) \PP( \boldC_{t,i} = \boldc | H_{t-1}^{(n)}, M_t^{(n)}=1 ) }{ \PP( A_{t,i} = k | H_{t-1}^{(n)}, M_t^{(n)}=1 ) } 
\end{equation*}
\begin{equation*}
\geq \frac{ f(n) ~ \PP( \boldC_{t,i} = \boldc | H_{t-1}^{(n)}, M_t^{(n)}=1 ) }{ 1 }.
\end{equation*}
Thus, we have that 
\begin{equation*}
\underline{\boldZ}_{t,k} M_t^{(n)} = \E \big[ \boldC_{t,i} \boldC_{t,i}^\top \big| H_{t-1}^{(n)}, A_{t,i} = k \big] M_t^{(n)}
= \E \big[ \boldC_{t,i} \boldC_{t,i}^\top \big| H_{t-1}^{(n)}, A_{t,i} = k, M_t^{(n)} = 1 \big] M_t^{(n)}
\end{equation*}
\begin{equation*}
\succcurlyeq f(n) \E \big[ \boldC_{t,i} \boldC_{t,i}^\top \big| H_{t-1}^{(n)}, M_t^{(n)} = 1 \big] M_t^{(n)}
= f(n) \E \big[ \boldC_{t,i} \boldC_{t,i}^\top \big| H_{t-1}^{(n)} \big] M_t^{(n)}
= f(n) \underline{ \bs{\Sigma} }_t^{(n)} M_t^{(n)}.
\end{equation*}
By apply matrix inverses to both sides of the above inequality, we get that
%\begin{equation}
%
%\PP \bigg( \lambda_{\max} ( \underline{\boldZ}_{t,k} ^{-1} M_t^{(n)} ) \leq \frac{1}{ f(n) } \lambda_{\max} \bigg( \big( \underline{ \bs{\Sigma} }_t^{(n)} \big)^{-1} \bigg) M_t^{(n)} \bigg) \to 1 \leq \frac{1}{ l ~ f(n) }
%\end{equation}
\begin{equation}
\label{eqn:eigenvaluebound}
\lambda_{\max} ( \underline{\boldZ}_{t,k} ^{-1} M_t^{(n)} ) \leq \frac{1}{ f(n) } \lambda_{\max} \bigg( \big( \underline{ \bs{\Sigma} }_t^{(n)} \big)^{-1} \bigg) M_t^{(n)} \leq \frac{1}{ l ~ f(n) }
\end{equation}
where the last inequality above holds for constant $l$ by Condition \ref{cond:boundedcontext}.
Recall that $P_{t,k} = \PP( A_{t,i} = k ~|~ H_{t-1}^{(n)} )$, so $P_{t,k} ~|~ (M_t^{(n)} = 1) \geq f(n)$.
Thus, equation \eqref{eqn:lemma2} is bounded above by the following
\begin{equation*}
\leq \frac{2 \max(d u^2, 1) }{\epsilon^2 n^2 l f(n)^2  } + o_p(1) \Pto 0 
\end{equation*}
where the limit above holds because we assume that $f(n) = c$ for some $0 < c \leq \frac{1}{2}$ $\qed$.
%where the limit above holds because we assume that for $0 < \alpha < \frac{1}{2}$, thus $n f(n)^2 = n \frac{1}{n^{2\alpha}} \to \infty$ $\qed$.

\bo{Proving Equation \eqref{eqn:lemma3b}: }
By Condition \ref{cond:boundedcontext}, $\| \frac{1}{n} \under{\boldC}_{t,k} \|_{\max} \leq u$ and $\| \under{\boldZ}_{t,k} P_{t,k} \|_{\max} \leq u$.
Thus, any continuous function of $\frac{1}{n} \under{\boldC}_{t,k}$ and $\under{\boldZ}_{t,k} P_{t,k}$ will have compact support and thus be uniformly continuous.
For any uniformly continuous function $f: \real^{d \by d} \to \real^{d \by d}$, for any $\epsilon > 0$, there exists a $\delta > 0$ such that for any matrices $\underline{\boldA}, \underline{\boldB} \in \real^{d \by d}$, whenever $\| \underline{\boldA} - \underline{\boldB} \|_{\TN{op}} < \delta$, then $\| f( \underline{\boldA} ) - f( \underline{\boldB} ) \|_{\TN{op}} < \epsilon$.
Thus, for any $\epsilon > 0$, there exists some $\delta > 0$ such that 
\begin{equation*}
\PP \bigg( \bigg\| \bigg( \frac{1}{n} \sum_{i=1}^n \II_{(A_{t,k} = k)} \boldC_{t,i} \boldC_{t,i}^\top \bigg) - \under{\boldZ}_{t,k} P_{t,k} \bigg\|_{\TN{op}} > \delta \bigg) \to 0
\end{equation*}
implies
\begin{equation*}
\PP \bigg( \bigg\| f \bigg( \frac{1}{n} \sum_{i=1}^n \II_{(A_{t,k} = k)} \boldC_{t,i} \boldC_{t,i}^\top \bigg) - f( \under{\boldZ}_{t,k} P_{t,k} ) \bigg\|_{\TN{op}} > \epsilon \bigg) \to 0
\end{equation*}
Thus, by letting $f$ be the matrix square-root function,
\begin{equation*}
\bigg( \frac{1}{n} \sum_{i=1}^n \II_{(A_{t,k} = k)} \boldC_{t,i} \boldC_{t,i}^\top \bigg)^{1/2} - ( \under{\boldZ}_{t,k} P_{t,k} )^{1/2} \Pto \under{\bs{0}}.
\end{equation*}

We now want to show that for some constant $r > 0$, $\PP \big( \big\| \under{\boldZ}_{t,k}^{-1} \frac{1}{ P_{t,k} } \big\|_{\TN{op}} > r \big)$, because this would imply that
\begin{equation*}
\bigg[ \bigg( \frac{1}{n} \sum_{i=1}^n \II_{(A_{t,k} = k)} \boldC_{t,i} \boldC_{t,i}^\top \bigg)^{1/2} - ( \under{\boldZ}_{t,k} P_{t,k} )^{1/2} \bigg] ( \under{\boldZ}_{t,k} P_{t,k} )^{-1/2}  \Pto \under{\bs{0}}.
\end{equation*}

Recall that for $M_t^{(n)} = \II_{( \forall ~\boldc \in \real^d, ~ \MC{A}_t( H_{t-1}^{(n)}, \boldc) \in [ f(n), 1-f(n) ]^K )}$, representing whether the conditional clipping condition is satisfied,
\begin{equation*}
\under{\boldZ}_{t,k}^{-1} = \under{\boldZ}_{t,k}^{-1} ( M_t^{(n)} + (1- M_t^{(n)})) = \under{\boldZ}_{t,k}^{-1} M_t^{(n)} + o_p(1).
\end{equation*}
Thus it is sufficient to show that $\PP \big( \big\| \under{\boldZ}_{t,k}^{-1} \frac{1}{ P_{t,k} }M_t^{(n)} \big\|_{\TN{op}} > r \big)$.
Recall that by equation \eqref{eqn:eigenvaluebound} we have that
\begin{equation*}
\lambda_{\max} ( \underline{\boldZ}_{t,k}^{-1} M_t^{(n)} ) \leq \frac{1}{ f(n) } \lambda_{\max} \bigg( \big( \underline{ \bs{\Sigma} }_t^{(n)} \big)^{-1} \bigg) M_t^{(n)} \leq \frac{1}{ l ~ f(n) }
\end{equation*}
Also note that $P_{t,k} = \PP( A_{t,i} = k ~|~ H_{t-1}^{(n)} )$, so $P_{t,k} ~|~ (M_t^{(n)} = 1) \geq f(n)$.
Thus we have that
\begin{equation*}
\PP \bigg( \bigg\| \under{\boldZ}_{t,k}^{-1} \frac{1}{ P_{t,k} }M_t^{(n)} \bigg\|_{\TN{op}} > r \bigg) \leq \II_{ ( \frac{1}{ l ~ f(n)^2 } > r )} = 0
\end{equation*}
for $r > \frac{1}{ l ~ f(n)^2 } = \frac{1}{ l c^2 }$, since we assume that $f(n) = c$ for some $0 < c \leq \frac{1}{2}$. $\qed$

%It can be shown that the square root function on the cone of positive definite matrices is uniformly continuous. 
%For any uniformly continuous function $f: \real^{d \by d} \to \real^{d \by d}$, for any $\epsilon > 0$, there exists a $\delta > 0$ such that for any matrices $\underline{\boldA}, \underline{\boldB} \in \real^{d \by d}$, whenever
%$\| \underline{\boldA} - \underline{\boldB} \|_{op}< \delta$, then $\| f( \underline{\boldA} ) - f( \underline{\boldB} ) \|_{op} < \epsilon$.
%
%Since 
%\begin{equation*}
%\PP \bigg( \bigg\| \bigg[ \sum_{i=1}^n \II_{ A_{t,i} = k } \boldC_{t,i} \boldC_{t,i}^\top \bigg] \big[ n \underline{\boldZ}_{t,k} P_{t,k} \big]^{-1}  - \under{\bs{I}}_d \bigg\|_{op} > \delta \bigg) \to 0
%\end{equation*}
%thus
%\begin{equation*}
%\PP \bigg( \bigg\| f \bigg( \bigg[ \sum_{i=1}^n \II_{ A_{t,i} = k } \boldC_{t,i} \boldC_{t,i}^\top \bigg] \big[ n \underline{\boldZ}_{t,k} P_{t,k} \big]^{-1} \bigg) - f \big( \under{\bs{I}}_d \big) \bigg\|_{op} > \epsilon \bigg) \to 0
%\end{equation*}
%The final result holds by letting $f$ be the matrix square root function. $\qed$

\paragraph{Proof of Theorem \ref{thm:bolscontext}:}
We define $P_{t,k} := \PP( A_{t,i} = k | H_{t-1}^{(n)} )$ and $\underline{\boldZ}_{t,k} := \E \big[ \boldC_{t,i} \boldC_{t,i}^\top \big| H_{t-1}^{(n)}, A_{t,i} = k \big]$.
% for $\boldC_{t,i} \sim [ \bs{\mu}_t^{(n)}, \underline{ \bs{ \Sigma } }_t^{(n)} ]$.
We also define
\begin{equation*}
\boldD_t^{(n)} := \TN{Diagonal} \big[ \underline{\boldC}_{t,0}, \underline{\boldC}_{t,1}, ..., \underline{\boldC}_{t,K-1} \big]^{1/2} ( \bs{\betahat}_t - \bs{\beta}_t )
= \sum_{i=1}^n \begin{bmatrix}
	\underline{\boldC}_{t,0}^{-1/2} ~ \boldC_{t,i} \II_{ A_{t,i} = 0 }  \\
	\underline{\boldC}_{t,1}^{-1/2} ~ \boldC_{t,i} \II_{ A_{t,i} = 1 } \\
	\vdots \\
	\underline{\boldC}_{t,K-1}^{-1/2} ~ \boldC_{t,i} \II_{ A_{t,i} = K-1 } \\
\end{bmatrix} \epsilon_{t,i}
\end{equation*}
%%%%%%%%%%%%%%%%%%%%%%%%%%%%%%%
We want to show that
$[ \boldD_1^{(n)}, \boldD_2^{(n)}, ..., \boldD_T^{(n)} ]^\top \Dto \N( \bs{0} , \sigma^2 \underline{\bo{I}}_{TKd} )$. 
%%%%%%%%%%%%%%%%%%%%%%%%%%%%%%%
By Lemma \ref{lemma:contextratio} and Slutsky's Theorem, it sufficient to show that as $n \to \infty$,
$[ \boldQ_1^{(n)}, \boldQ_2^{(n)}, ..., \boldQ_T^{(n)} ]^\top \Dto \N( \bs{0} , \sigma^2 \underline{\bo{I}}_{TKd} )$ for
\begin{equation*}
\boldQ_t^{(n)} := \sum_{i=1}^n \begin{bmatrix}
	\frac{1}{ \sqrt{ n P_{t,0} } } \underline{\boldZ}_{t,0}^{-1/2} \boldC_{t,i} \II_{ A_{t,i} = 0 }  \\
	\frac{1}{ \sqrt{ n P_{t,1} } } \underline{\boldZ}_{t,1}^{-1/2} \boldC_{t,i} \II_{ A_{t,i} = 1 }  \\
	\vdots \\
	\frac{1}{ \sqrt{ n P_{t,K-1} } }  \underline{\boldZ}_{t,K-1}^{-1/2} \boldC_{t,i} \II_{ A_{t,i} = K-1 } \\
\end{bmatrix} \epsilon_{t,i}
\end{equation*}
%%%%%%%%%%%%%%%%%%%%%%%%%%%%%%%
By Cramer Wold device, it is sufficient to show that for any $\boldb \in \real^{TKd}$ with $\| \boldb \|_2 = 1$, where $\boldb = [ \boldb_1, \boldb_2, ..., \boldb_T]$ for $\boldb_t \in \real^{Kd}$,
%$\sum_{t=1}^T \boldb_t^\top \boldQ_t^{(n)} \Dto \N(0, \sigma^2)$ 
as $n \to \infty$.
\begin{equation}
\label{eqn:qnormality}
\sum_{t=1}^T 
\boldb_t^\top \boldQ_t^{(n)} 
\Dto \N(0, \sigma^2)
\end{equation}
%%%%%%%%%%%%%%%%%%%%%%%%%%%%%%%
We can further define for all $t \in [1 \colon T]$, $\boldb_t = [\boldb_{t,0}, \boldb_{t,1}, ..., \boldb_{t,K-1}]$ with $\boldb_{t,k} \in \real^d$.
Thus to show \eqref{eqn:qnormality} it is equivalent to show that
\begin{equation*}
\sum_{t=1}^T \sum_{k=0}^{K-1} \boldb_{t,k}^\top \frac{1}{ \sqrt{ n P_{t,k} } }  \underline{\boldZ}_{t,k}^{-1/2} 
\sum_{i=1}^n \II_{ A_{t,i} = k } \boldC_{t,i} \epsilon_{t,i} \Dto \N( 0 , \sigma^2 )
\end{equation*}
%%%%%%%%%%%%%%%%%%%%%%%%%%%%%%%
We define $Y_{t,i}^{(n)} := \sum_{k=0}^{K-1} \boldb_{t,k}^\top \frac{ 1 }{ \sqrt{ n P_{t,k} } } \II_{ A_{t,i} = k } \underline{\boldZ}_{t,k}^{-1/2} \boldC_{t,i} \epsilon_{t,i} $.
The sequence $Y_{1,1}^{(n)}, Y_{1,2}^{(n)}, ..., Y_{1,n}^{(n)}, ... Y_{T,1}^{(n)}, Y_{T,2}^{(n)}, ..., Y_{T,n}^{(n)}$ %Y_{2,1}^{(n)}, Y_{2,2}^{(n)}, ..., Y_{2,n}^{(n)}, 
is a martingale difference array with respect to the sequence of histories $\{ H_{t-1}^{(n)} \}_{t=1}^T$ because
$\E[ Y_{t,i}^{(n)} | H_{t-1}^{(n)}]
= \E \bigg[ \E[ Y_{t,i}^{(n)} | H_{t-1}^{(n)}, A_{t,i}, \boldC_{t,i} ] \bigg| H_{t-1}^{(n)} \bigg] = 0$
for all $i \in [1 \colon n]$ and all $t \in [1 \colon T]$. 
We then apply the martingale central limit theorem of \cite{dvoretzky1972asymptotic} to $Y_{t,i}^{(n)}$ to show the desired result (see the proof of Theorem \ref{thm:triangularCLT} in Appendix \ref{appendix:triangularCLT} for the statement of the martingale CLT conditions).
Note that the first condition (a) of the martingale CLT is already satisfied, as we just showed that $Y_{t,i}^{(n)}$ form a martingale difference array with respect to $H_{t-1}^{(n)}$.

\paragraph{Condition(b): Conditional Variance}
\begin{equation*}
\sum_{t=1}^T \sum_{i=1}^{n} \E[ Y_{t,i}^2 | H_{t-1}^{(n)}]
= \sum_{t=1}^T \sum_{i=1}^{n}  \E \bigg[ \bigg( \sum_{k=0}^{K-1} \boldb_{t,k}^\top \frac{1}{ \sqrt{ n P_{t,k} } } \underline{\boldZ}_{t,k}^{-1/2}
\II_{ A_{t,i} = k } \boldC_{t,i} \epsilon_{t,i} \bigg)^2 \bigg| H_{t-1}^{(n)} \bigg]
\end{equation*}
\begin{equation*}
= \sum_{t=1}^T \sum_{i=1}^{n} \sum_{k=0}^{K-1} \frac{1}{n P_{t,k} } \boldb_{t,k}^\top \under{\boldZ}_{t,k}^{-1/2} \E \bigg[ 
 \II_{ A_{t,i} = k } \boldC_{t,i} \boldC_{t,i}^\top \epsilon_{t,i}^2 \bigg| H_{t-1}^{(n)} \bigg] \under{\boldZ}_{t,k}^{-1/2} \boldb_{t,k}
 \end{equation*}
 %%%%%%%%%%%%%%%%%%%%%%%%%%%%%%%
 By law of iterated expectations (conditioning on $H_{t-1}^{(n)}, A_{t,i}, \boldC_{t,i}$) and Condition \ref{cond:weakmoments},
\begin{equation*}
= \frac{1}{n}  \sum_{t=1}^T \sum_{i=1}^{n} \sum_{k=0}^{K-1} \frac{1}{ P_{t,k} } \boldb_{t,k}^\top \under{\boldZ}_{t,k}^{-1/2} \E \bigg[ 
 \II_{ A_{t,i} = k } \boldC_{t,i} \boldC_{t,i}^\top \bigg| H_{t-1}^{(n)} \bigg] \under{\boldZ}_{t,k}^{-1/2} \boldb_{t,k} \sigma^2
 \end{equation*}
\begin{equation*}
= \frac{1}{n}  \sum_{t=1}^T \sum_{i=1}^{n} \sum_{k=0}^{K-1} \frac{1}{ P_{t,k} } \boldb_{t,k}^\top \under{\boldZ}_{t,k}^{-1/2} 
 \E \bigg[ \boldC_{t,i} \boldC_{t,i}^\top \bigg| H_{t-1}^{(n)}, A_{t,i} = k \bigg] P_{t,k} \under{\boldZ}_{t,k}^{-1/2} \boldb_{t,k} \sigma^2
 \end{equation*}
\begin{equation*}
= \frac{1}{n} \sum_{t=1}^T \sum_{i=1}^{n} \sum_{k=0}^{K-1} \boldb_{t,k}^\top \under{\bo{I}}_d \boldb_{t,k} \sigma^2
= \sigma^2 \sum_{t=1}^T \sum_{k=0}^{K-1} \boldb_{t,k}^\top \boldb_{t,k} 
= \sigma^2
\end{equation*}

\paragraph{Condition(c): Lindeberg Condition}
Let $\delta > 0$.
\small
\begin{equation*}
\sum_{t=1}^T \sum_{i=1}^{n} \E \big[ Y_{t,i}^2  \II_{( |Y_{t,i}| > \delta )} \big| H_{t-1}^{(n)} \big]
= \sum_{t=1}^T \sum_{i=1}^{n}  \E \bigg[ \bigg( \sum_{k=0}^{K-1} \boldb_{t,k}^\top \frac{1}{ \sqrt{ n P_{t,k} } } \boldZ_{t,i}^{-1/2}
\II_{ A_{t,i} = k } \boldC_{t,i} \epsilon_{t,i} \bigg)^2 \II_{( Y_{t,i}^2 > \delta^2 )} \bigg| H_{t-1}^{(n)} \bigg]
\end{equation*}
\normalsize
\begin{equation*}
= \sum_{t=1}^T \sum_{i=1}^{n} \sum_{k=0}^{K-1} \frac{1}{n P_{t,k} } \boldb_{t,k}^\top \boldZ_{t,i}^{-1/2} \E \bigg[ 
\II_{ A_{t,i} = k } \boldC_{t,i} \boldC_{t,i}^\top \epsilon_{t,i}^2 \II_{( Y_{t,i}^2 > \delta^2 )} \bigg| H_{t-1}^{(n)} \bigg] \boldZ_{t,i}^{-1/2} \boldb_{t,k}
\end{equation*}
%%%%%%%%%%%%%%%%%%%%%%%%%%%%%%%
It is sufficient to show that for any $t \in [1 \colon T]$ and any $k \in [0:K-1]$ the following converges in probability to zero:
\begin{equation*}
\sum_{i=1}^{n} \frac{1}{ n P_{t,k} } \boldb_{t,k}^\top \boldZ_{t,i}^{-1/2} \E \bigg[ 
 \II_{ A_{t,i} = k } \boldC_{t,i} \boldC_{t,i}^\top \epsilon_{t,i}^2 \II_{( Y_{t,i}^2 > \delta^2 )} \bigg| H_{t-1}^{(n)} \bigg] \boldZ_{t,i}^{-1/2} \boldb_{t,k}
 \end{equation*}
 Recall that $Y_{t,i} = \sum_{k=0}^{K-1} \boldb_{t,k}^\top \frac{ 1 }{ \sqrt{ n P_{t,k} } } \II_{ A_{t,i} = k } \underline{\boldZ}_{t,k}^{-1/2} \boldC_{t,i} \epsilon_{t,i} $. 
\begin{equation*}
= \frac{1}{ n } \sum_{i=1}^{n} \boldb_{t,k}^\top \boldZ_{t,i}^{-1/2} \E \bigg[ 
\boldC_{t,i} \boldC_{t,i}^\top \epsilon_{t,i}^2 \II_{( \frac{ 1 }{ n P_{t,k} } \boldb_{t,k}^\top \underline{\boldZ}_{t,k}^{-1/2} \boldC_{t,i} \boldC_{t,i}^\top \underline{\boldZ}_{t,k}^{-1/2} \boldb_{t,k} \epsilon_{t,i}^2 > \delta^2 )} \bigg| H_{t-1}^{(n)}, A_{t,i}=k \bigg] 
\boldZ_{t,i}^{-1/2} \boldb_{t,k}
\end{equation*} 
Since $\boldc \in [-u,u]$, by the Gershgorin circle theorem, we can bound the maximum eigenvalue of $\boldc \boldc^\top$ by some constant $a > 0$.
\begin{equation*}
\leq \frac{a}{ n } \sum_{i=1}^{n} \boldb_{t,k}^\top \boldZ_{t,i}^{-1} \boldb_{t,k} \E \bigg[ \epsilon_{t,i}^2 \II_{( \frac{ a }{ n P_{t,k} } \boldb_{t,k}^\top \underline{\boldZ}_{t,k}^{-1} \boldb_{t,k} \epsilon_{t,i}^2 > \delta^2 )} ~\bigg|~ H_{t-1}^{(n)}, A_{t,i}=k \bigg]
\end{equation*}

We define random variable $M_t^{(n)} = \II_{( \forall ~\boldc \in \real^d, ~ \MC{A}_t( H_{t-1}^{(n)}, \boldc) \in [ f(n), 1-f(n) ]^K )}$, representing whether the conditional clipping condition is satisfied.
Note that by our conditional clipping assumption, $M_t^{(n)} \Pto 1$ as $n \to \infty$.
\begin{equation*}
= \frac{a}{ n } \sum_{i=1}^{n} \boldb_{t,k}^\top \underline{\boldZ}_{t,k}^{-1} \boldb_{t,k} \E \bigg[ \epsilon_{t,i}^2 \II_{( \frac{ a }{ n P_{t,k} } \boldb_{t,k}^\top \underline{\boldZ}_{t,k}^{-1} \boldb_{t,k} \epsilon_{t,i}^2 > \delta^2 )} \bigg| H_{t-1}^{(n)}, A_{t,i}=k \bigg] \bigg( M_t^{(n)} + (1-M_t^{(n)}) \bigg)
\end{equation*}
\begin{equation}
\label{eqn:lindeberg1}
= \frac{ a }{ n }  \sum_{i=1}^{n} \boldb_{t,k}^\top \underline{\boldZ}_{t,k}^{-1} \boldb_{t,k} \E \bigg[ \epsilon_{t,i}^2 \II_{( \frac{ a }{ n P_{t,k} } \boldb_{t,k}^\top \underline{\boldZ}_{t,k}^{-1} \boldb_{t,k} \epsilon_{t,i}^2 > \delta^2 )} \bigg| H_{t-1}^{(n)}, A_{t,i}=k \bigg] M_t^{(n)} + o_p(1)
\end{equation}
 
By equation \eqref{eqn:eigenvaluebound}, have that 
\begin{equation*}
\lambda_{\max} ( \underline{\boldZ}_{t,k} ^{-1} ) \leq \frac{1}{ f(n) } \lambda_{\max} \bigg( \big( \underline{ \bs{\Sigma} }_t^{(n)} \big)^{-1} \bigg) \leq \frac{1}{ l ~ f(n) }
\end{equation*}
Recall that $P_{t,k} = \PP( A_{t,i} = k ~|~ H_{t-1}^{(n)} )$, so $P_{t,k} ~|~ (M_t^{(n)} = 1) \geq f(n)$.
Thus we have that equation \eqref{eqn:lindeberg1} is upper bounded by the following:
\begin{equation*}
\leq \frac{1}{ n }  \sum_{i=1}^{n} \frac{ \boldb_{t,k}^\top \boldb_{t,k} }{ l ~ f(n) } \E \bigg[ \epsilon_{t,i}^2 \II_{( \frac{ a }{ n f(n) } \frac{ \boldb_{t,k}^\top \boldb_{t,k} }{ l ~ f(n) } \epsilon_{t,i}^2 > \delta^2 )} \bigg| H_{t-1}^{(n)}, A_{t,i}=k \bigg] + o_p(1)
\end{equation*}
\begin{equation*}
= \frac{1}{ n }  \sum_{i=1}^{n} \frac{ \boldb_{t,k}^\top \boldb_{t,k} }{ l ~ f(n) } \E \bigg[ \epsilon_{t,i}^2 \II_{( \epsilon_{t,i}^2 > \delta^2 \frac{ l ~ n f(n)^2 }{ a \boldb_{t,k}^\top \boldb_{t,k} } )} \bigg| H_{t-1}^{(n)}, A_{t,i}=k \bigg] + o_p(1)
\end{equation*}

It is sufficient to show that 
\begin{equation}
\label{eqn:lindeberg2}
\lim_{n \to \infty} \max_{i \in [1 \colon n]} \frac{ 1 }{ f(n) } \E \bigg[ \epsilon_{t,i}^2 \II_{( \epsilon_{t,i}^2 > \delta^2 \frac{ l ~ n f(n)^2 }{ a \boldb_{t,k}^\top \boldb_{t,k} } )} \bigg| H_{t-1}^{(n)}, A_{t,i}=k \bigg] = 0.
\end{equation}

By Condition \ref{cond:weakmoments},  we have that for all $n \geq 1$,
$\max_{t \in [1 \colon T], i \in [1 \colon n]} \E[ \varphi( \epsilon_{t,i}^2 ) | H_{t-1}^{(n)}, A_{t,i}=k ] < M$.

Since we assume that $\lim_{x \to \infty} \frac{ \varphi(x) }{x} = \infty$, for all $m \geq 1$, there exists a $b_m$ s.t. $\varphi(x) \geq m M x$ for all $x \geq b_m$. 
So, for all $n, t, i$,
\begin{equation*}
M \geq \E[ \varphi( \epsilon_{t,i}^2 ) | H_{t-1}^{(n)}, A_{t,i}=k  ] \geq \E[ \varphi( \epsilon_{t,i}^2 ) \II_{( \epsilon_{t,i}^2 \geq b_m )} | H_{t-1}^{(n)}, A_{t,i}=k  ] 
\end{equation*}
\begin{equation*}
\geq m M \E[ \epsilon_{t,i}^2 \II_{( \epsilon_{t,i}^2 \geq b_m )} | H_{t-1}^{(n)}, A_{t,i}=k  ]
\end{equation*}
Thus, $\max_{i \in [1 \colon n]} \E[ \epsilon_{t,i}^2 \II_{( \epsilon_{t,i}^2 \geq b_m )} | H_{t-1}^{(n)}, A_{t,i}=k  ] \leq \frac{1}{m}$;
so $\lim_{m \to \infty} \max_{i \in [1 \colon n]} \E[ \epsilon_{t,i}^2 \II_{( \epsilon_{t,i}^2 \geq b_m )} | H_{t-1}^{(n)}, A_{t,i}=k  ] = 0$.
Since by our conditional clipping assumption, $f(n) = c$ for some $0 < c \leq \frac{1}{2}$ thus $n f(n)^2 \to \infty$. 
So equation \eqref{eqn:lindeberg2} holds. $\qed$

\clearpage
\begin{corollary}[Asymptotic Normality of the Batched OLS for Margin with Context Statistic]
Assume the same conditions as Theorem \ref{thm:bolscontext}. For any two arms $x, y \in [0 \colon K-1]$ for all $t \in [1 \colon T]$, we have the BOLS estimator for $\bs{ \Delta_{t, x-y} }:= \bs{\beta}_{t,x} - \bs{\beta}_{t,y}$. We show that as $n \to \infty$,
%\begin{equation*}
%\underline{\boldC}_{t,k} := \sum_{i=1}^n \II_{ A_{t,i}^{(n)} = k } \boldC_{t,i} \boldC_{t,i}^\top \in \real^{d \by d}
%\end{equation*}
%%%%%%%%%%%%%%%%%%%%%%%%%%%%%%%
%where $\underline{\boldC}_{t,k} := \sum_{i=1}^n \II_{ A_{t,i}^{(n)} = k } \boldC_{t,i} \boldC_{t,i}^\top \in \real^{d \by d}$. 
\begin{equation*}
\begin{bmatrix} 
\bigg[ \underline{\boldC}^{-1} _{1,x}+ \underline{\boldC}_{1,y}^{-1} \bigg]^{ 1/2 } ( \bs{ \hat{\Delta} }_{1, x-y}^{\BOLS} - \bs{\Delta}_{1, x-y} )  \\
\bigg[ \underline{\boldC}^{-1} _{2,x} + \underline{\boldC}_{2,y}^{-1} \bigg]^{ 1/2 } ( \bs{ \hat{\Delta} }_{2, x-y}^{\BOLS} - \bs{\Delta}_{2, x-y} )   \\
\vdots \\
\bigg[ \underline{\boldC}^{-1} _{T,x} + \underline{\boldC}_{T,y}^{-1} \bigg]^{ 1/2 } ( \bs{ \hat{\Delta} }_{T, x-y}^{\BOLS} - \bs{\Delta}_{T, x-y} )   \\
 \end{bmatrix}
\Dto \N(0, \sigma^2 \under{\bo{I}}_{Td} )
\end{equation*}
\end{corollary}
where 
\begin{equation*}
\bs{ \hat{\Delta} }_{t, x-y} ^{\BOLS} = \bigg[ \underline{\boldC}_{t,x}^{-1} + \underline{\boldC}_{t,y}^{-1} \bigg]^{ -1 } \bigg( \underline{\boldC}_{t,y}^{-1} \sum_{i=1}^n A_{t,i} \boldC_{t,i} R_{t,i} - \underline{\boldC}_{t,x}^{-1} \sum_{i=1}^n (1-A_{t,i}) \boldC_{t,i} R_{t,i} \bigg).
\end{equation*}

\paragraph{Proof:}
By Cramer-Wold device, it is sufficient to show that for any fixed vector $\boldd \in \real^{Td}$ s.t. $\| \boldd \|_2 = 1$, where $\boldd = [ \boldd_1, \boldd_2, ..., \boldd_T]$ for $\boldd_t \in \real^d$,
$\sum_{t=1}^T \boldd_t^\top \big[ \underline{\boldC}_{t,x}^{-1} + \underline{\boldC}_{t,y}^{-1} \big]^{ 1/2 } ( \bs{ \hat{\Delta} }_{t, x-y}^{\BOLS} - \bs{\Delta}_{t, x-y} ) \Dto \N(0, \sigma^2)$, as $n \to \infty$.
\begin{multline*}
\sum_{t=1}^T \boldd_t^\top \bigg[ \underline{\boldC}_{t,x}^{-1} + \underline{\boldC}_{t,y}^{-1} \bigg]^{ 1/2 } ( \bs{ \hat{\Delta} }_{t, x-y} ^{\BOLS} - \bs{\Delta}_{t, x-y} ) \\
= \sum_{t=1}^T \boldd_t^\top \bigg[ \underline{\boldC}_{t,x}^{-1} + \underline{\boldC}_{t,y}^{-1} \bigg]^{ -1/2 } 
\bigg( \underline{\boldC}_{t,y}^{-1} \sum_{i=1}^n A_{t,i} \boldC_{t,i} \epsilon_{t,i} - \underline{\boldC}_{t,x}^{-1} \sum_{i=1}^n (1-A_{t,i}) \boldC_{t,i} \epsilon_{t,i} \bigg)
\end{multline*}
%%%%%%%%%%%%%%%%%%%%%%%%%%%%%%%
By Lemma \ref{lemma:contextratio}, as $n \to \infty$, $\frac{1}{ n P_{t,x} } \underline{\boldZ}_{t,x}^{-1}  \underline{\boldC}_{t,x} \Pto \under{\bo{I}}_d$ and 
 $\frac{1}{ n P_{t,y} } \underline{\boldZ}_{t,y}^{-1} \underline{\boldC}_{t,y} \Pto \under{\bo{I}}_d$, so by Slutsky's Theorem it is sufficient to that as $n \to \infty$, 
 \small
\begin{equation*}
\sum_{t=1}^T \boldd_t^\top \bigg[ \underline{\boldC}_{t,x}^{-1} + \underline{\boldC}_{t,y}^{-1} \bigg]^{ -1/2 } 
\bigg( \frac{1}{ n P_{t,y} } \underline{\boldZ}_{t,y}^{-1} \sum_{i=1}^n A_{t,i} \boldC_{t,i} \epsilon_{t,i} - \frac{1}{ n P_{t,x} } \underline{\boldZ}_{t,x}^{-1} \sum_{i=1}^n (1-A_{t,i}) \boldC_{t,i} \epsilon_{t,i} \bigg)
\Dto \N(0, \sigma^2)
\end{equation*}
\normalsize
%%%%%%%%%%%%%%%%%%%%%%%%%%%%%%%
%We will now show that
%$\big[ \frac{1}{ P_{t,x} } \underline{\boldZ}_{t,x}^{-1} + \frac{1}{ P_{t,y} } \underline{\boldZ}_{t,y}^{-1} \big]^{ -1/2 }  
%\big[ n \underline{\boldC}_{t,x}^{-1} + n \underline{\boldC}_{t,y}^{-1} \big]^{ 1/2 } \Pto \under{\bo{I}}_d$. \\
We know that
$\bigg[ \frac{1}{ P_{t,x} } \underline{\boldZ}_{t,x} ^{-1} + \frac{1}{ P_{t,y} } \underline{\boldZ}_{t,y}^{-1} \bigg]^{ -1/2 }  
\bigg[ \frac{1}{ P_{t,x} } \underline{\boldZ}_{t,x}^{-1} + \frac{1}{ P_{t,y} } \underline{\boldZ}_{t,y}^{-1} \bigg]^{ 1/2 } \Pto \under{\bo{I}}_d$. \\
%%%%%%%%%%%%%%%%%%%%%%%%%%%%%%%
By Lemma \ref{lemma:contextratio} and continuous mapping theorem, $n P_{t,x} \underline{\boldZ}_{t,x} \underline{\boldC}_{t,x}^{-1} \Pto \under{\bo{I}}_d$ and 
 $ n P_{t,y} \underline{\boldZ}_{t,y} \underline{\boldC}_{t,y}^{-1} \Pto \under{\bo{I}}_d$. So by Slutsky's Theorem,
\begin{equation*}
\bigg[ \frac{1}{ P_{t,x} } \underline{\boldZ}_{t,x}^{-1} + \frac{1}{ P_{t,y} } \underline{\boldZ}_{t,y}^{-1} \bigg]^{ -1/2 }  
\bigg[ n \underline{\boldC}_{t,x}^{-1} + n \underline{\boldC}_{t,y}^{-1} \bigg]^{ 1/2 } \Pto \under{\bo{I}}_d
\end{equation*}
%%%%%%%%%%%%%%%%%%%%%%%%%%%%%%%
So, returning to our CLT, by Slutsky's Theorem, it is sufficient to show that as $n \to \infty$, 
\begin{equation*}
\sum_{t=1}^T \boldd_t^\top \bigg[ \frac{1}{ n P_{t,x} } \underline{\boldZ}_{t,x}^{-1} + \frac{1}{ n P_{t,y} } \underline{\boldZ}_{t,y}^{-1} \bigg]^{ -1/2 } 
 \frac{1}{ n P_{t,y} } \underline{\boldZ}_{t,y}^{-1} \sum_{i=1}^n A_{t,i} \boldC_{t,i} \epsilon_{t,i} 
\end{equation*}
\begin{equation*}
- \sum_{t=1}^T \boldd_t^\top \bigg[ \frac{1}{ n P_{t,x} } \underline{\boldZ}_{t,x}^{-1} + \frac{1}{ n P_{t,y} } \underline{\boldZ}_{t,y}^{-1} \bigg]^{ -1/2 } 
 \frac{1}{ n P_{t,x} } \underline{\boldZ}_{t,x}^{-1} \sum_{i=1}^n (1-A_{t,i}) \boldC_{t,i} \epsilon_{t,i} \Dto \N(0, \sigma^2)
 \end{equation*}
 %%%%%%%%%%%%%%%%%%%%%%%%%%%%%%%
The above sum equals the following:
\begin{equation*}
= \sum_{t=1}^T \boldd_t^\top \bigg[ \frac{1}{ n P_{t,x} } \underline{\boldZ}_{t,x}^{-1} + \frac{1}{ n P_{t,y} } \underline{\boldZ}_{t,y}^{-1} \bigg]^{ -1/2 } 
\frac{1}{ \sqrt{ n P_{t,x} } }  \underline{\boldZ}_{t,x}^{-1/2} \bigg( \frac{1}{ \sqrt{ n P_{t,x} } } \underline{\boldZ}_{t,x}^{-1/2} \sum_{i=1}^n A_{t,i} \boldC_{t,i} \epsilon_{t,i} \bigg)
\end{equation*}
\begin{equation*}
- \sum_{t=1}^T \boldd_t^\top \bigg[ \frac{1}{ n P_{t,x} } \underline{\boldZ}_{t,x}^{-1} + \frac{1}{ \sqrt{ n P_{t,x} } } \underline{\boldZ}_{t,y}^{-1} \bigg]^{ -1/2 }  
\frac{1}{ \sqrt{ n P_{t,y} } }  \underline{\boldZ}_{t,y}^{-1/2} \bigg( \frac{1}{ \sqrt{ n P_{t,y} } } \underline{\boldZ}_{t,y}^{-1/2} \sum_{i=1}^n (1-A_{t,i}) \boldC_{t,i} \epsilon_{t,i} \bigg)
\end{equation*}
%%%%%%%%%%%%%%%%%%%%%%%%%%%%%%%
Asymptotic normality holds by the same martingale CLT as we used in the proof of Theorem \ref{thm:bolscontext}. 
The only difference is that we adjust our $\boldb_{t,k}$ vector from Theorem \ref{thm:bolscontext} to the following:
\begin{equation*}
\boldb_{t,k} := \begin{cases}
	\bo{0} & \TN{ if } k \not\in \{ x, y \} \\
	\boldd_t^\top \bigg[ \frac{1}{ n P_{t,x} } \underline{\boldZ}_{t,x}^{-1} + \frac{1}{ n P_{t,y} } \underline{\boldZ}_{t,y}^{-1} \bigg]^{ -1/2 }  \frac{1}{ \sqrt{ n P_{t,x} } } \underline{\boldZ}_{t,x}^{-1/2} & \TN{ if } k = x \\
	\boldd_t^\top \bigg[ \frac{1}{ n P_{t,x} } \underline{\boldZ}_{t,x}^{-1} + \frac{1}{ n P_{t,y} } \underline{\boldZ}_{t,y}^{-1} \bigg]^{ -1/2 }  \frac{1}{ \sqrt{ n P_{t,y} } } \underline{\boldZ}_{t,y}^{-1/2} & \TN{ if } k = y \\
\end{cases}
\end{equation*}
The proof still goes through with this adjustment because for all $k \in [0 \colon K-1]$,
(i) $\boldb_{t,k} \in H_{t-1}^{(n)}$,
(ii) $\sum_{t=1}^T \sum_{k=0}^{K-1} \boldb_{t,k}^\top \boldb_{t,k} = \sum_{t=1}^T \boldd_t^\top \boldd_t = 1$.
and (iii) $\frac{ l ~ n f(n)^2 }{ a \boldb_{t,k}^\top \boldb_{t,k} } \to \infty$ still holds because $\boldb_{t,k}^\top \boldb_{t,k}$ is bounded above by one. 
$\qed$

%% file: appendix/wdecorrelated.tex
\section{W-Decorrelated Estimator \cite{deshpande}}
\label{appendix:wdecorrelated}

To better understand why the W-decorrelated estimator has relatively low power, but is still able to guarantee asymptotic normality, we now investigate the form of the W-decorrelated estimator in the two-arm bandit setting.

\subsection{Decorrelation Approach}
We now assume we are in the unbatched setting (i.e., batch size of one), as the W-decorrelated estimator was developed for this setting; 
however, these results easily translate to the batched setting.
We now let $n$ index the number of samples total (previously this was $nT$) and examine asymptotics as $n \to \infty$.
We assume the following model:
$$\boldR_n = \under{\boldX}_n^\top \bs{\beta} + \bs{\epsilon}_n$$
where $\boldR_n, \bs{\epsilon}_n \in \real^n$ and $\under{ \boldX }_n \in \real^{n \by p}$ and $\bs{\beta} \in \real^p$.
The W-decorrelated OLS estimator is defined as follows:
$$\bs{\betahat}^d = \bs{\betahat}_{\OLS} + \under{\boldW}_n (\boldR_n - \under{\boldX}_n \bs{\betahat}_{\OLS})$$
%%%%%%%%%%%%%%%%%%%%%%%%%%%%%%%%%%%%%%%%%%%%
With this definition we have that,
$$\bs{\betahat}^d - \bs{\beta} = \bs{\betahat}_{\OLS} + \under{\boldW}_n (\boldR_n - \under{\boldX}_n \bs{\betahat}_{\OLS}) - \bs{\beta}$$
$$= \bs{\betahat}_{\OLS} + \under{\boldW}_n (\under{\boldX}_n \bs{\beta} + \bs{\epsilon}_n) - \under{\boldW}_n \under{\boldX}_n \bs{\betahat}_{\OLS} - \bs{\beta}$$
$$= (\under{\boldI}_p - \under{\boldW}_n \under{\boldX}_n) (\bs{\betahat}_{\OLS} - \bs{\beta}) + \under{\boldW}_n \bs{ \epsilon }_n$$
%%%%%%%%%%%%%%%%%%%%%%%%%%%%%%%%%%%%%%%%%%%%
Note that if $\E[ \under{\boldW}_n \bs{ \epsilon }_n] = \E \big[ \sum_{i=1}^n \boldW_i \epsilon_i \big] = 0$ (where $\boldW_i$ is the $i^{th}$ column of $\under{\boldW}_n$), then 
$\E[ (\under{\boldI}_p - \under{\boldW}_n \under{\boldX}_n) (\bs{\betahat}_{\OLS} - \bs{\beta}) ]$ would be the bias of the estimator.
We assume $\{ \epsilon_i \}$ is a martingale difference sequence w.r.t. filtration $\{ \G_i \}_{i=1}^n$.
Thus, if we constrain $\boldW_i$ to be $\G_{i-1}$ measurable, 
$$\E[ \under{\boldW}_n \bs{\epsilon}_n] = \E \big[ \sum_{i=1}^n \boldW_i \epsilon_i \big] = \sum_{i=1}^n \E \bigg[ \E[ \boldW_i \epsilon_i | \G_{i-1} ] \bigg] = \sum_{i=1}^n \E \bigg[ \boldW_i \E[ \epsilon_i | \G_{i-1} ] \bigg] = 0$$

\paragraph{Trading off Bias and Variance}
While decreasing $\E[ (\under{\boldI}_p - \under{\boldW}_n \under{\boldX}_n) (\bs{\betahat}_{\OLS} - \bs{\beta}) ]$ will decrease the bias, making $\under{\boldW}_n$ larger in norm will increase the variance.
So the trade-off between bias and variance can be adjusted with different values of $\lambda$ for the following optimization problem:
$$|| \under{\boldI}_p - \under{\boldW}_n \under{\boldX}_n \|_F^2 + \lambda \| \under{\boldW}_n \|_F^2
= \| \under{\boldI}_p - \under{\boldW}_n \under{\boldX}_n \|_F^2 + \lambda \Tr (\under{\boldW}_n \under{\boldW}_n^\top)$$

\paragraph{Optimizing for $\under{\boldW}_n$}
The authors propose to optimize for $\under{\boldW}_n$ in a recursive fashion, so that the $i^{th}$ column, $\boldW_i$, only depends on $\{ \boldX_j \}_{j \leq i} \cup \{ \epsilon_j \}_{j \leq i-1}$ (so $\sum_{i=1}^n \E[ \boldW_i \epsilon_i ] = 0$). 
We let $\boldW_0 = 0$, $\boldX_0 = 0$, and recursively define $\under{\boldW}_n := [\under{\boldW}_{n-1} \boldW_n]$ where
$$\boldW_n = \argmin_{\boldW \in \real^p} \| \under{\boldI}_p - \under{\boldW}_{n-1} \under{\boldX}_{n-1} - \boldW \boldX_n^\top \|^2_F + \lambda \| \boldW \|^2_2$$
where $\under{\boldW}_{n-1} = [ \boldW_1; \boldW_2; ...; \boldW_{n-1} ] \in \real^{p \by (n-1)}$ and $\under{\boldX}_{n-1} = [ \boldX_1; \boldX_2; ...; \boldX_{n-1} ]^\top \in \real^{(n-1) \by p}$.
Now, let us find the closed form solution for each step of this minimization:
$$\frac{d}{d \boldW} \| \under{\boldI}_p - \under{\boldW}_{n-1} \under{\boldX}_{n-1} - \boldW \boldX_n^\top \|^2_F + \lambda \| \boldW \|^2_2
= 2 ( \under{\boldI}_p - \under{\boldW}_{n-1} \under{\boldX}_{n-1} - \boldW \boldX_n^\top ) (- \boldX_n) + 2 \lambda \boldW$$
%%%%%%%%%%%%%%%%%%%%%%%%%%%%%%%%%%%%%%%%%%%%
Note that since the Hessian is positive definite, so we can find the minimizing $\boldW$ by setting the first derivative to $0$:
$$\frac{d^2}{d \boldW d \boldW^\top} \| \under{\boldI}_p - \under{\boldW}_{n-1} \under{\boldX}_{n-1} - \boldW \boldX_n^\top \|^2_F + \lambda \| \boldW \|^2_2
= 2 \boldX_n \boldX_n^\top + 2 \lambda \under{\boldI}_p \succcurlyeq 0$$

$$0 = 2 ( \under{\boldI}_p - \under{\boldW}_{n-1} \under{\boldX}_{n-1} - \boldW \boldX_n^\top ) (- \boldX_n) + 2 \lambda \boldW$$
$$( \under{\boldI}_p - \under{\boldW}_{n-1} \under{\boldX}_{n-1} - \boldW \boldX_n^\top ) \boldX_n = \lambda \boldW$$
$$( \under{\boldI}_p - \under{\boldW}_{n-1} \under{\boldX}_{n-1}) \boldX_n = \lambda \boldW + \boldW \boldX_n^\top \boldX_n
= ( \lambda + \| \boldX_n \|^2_2 ) \boldW$$
$$\boldW^* = ( \under{\boldI}_p - \under{\boldW}_{n-1} \under{\boldX}_{n-1}) \frac{\boldX_n}{\lambda + \| \boldX_n \|^2_2}$$

%\paragraph{How is $\lambda$ set?}
%According to the theory of \cite{deshpande}, if $\lambda$ is set to be a high probability lower bound on the minimum eigenvalue of the Gram matrix
%( i.e. $P(  \lambda_{\min} ( \under{\boldX}_n \under{\boldX}_n^\top ) \leq \lambda \log \log n) \leq \frac{1}{n}$), then the bias of the estimator will be dominated by the variance. 
%In their experiments, they set $\lambda$ such that $P(  \lambda_{\min} ( \under{\boldX}_n \under{\boldX}_n^\top ) \leq \lambda ) \leq 0.05$. 

\begin{proposition}[W-decorrelated estimator and time discounting in the two-arm bandit setting]
%\subsection{Proposition 4: Time discounting in 2-arm bandit setting}
Suppose we have a 2-arm bandit. $A_i$ is an indicator that equals 1 if arm 1 is chosen for the $i^{th}$ sample, and 0 if arm 0 is chosen.
We define $\boldX_i := [ 1-A_i, A_i ] \in \real^2$.
We assume the following model of rewards:
$$R_i = \boldX_i^\top \bs{\beta} + \epsilon_i
= A_i \beta_1 + (1-A_i) \beta_0 + \epsilon_i$$
We further assume that $\{ \epsilon_i \}_{i=1}^n$ are a martingale difference sequence with respect to filtration $\{ \G_i \}_{i=1}^n$.
We also assume that $\boldX_i$ are non-anticipating with respect to filtration $\{ \G_i \}_{i=1}^n$.
Note the W-decorrelated estimator:
$$\bs{\betahat}^d = \bs{\betahat}_{\OLS} + \under{\boldW}_n (\boldR_n - \under{\boldX}_n \bs{\betahat}_{\OLS})$$
We show that for $\under{\boldW}_n = [ \boldW_1; \boldW_2; ...; \boldW_n ] \in \real^{p \by n}$ and choice of constant $\lambda$,
$$\boldW_i = \begin{bmatrix}
(1- \frac{1}{ \lambda + 1 })^{ \sum_{i=1}^n (1-A_i) } \frac{1}{ \lambda + 1 } \\
(1- \frac{1}{ \lambda + 1 })^{ \sum_{i=1}^n A_i } \frac{1}{ \lambda + 1 } 
\end{bmatrix} \in \real^2$$
Moreover, we show that the W-decorrelated estimator for the mean of arm 1, $\beta_1$, is as follows:
$$\betahat_1^d= \bigg( 1 - \sum_{i=1}^n A_t \frac{1}{\lambda + 1} \bigg( 1- \frac{1}{\lambda + 1} \bigg)^{N_{1,i} - 1} \bigg) \betahat_1^{\OLS} + \sum_{i=1}^n A_t R_t \cdot \frac{1}{\lambda + 1} \bigg( 1-\frac{1}{\lambda + 1} \bigg)^{N_{1,i} - 1}$$ 
where $\betahat_1^{\OLS} = \frac{ \sum_{i=1}^n A_i R_i }{ N_{1,n} }$ for $N_{1,n} = \sum_{i=1}^n A_i$.
Since \cite{deshpande} require that $\lambda \geq 1$ for their CLT results to hold, thus, the W-decorrelated estimators is down-weighting samples drawn later on in the study and up-weighting earlier samples.
\end{proposition}

\paragraph{Proof:}
Recall the formula for $\boldW_i$,
$$\boldW_i = ( \under{\boldI}_p - \under{\boldW}_{i-1} \under{\boldX}_{i-1}) \frac{\boldX_i}{\lambda + \| \boldX_i \|^2_2}$$
We let $\boldW_i = [ W_{0,i}, W_{1,i} ]^\top$. For notational simplicity, we let $r = \frac{1}{\lambda + 1}$. We now solve for $W_{1,n}$:
$$W_{1,1} = (1 - 0) \cdot r A_1= r A_1$$
$$W_{1,2} = (1 - W_{1,1} \cdot A_1 ) \cdot r A_2 = ( 1 - r A_1 ) r A_2$$
$$W_{1,3} = \bigg( 1 - \sum_{i=1}^2 \boldW_{1,i} \cdot A_i \bigg) \cdot r A_3
= \bigg( 1 - r A_1 - ( 1 - r A_1 ) r A_2 \bigg) \cdot r A_3
= ( 1 - r A_1 ) ( 1- r A_2 ) \cdot r A_3$$
$$W_{1,4} =  \bigg( 1 - \sum_{i=1}^3 \boldW_{1,i} \cdot A_i \bigg) \cdot r A_4
= \bigg( 1 - r A_1 - ( 1 - r A_1 ) r A_2 - ( 1 - r A_1 ) ( 1- r A_2 ) \cdot r A_3 \bigg) \cdot r A_4$$
$$= ( 1 - r A_1 ) \big( 1 - r A_2 - ( 1 - r A_2 ) r A_3 ) \big) \cdot r A_4
= ( 1 - r A_1 ) ( 1 - r A_2 ) ( 1 - r A_3 ) \cdot r A_4$$
We have that for arbitrary $n$, 
$$W_{1,n} = \bigg( 1 - \sum_{i=1}^{n-1} \boldW_{1,i} \cdot A_i \bigg) \cdot r A_n
= r A_n \prod_{i=1}^{n-1} ( 1 - r A_i )
= r A_n ( 1 - r )^{ \sum_{i=1}^{n-1} A_i }
= r A_n ( 1 - r )^{ N_{1,n-1} }$$
By symmetry, we have that 
$$W_{0,n} = \bigg( 1 - \sum_{i=1}^{n-1} \boldW_{1,i} \cdot (1-A_i) \bigg) \cdot r (1-A_n)
= r (1-A_n) ( 1 - r )^{ N_{0,n-1} }$$
%%%%%%%%%%%%%%%%%%%%%%5
Note the W-decorrelated estimator for $\beta_1$:
$$\betahat_1^d = \betahat_1^{\OLS} + \sum_{i=1}^n A_i \bigg( R_i - \betahat_1^{\OLS} \bigg)  r (1-r)^{N_{1,i-1}}$$
$$= \bigg( 1 - \sum_{i=1}^n A_i r (1-r)^{N_{1,i-1}} \bigg) \betahat_1^{\OLS} + \sum_{i=1}^n A_i R_i \cdot r (1-r)^{N_{1,i-1}} ~~~ \square$$